\documentclass[twoside,11pt]{article}

\usepackage{jmlr2e}
\usepackage{booktabs}
\usepackage{multirow}
\usepackage{multicol}
\usepackage{wrapfig}
\usepackage{subcaption}
\usepackage{amsmath}

\jmlrheading{1}{}{}{}{}{}

\firstpageno{1}

\begin{document}

\title{Demystifying the Base and Novel Performances\\for Few-shot Class-incremental Learning}

       
\author{\name Jaehoon Oh \email jhoon.oh@kaist.ac.kr \\
       \addr Graduate School of Data Science, KAIST, Korea
       \AND
       \name Se-Young Yun \email yunseyoung@kaist.ac.kr \\
       \addr Graduate School of Artificial Intelligence, KAIST, Korea}


\maketitle

\begin{abstract}
Few-shot class-incremental learning (FSCIL) has addressed challenging real-world scenarios where unseen novel classes continually arrive with few samples.
In these scenarios, it is required to develop a model that recognizes the novel classes without forgetting prior knowledge.
In other words, FSCIL aims to maintain the base performance and improve the novel performance simultaneously.
However, there is little study to investigate the two performances separately.
In this paper, we first decompose the entire model into four types of parameters and demonstrate that the tendency of the two performances varies greatly with the updated parameters when the novel classes appear.
Based on the analysis, we propose a simple method for FSCIL, coined as NoNPC, which uses normalized prototype classifiers without further training for incremental novel classes. It is shown that our straightforward method has comparable performance with the sophisticated state-of-the-art algorithms.
\end{abstract}

\begin{keywords}
  Few-shot class-incremental learning, parameter decomposition, prototype classifier
\end{keywords}

\section{Introduction}\label{sec:intro}
Deep learning has achieved considerable success under the IID\,(independent and identically distribution) and stationary assumption. However, in real-world scenarios, dynamic and open environments are more natural, discouraging practitioners from developing deep models in many applications.
To address this concern, class-incremental learning (CIL) has gained much attention\,\citep{belouadah2019IL2M, masana2020class, zhu2021class, shim2021online,mai2022online}, where the unseen \emph{novel} classes continually appear.
CIL aims to learn novel classes without forgetting the past knowledge, where previously seen data are not available due to privacy\,\citep{mai2022online,joseph2022energy} and memory\,\citep{fini2020online} issues.
%
%

Conventional CIL methods have been studied based on a large amount of data for the novel classes.
However, when only a small amount of training data for the novel classes is available, they are known to suffer from severe performance degradation\,\citep{shi2021overcoming}.  
To address this challenging scenario, \emph{few-shot class-incremental learning} (FSCIL)\,\citep{tao2020few, zhang2021few, shi2021overcoming, zhou2022forward} has emerged.
FSCIL learns base classes consisting of a huge amount of data (i.e., \emph{many-shot}), and then learns novel classes incrementally with very few data (i.e., \emph{few-shot}). 
In this regime, FSCIL aims to recognize novel classes from few novel data that arrive sequentially without forgetting the knowledge of base classes.
Namely, it is required to maintain the base performance and improve the novel performance simultaneously.

Recent methods for FSCIL have improved the \emph{weighted} performance, which weighs the \emph{base} and \emph{novel} performances based on the number of base and novel classes.
However, the improved weighted performance does not mean that \emph{base} and \emph{novel} performances are both improved.
%
In this paper, we analyze which components in a model are responsible for the base and novel performances by parameter decomposition. Based on this analysis, we propose a simple yet effective method, \textbf{NoNPC}, which uses normalized prototype classifiers without training when novel classes appear. This simple algorithm achieves comparable performance to SOTA algorithms.

\section{Related Work}\label{sec:related}

\begin{figure}[t]
    \centering
    \includegraphics[width=\linewidth]{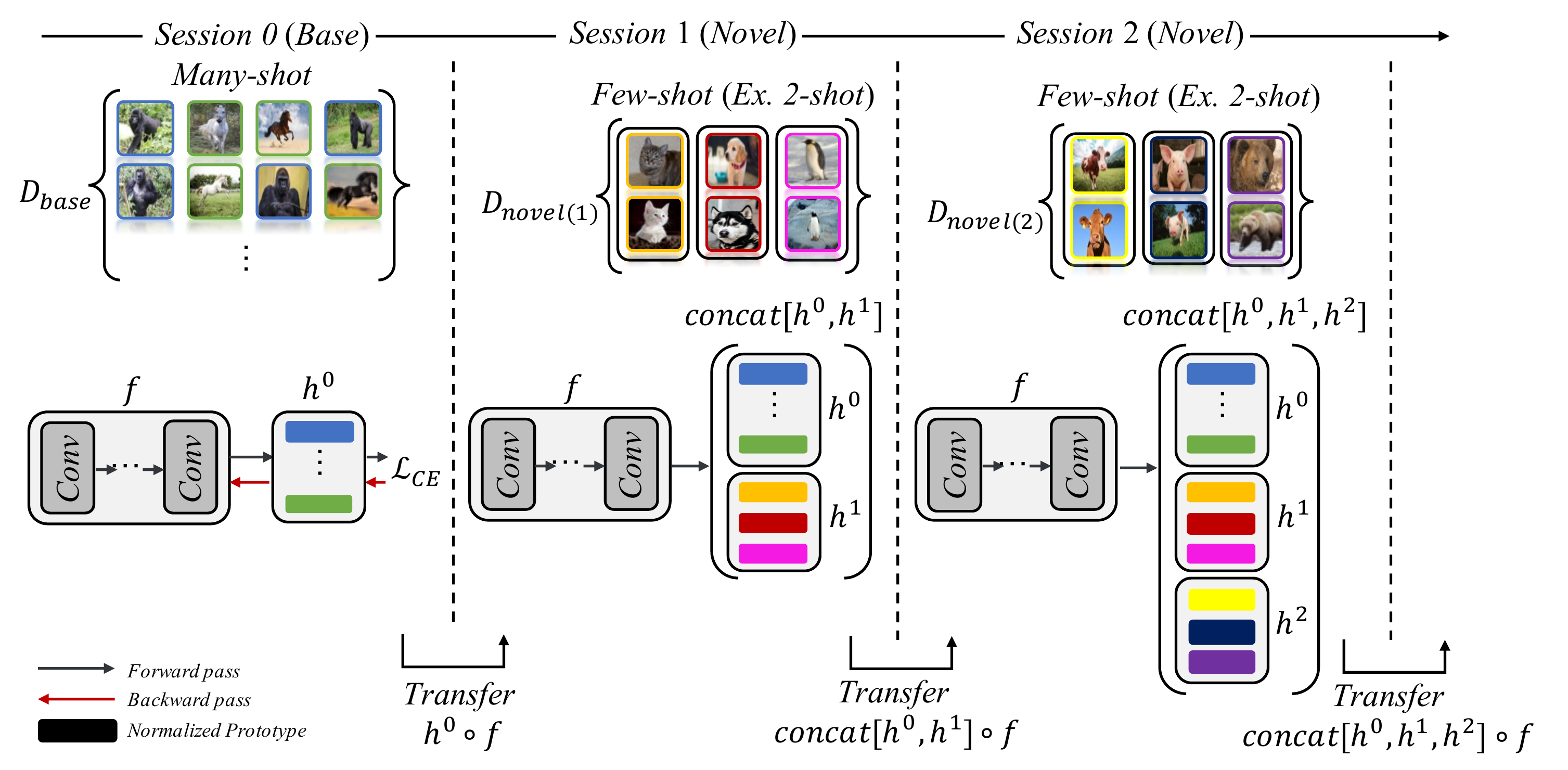}
    \caption{Overview of NoNPC (\textbf{No} Training with \textbf{N}ormalized \textbf{P}rototype \textbf{C}lassifiers). $f$ indicates a feature extractor. $h^{0}$ and $h^{i}$ indicate classifiers for base and novel classes for session $i$ ($i > 0$), respectively. Our algorithm does not learn a model during novel sessions.}
    \label{fig:fscil_overview}
    \vspace*{-5pt}
\end{figure}

\textbf{Few-shot Class-incremental Learning (FSCIL).}
FSCIL is a combination of class-incremental learning (CIL) and few-shot learning (FSL), aiming to incrementally update a classifier with limited data from novel classes to discriminate all classes seen before.
The incremental learning procedure consists of a base session followed by novel sessions.
Recent FSCIL methods \citep{tao2020few,mazumder2021few,shi2021overcoming,zhang2021few,zhou2022forward} can be categorized by which session they focus on, i.e., base or novel session.
The former prepares to learn novel classes in the base session by revising the standard training process in the base session\,\citep{shi2021overcoming, zhou2022forward} or training an additional network for the incoming novel classes\,\citep{zhang2021few}.
On the other hand, the latter devise how to fine-tune a model incrementally when they encounter unseen classes in the novel sessions by imposing regularization on fine-tuning in the novel session\,\citep{mazumder2021few, hersche2022cfscil}.
Previous studies focus on improving the weighted performance, while we analyze the \emph{base} and \emph{novel} performances separately and propose a simple yet effective method without additional costs.

\smallskip
\noindent \textbf{Parameter Decomposition.}
Parameter decomposition is widely used technique for in-depth analysis or improved algorithms in many fields such as long-tailed distribution\,\citep{Kang2020Decoupling, yu2020devil}, federated learning\,\citep{arivazhagan2019federated, collins2021exploiting, oh2022fedbabu}, meta-learning\,\citep{lee2018gradient, Raghu2020Rapid, oh2021boil}, and continual learning\,\citep{shi2021overcoming, davari2022probing}.
Most of the prior works decompose the entire model into two parts, feature extractors and classifiers.
Feature extractors can be transferred well, while classifiers are easily distorted under non-IID environments, meaning that they are susceptible to bias\,\citep{Kang2020Decoupling, oh2022fedbabu}.
We decompose entire parameters into four parts under FSCIL environments: extractors and three types of classifiers, described in Table~\ref{tab:notation}, to investigate the base and novel performances for FSCIL.
\section{Problem Setup}\label{sec:setup}

In this section, we formally summarize the problem setup.
The procedure of FSCIL includes continuous sessions, where each session consists of training and evaluation.
For the first session that we call base session\,(defined as session 0), a model $h^0 \circ f$ is trained using $D_{base}$ consisting of base classes, where $f$ is a feature extractor and $h^0$ is a classifier for the base classes.
Note that during the base session, we can use abundant data from base classes.
Let $x$ be an input and $d$ be the output dimension of $f$, then $f(x) \in \mathbb{R}^{d}$ and $h^0 \in \mathbb{R}^{K \times d}$ where $K$ is the number of base classes.
We define $h^0_j$ as the $j$-th row vector of $h^0$.

\begin{table}[t]
    \centering
    \footnotesize
    \caption{Notations of decomposed parameters and models according to the update parts for the current novel session $i$ ($i>0$).}\label{tab:notation}
    \vspace*{-0.15cm}
    \begin{tabular}{ll|ll}
    \toprule
    Notation & Description & Model & Description \\
    \midrule
    $f$      & Feature extractor & M1 & No update \\ 
    $h^0$    & Classifier for base classes & M2 & Update $h^i$ \\
    \multirow{2}{*}{$h^{1:i-1}$} & Classifiers for novel classes before session $i$ & M3 & Update $h^{1:i-1}$ and $h^i$ \\
             & (\emph{i.e.}, $\{h^{1}, \cdots, h^{i-1}\}$) & M4 & Update $h^0, h^{1:i-1}$ and $h^i$ \\
    $h^{i}$  & Classifier for novel classes in session $i$ & M5 & Update $f, h^0, h^{1:i-1}$ and $h^i$ \\
    \bottomrule
    \end{tabular}
\end{table}

For the subsequent session $i$ ($i \in \{1, \cdots, S\}$) that we call \emph{novel} sessions, an extended model $({concat}$ $[h^0, h^1, \cdots, h^i]) \circ f$ is trained using $D_{novel(i)}$ consisting of novel classes, where $h^i$ is a classifier for novel classes in session $i$.
Unlike $D_{base}$, $D_{novel(i)}$ consists of few samples, in general $nk$ samples, where $n$ is the number of incremental novel classes and $k$ is the number of samples per class. $k$ is 5 in our experiments.
The incrementally extended model $({concat}[h, h^1, \cdots, h^i]) \circ f$ is evaluated for both base and incremental novel classes until session $i$.
After \emph{S} sessions, the final classifier $({concat}[h, h^1, \cdots, h^S])$ is in $\mathbb{R}^{(K+nS) \times d}$. 

For evaluation, we use three metrics: base, novel, and weighted performances. The base and novel performance indicate accuracy on base and novel classes, respectively. These performances are weighted based on the number of classes for calculating the weighted performance.
All results are averaged by five runs.
The detailed implementation is described in Appendix~\ref{appx:detail}.
\section{Analysis: Parameter Decomposition during Novel Sessions}\label{sec:decompose}

\begin{figure}[t]
    \centering
    \includegraphics[width=0.45\linewidth]{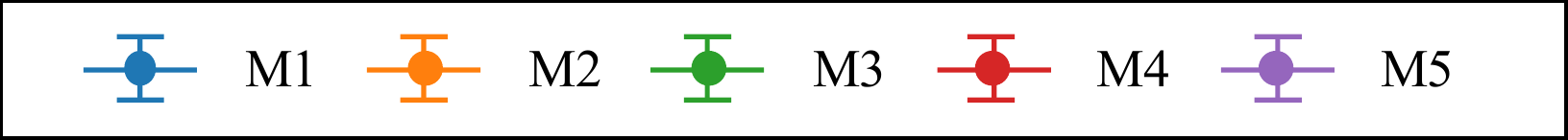}
    
    \begin{subfigure}[h]{0.32\linewidth}
    \includegraphics[width=\linewidth]{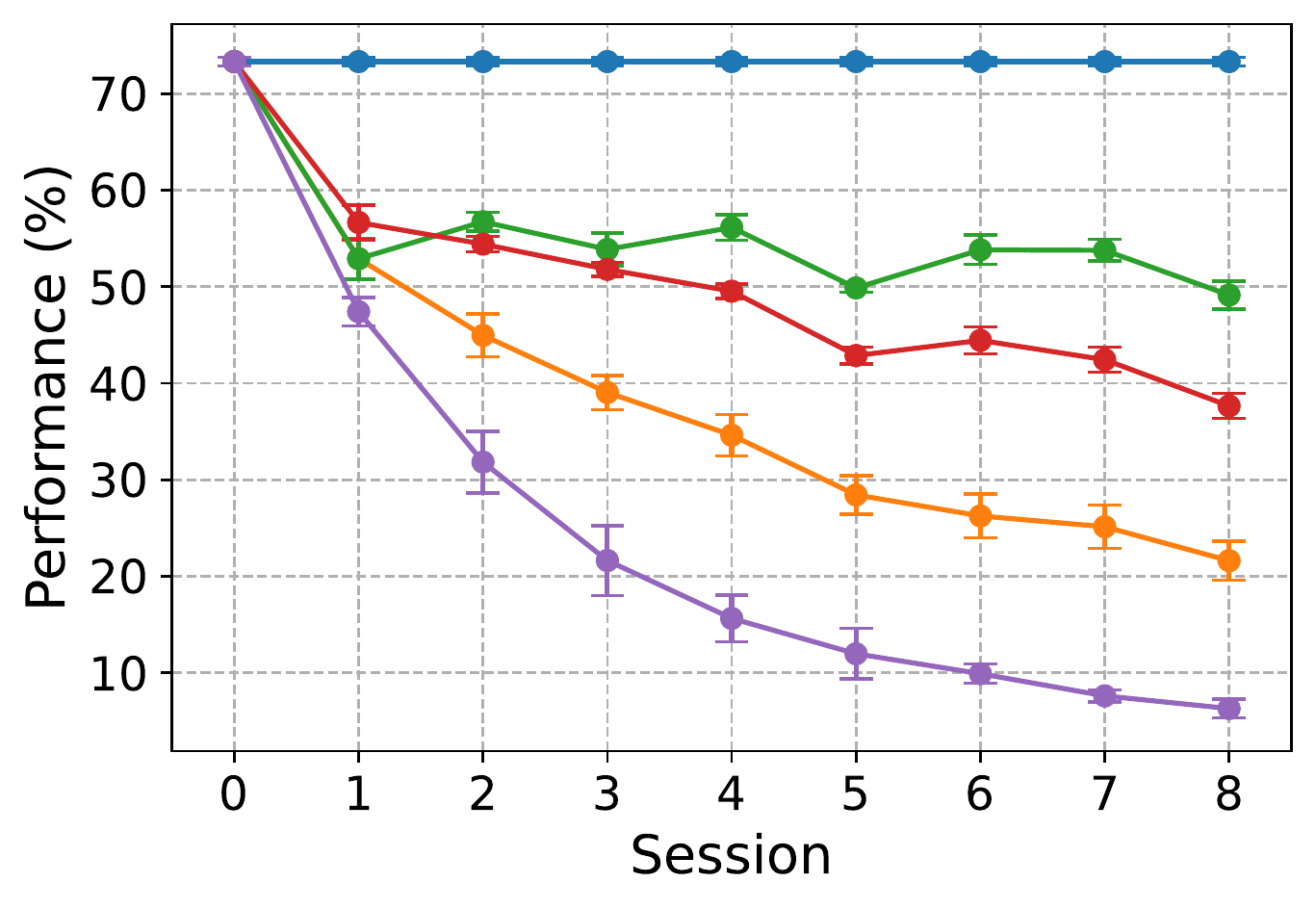}
        \caption{Base}
    \end{subfigure}
    \hfill
    \begin{subfigure}[h]{0.32\linewidth}
    \includegraphics[width=\linewidth]{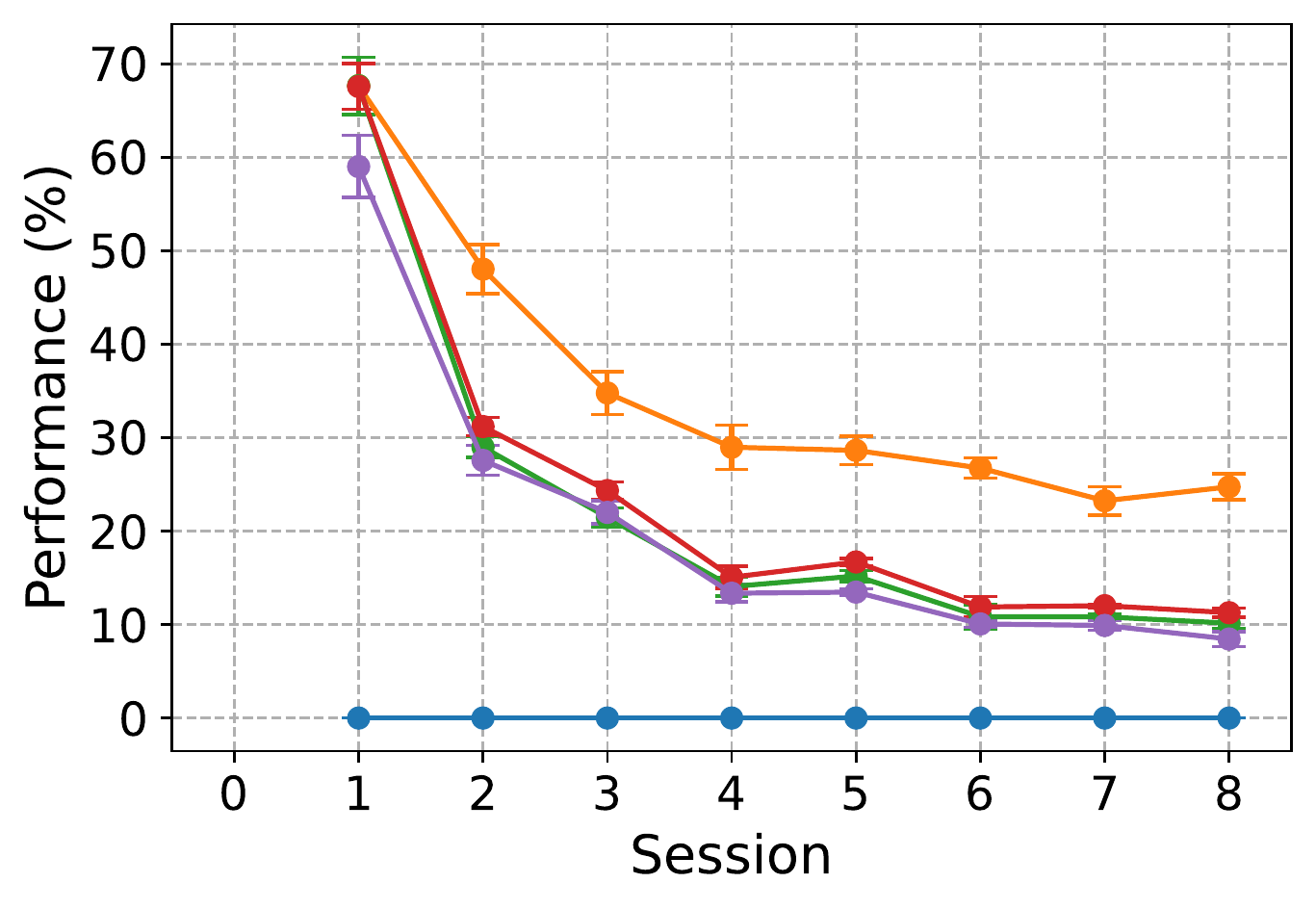}
        \caption{Novel}
    \end{subfigure}
    \hfill
    \begin{subfigure}[h]{0.32\linewidth}
    \includegraphics[width=\linewidth]{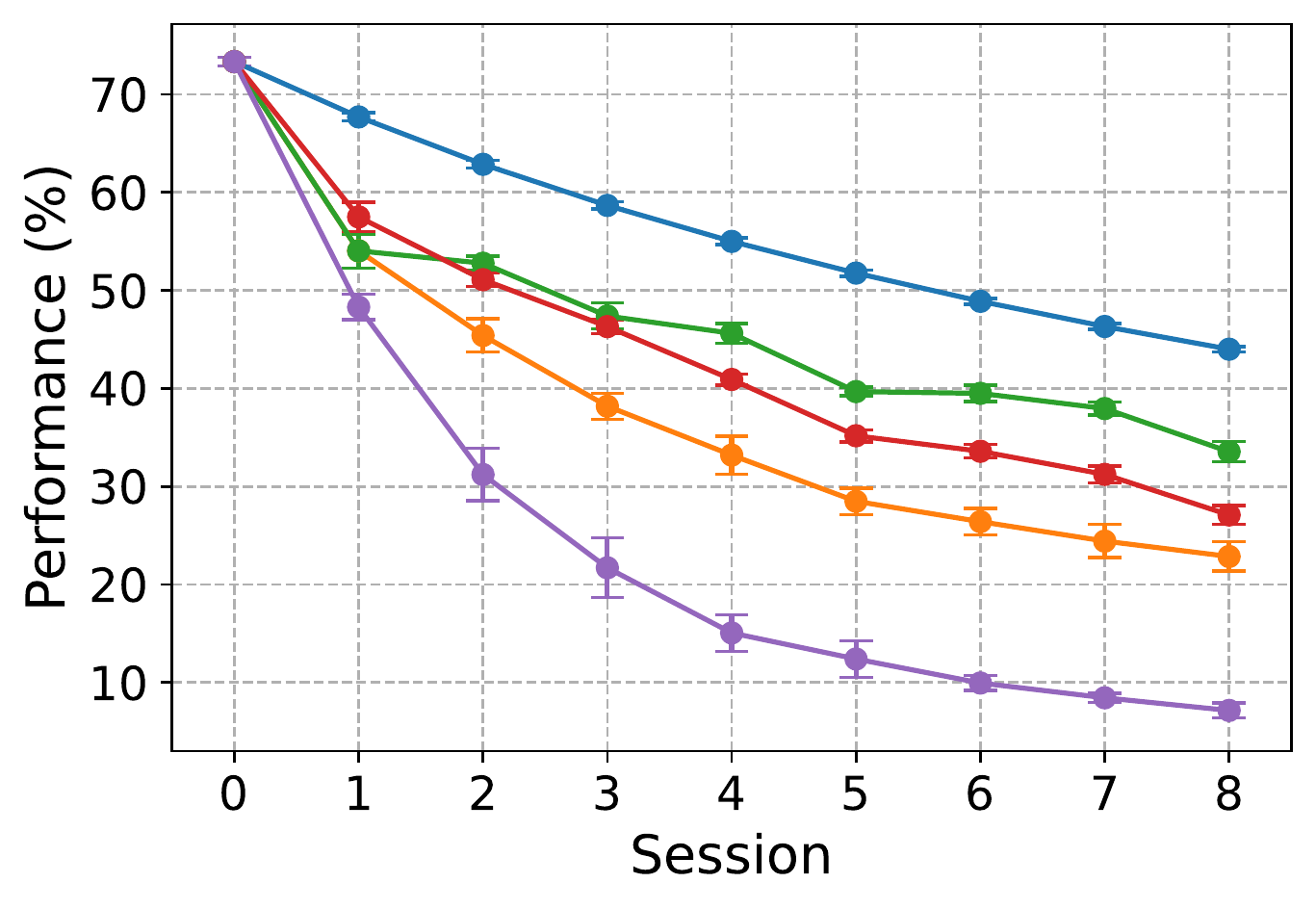}
        \caption{Weighted}
    \end{subfigure}
    \vspace*{-3pt}
    \caption{Base, novel, and weighted performances according to the update parts on CIFAR100. Models are described in Table~\ref{tab:notation}.}
    \label{fig:cifar100_decomp}
    \vspace*{-8pt}
\end{figure}

\begin{figure}[t]
    \centering
    \begin{subfigure}[h]{0.195\linewidth}
    \includegraphics[width=\linewidth]{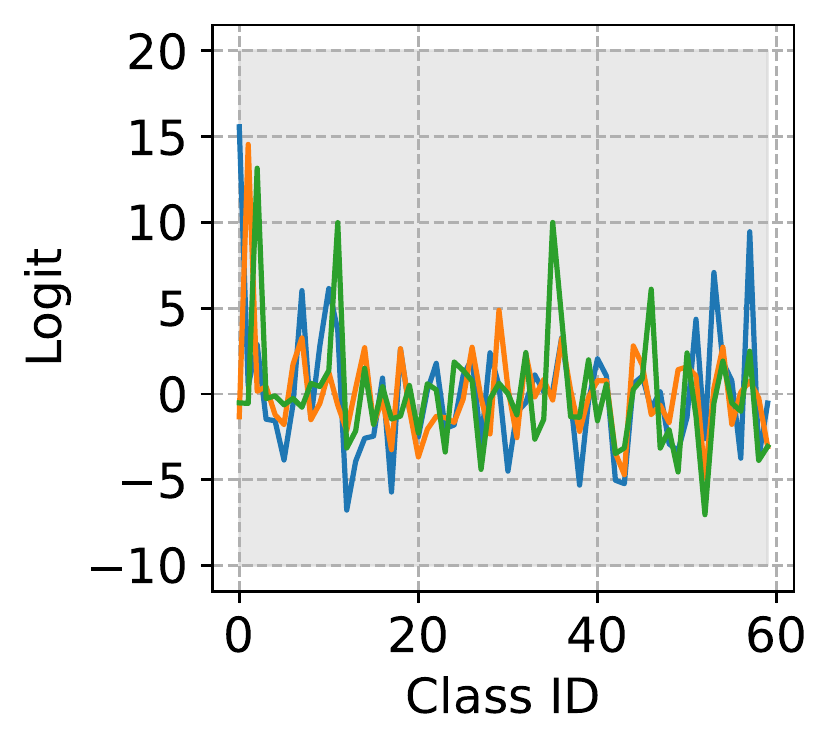}
        \caption{Session 0}
    \end{subfigure}
    \begin{subfigure}[h]{0.18\linewidth}
    \includegraphics[width=\linewidth]{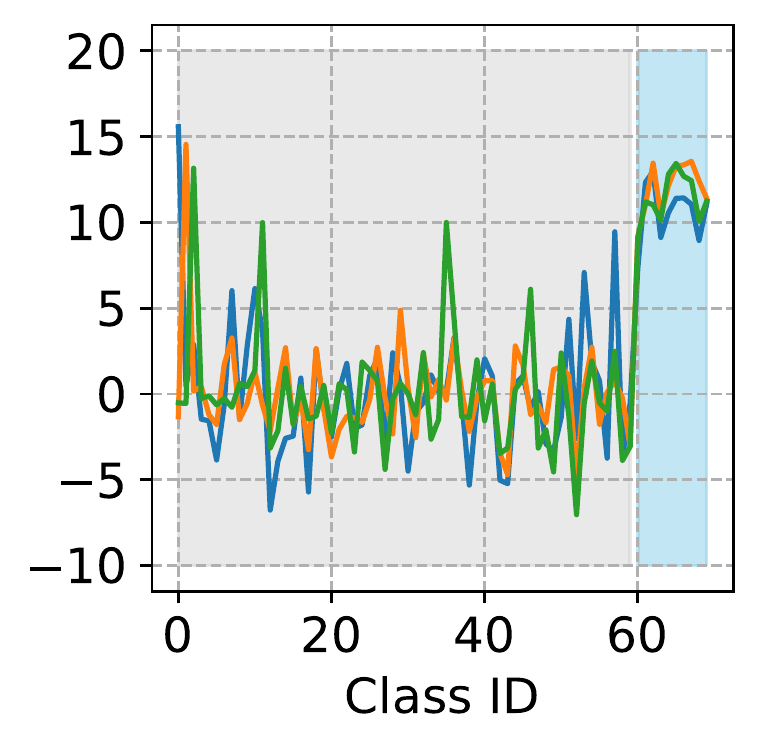}
        \caption{Session 2}
    \end{subfigure}
    \begin{subfigure}[h]{0.18\linewidth}
    \includegraphics[width=\linewidth]{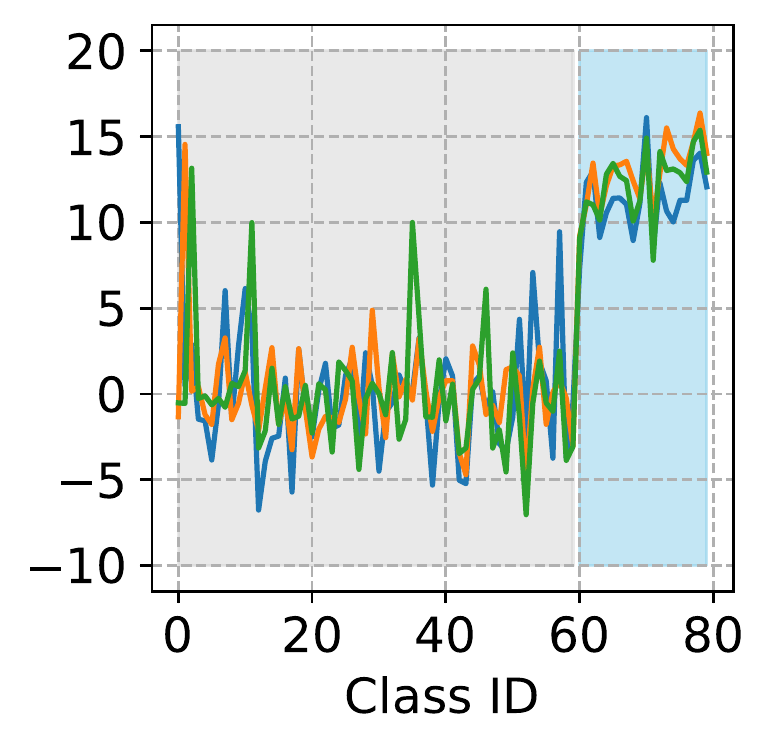}
        \caption{Session 4}
    \end{subfigure}
    \begin{subfigure}[h]{0.18\linewidth}
    \includegraphics[width=\linewidth]{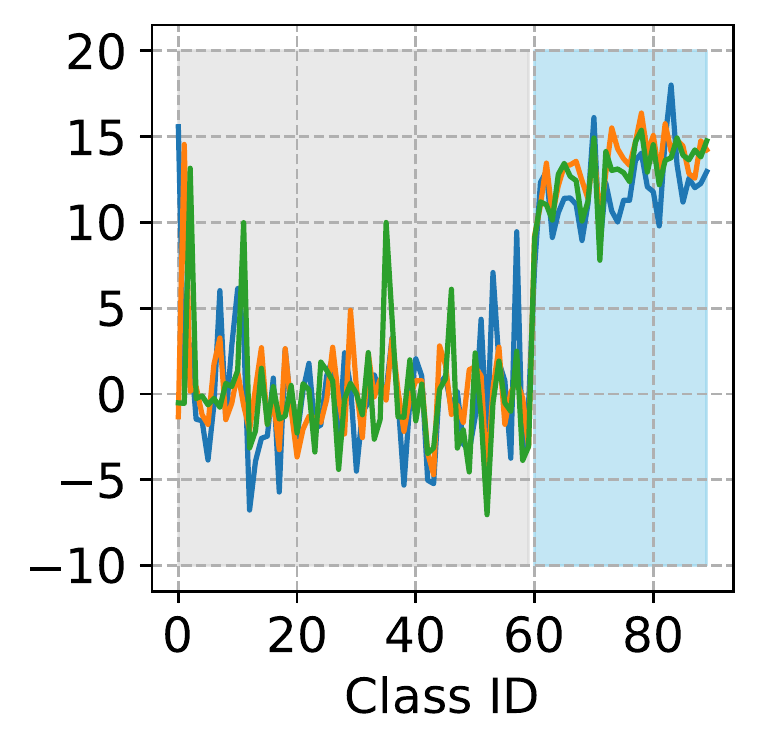}
        \caption{Session 6}
    \end{subfigure}
    \begin{subfigure}[h]{0.18\linewidth}
    \includegraphics[width=\linewidth]{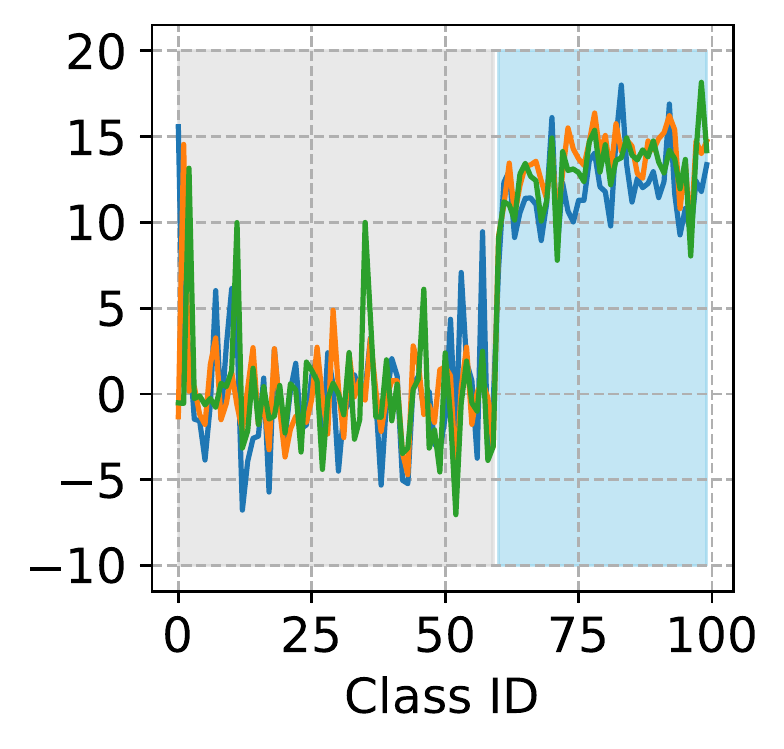}
        \caption{Session 8}
    \end{subfigure}
    \vspace*{-3pt}
    \caption{Logit distributions of base samples according to the session on CIFAR100 when training only $h^i$. Gray and blue backgrounds indicate the base and novel classes, respectively. Each line is the average of logits of samples belonging to class 0, 1, and 2.}
    \label{fig:cifar100_logit}
    \vspace*{-8pt}
\end{figure}

To investigate which parameters are relevant to each performance, we decompose the entire model into four update parts described in Section~\ref{sec:setup}, during novel sessions: $f$, $h^0$, $h^{1:i-1}$, and $h^i$.
Figure~\ref{fig:cifar100_decomp} describes the base, novel, and weighted performances according to the decomposed update parts on CIFAR100.
Following \citet{tao2020few}, for the cases where an encoder $f$ is not updated, the running statistics of batch normalization layers are also fixed based on $D_{base}$.
This result provides interesting observations as follows\footnote{The same results are observed on CUB200 and miniImageNet, reported in Appendix~\ref{appx:decomp}.}:
\begin{itemize}
    \item No training during novel sessions (M1 in Figure~\ref{fig:cifar100_decomp}) undoubtedly is the best on the base performance, whereas the worst on the novel performance. The weighted performance, which is the main evaluation measurement for FSCIL, can be misleading by the base performance.
    \item Training including an extractor $f$ (M5 in Figure~\ref{fig:cifar100_decomp}) significantly deteriorates the base performance, which is in line with \citet{jie2022alleviating}. Moreover, this training scheme has a similar novel performance to the models with better base performance (M3 and M4 in Figure~\ref{fig:cifar100_decomp}). Therefore, simply updating $f$ is not an appealing strategy.
    \item Training only the current classifier $h^i$ (M2 in Figure~\ref{fig:cifar100_decomp}) is the best strategy for improving the novel performance; however, this training scheme degrades the base performance even with the fixed extractor $f$.
\end{itemize}

We further investigate why the third observation occurs. Under M2, after the base session, every logit on base classes of samples $x$ belonging to the base classes are fixed to $h^0 f(x)$ because $f$ and $h^0$ are frozen. 
However, the predicted probability $p_j$ corresponding to the base class $j$ strictly decreases as novel classes increase as follows:

\begin{equation}
    p_j = \frac{exp(z_j)}{\sum_{k=1}^{K} exp(z_k)} \rightarrow \frac{exp(z_j)}{\sum_{k=1}^{K+nS} exp(z_k)}
\end{equation}

In this situation, we describe the logit distributions of base samples, depicted in Figure~\ref{fig:cifar100_logit}.
Training only $h^i$ leads the logit values of incremental novel classes to exceed the top logit value of base classes, regardless of which class comes in. The logit values for the novel classes are large overall by making the norm of $h^i_k \, (i>0$ and $k \in \{1,\cdots,n\})$ larger than the norms of row vectors of $h^0$. The logit distribution of base samples on other datasets are provided in Appendix~\ref{appx:logit_dist}.

\section{NoNPC: No Training with Normalized Prototype Classifiers}\label{sec:pre_exp}

\begin{figure}[t]
    \centering
    \includegraphics[width=0.8\linewidth]{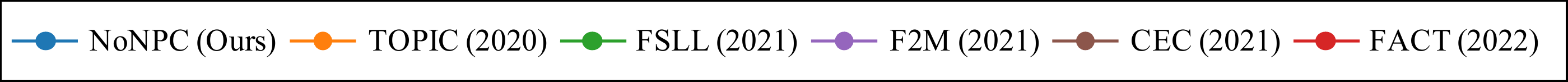}
    
    \begin{subfigure}[h]{0.32\linewidth}
    \includegraphics[width=\linewidth]{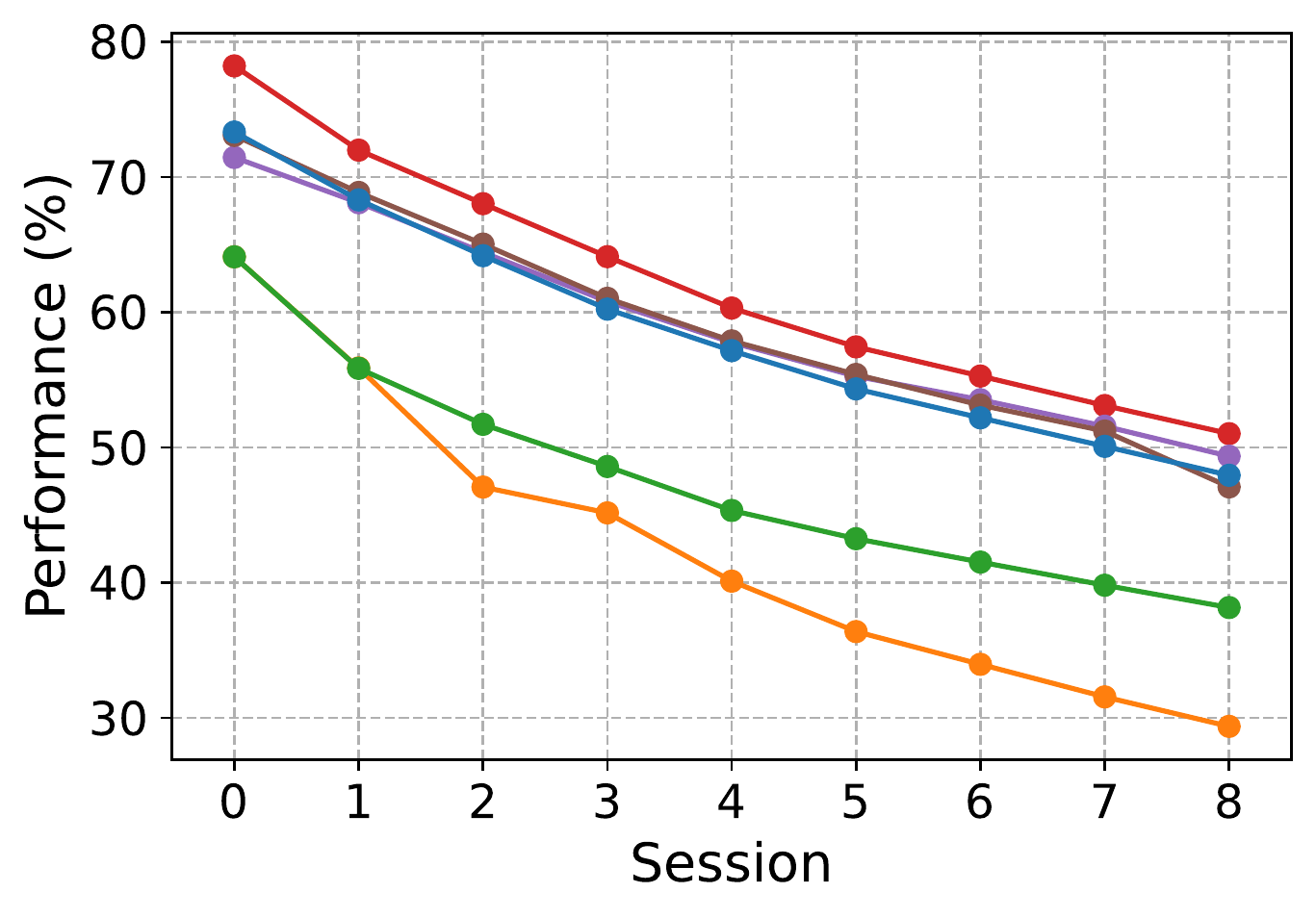}
        \caption{CIFAR100}
    \end{subfigure}
    \hfill
    \begin{subfigure}[h]{0.32\linewidth}
    \includegraphics[width=\linewidth]{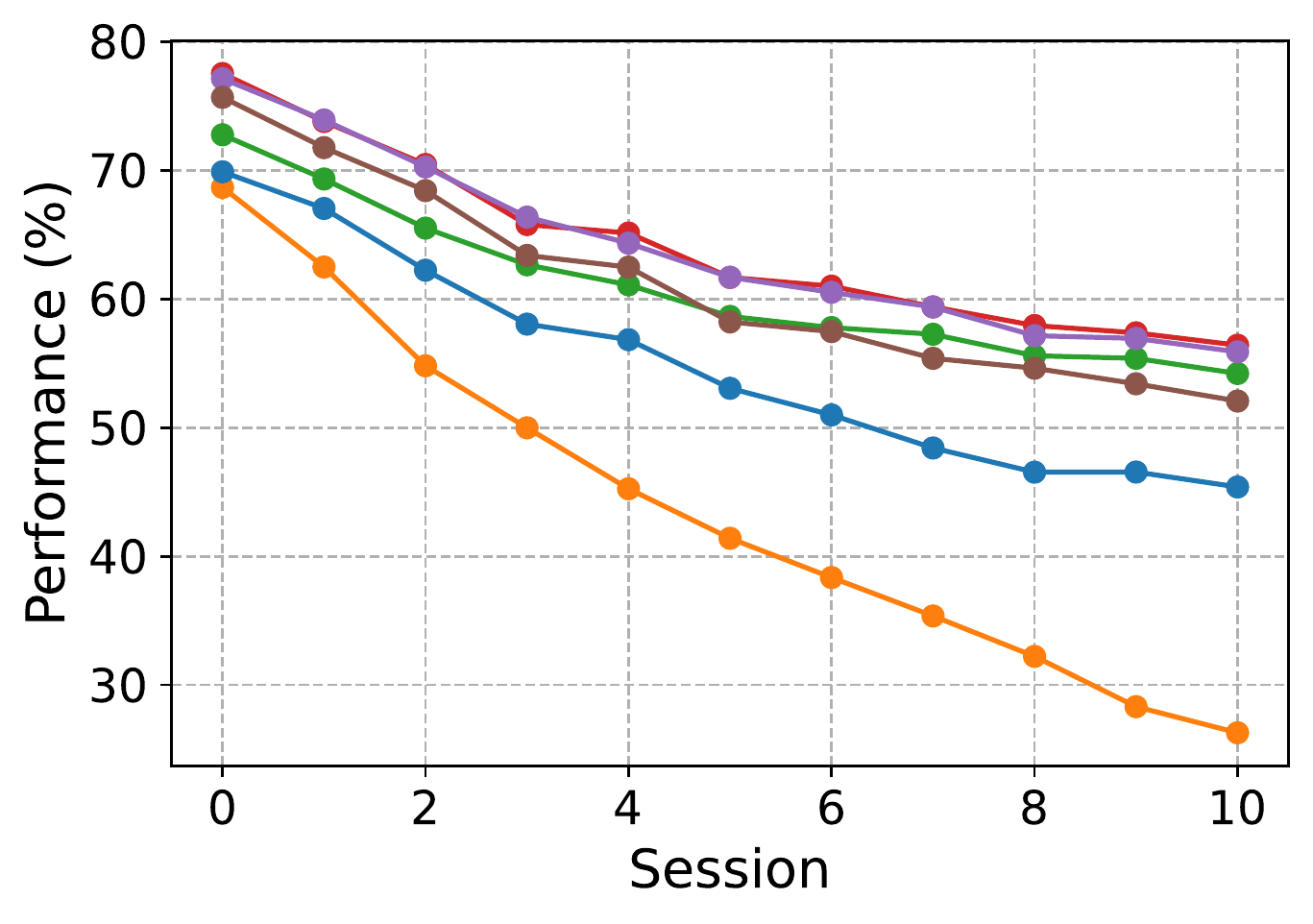}
        \caption{CUB200}
    \end{subfigure}
    \hfill
    \begin{subfigure}[h]{0.32\linewidth}
    \includegraphics[width=\linewidth]{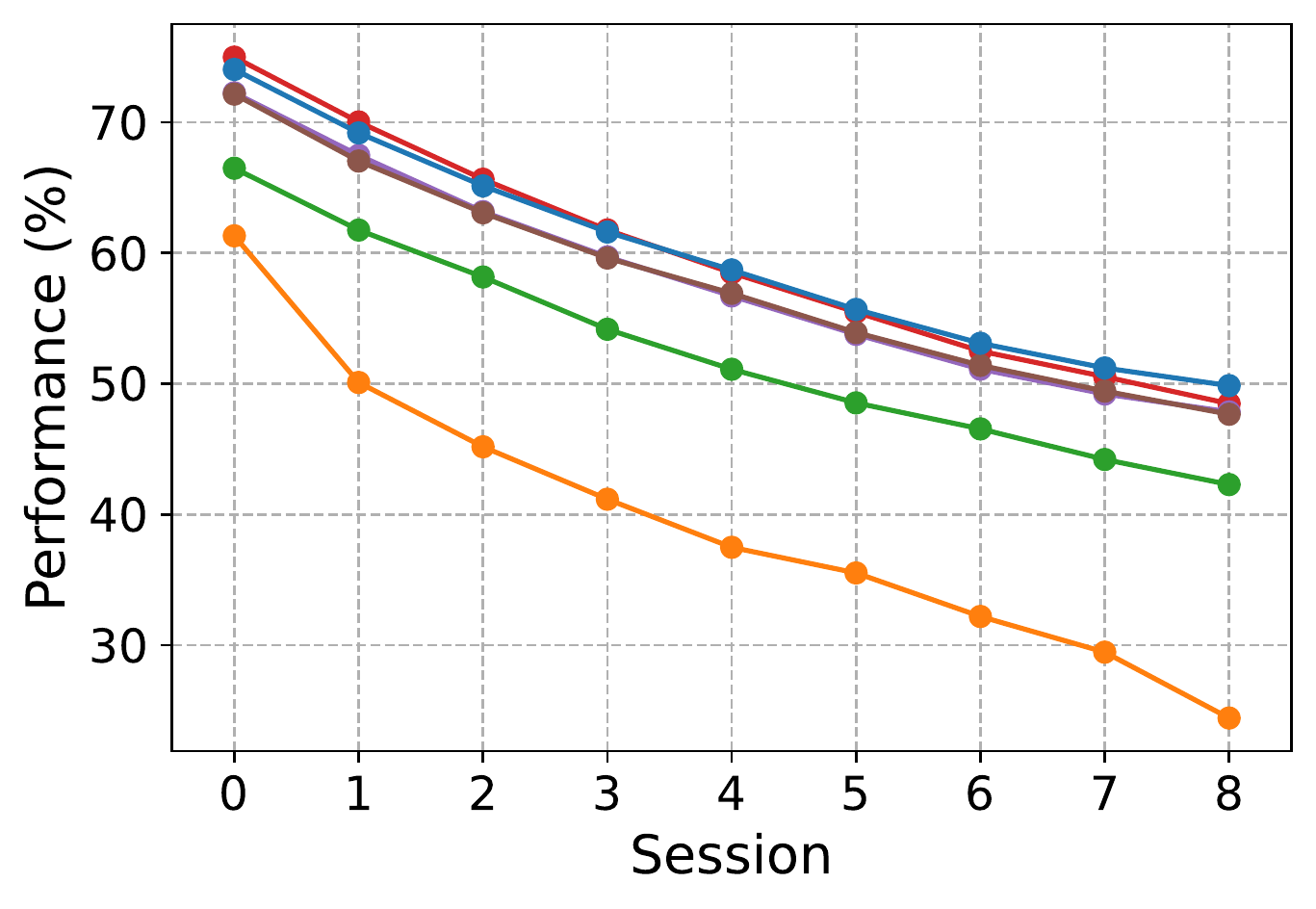}
        \caption{miniImageNet}
    \end{subfigure}
    \caption{Weighted performance comparison. CEC \citep{zhang2021few} and FacT \citep{zhou2022forward} are reproduced and other algorithms are from \citet{shi2021overcoming}. The base and novel performances comparisons are reported in Appendix~\ref{appx:existing}.}
    \label{fig:weighted_comp}
\end{figure}

We propose a simple yet powerful method called \textbf{NoNPC}, which means \textbf{No} training with \textbf{N}ormalized \textbf{P}rototype \textbf{C}lassifiers, inspired by observations in Section~\ref{sec:decompose}: combining (1) M1 to maintain the base performance and (2) M2 to improve the novel performance.
Note that M1 does not work on the novel classes at all, while M2 deteriorates the base performance.
To solve this conundrum, we use non-parametric prototype classifiers without training.
The prototype $c^i_j$ for the novel class $j$ in the session $i\,(i>0)$ are defined as:

\begin{equation}
    c^i_j = \frac{1}{k} \sum_{x \text{ belongs to class } j}{f(x)}
\end{equation}

However, $c^i_j$ is unstable because $k$ is significantly small (e.g., 1 or 5). To mitigate this issue, we transform prototype $c^i_j$ through L2-normalization, following \citet{wang2019simpleshot}, in which it is shown that feature transformation improves performance.
Finally, we use the normalized prototype as a classifier, i.e., $h^i_j = \frac{c^i_j}{\|c^i_j\|_2}$.
Furthermore, to match the characteristics between $h^0$ and $h^i\,(i>0)$, we replace the base classifier $h^0$ optimized via stochastic gradient descent with a normalized prototype classifier in the same way after the base session. The ablation studies related to prototype normalization are described in Appendix~\ref{appx:proto}.

Figure~\ref{fig:weighted_comp} describes the weighted performance comparison on CIFAR100, CUB200, and miniImageNet.
It is observed that our simple method has the best performance on miniImageNet and comparable performance to recent algorithms on CIFAR100.
However, on CUB dataset, our algorithm does not achieve the desired performance.

\subsection{Label Smoothing on Fine-grained Dataset}

\begin{figure}[t]
    \centering
    \begin{subfigure}[h]{0.29\linewidth}
    \includegraphics[width=\linewidth]{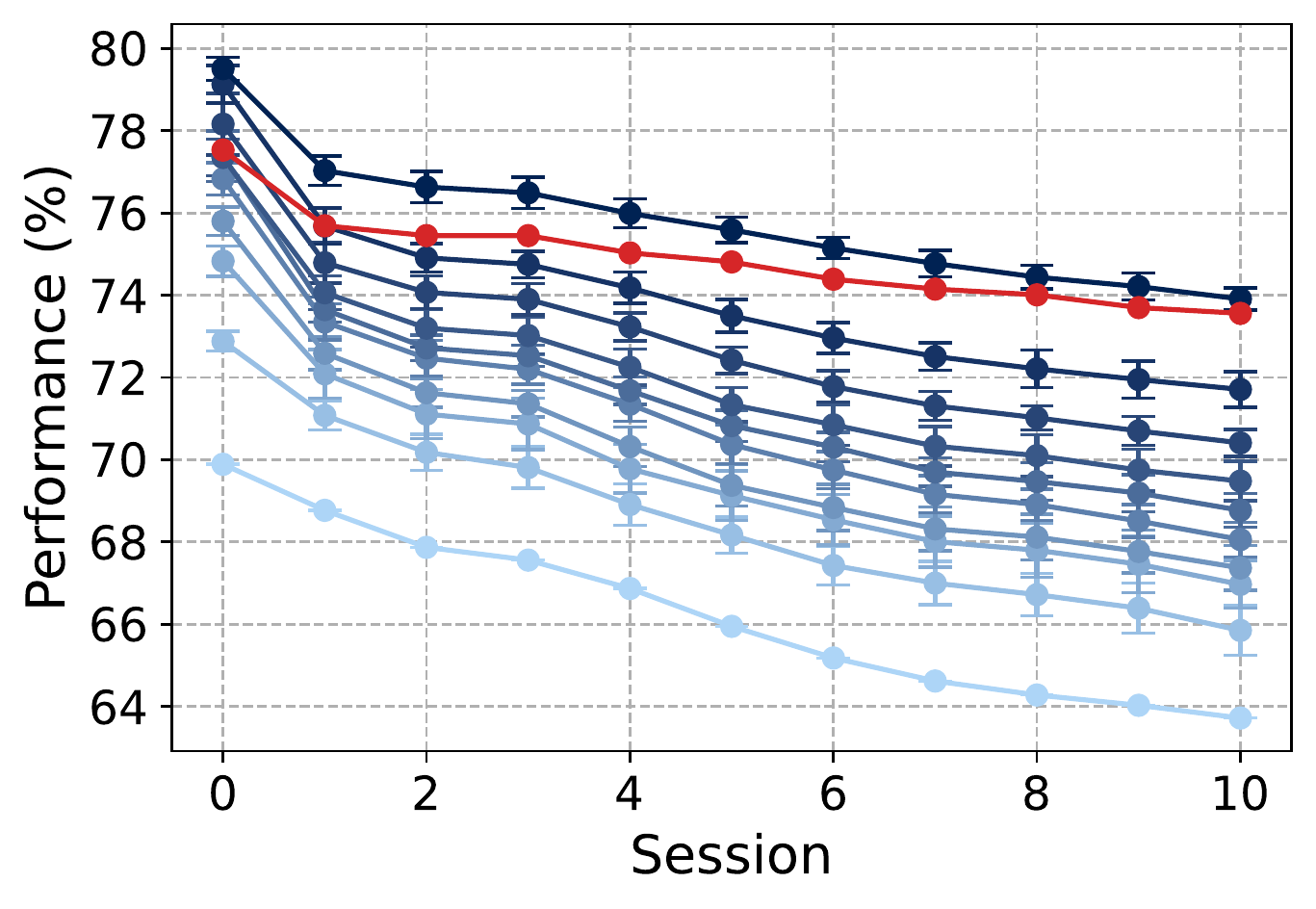}
        \caption{Base}
    \end{subfigure}
    \hfill
    \begin{subfigure}[h]{0.29\linewidth}
    \includegraphics[width=\linewidth]{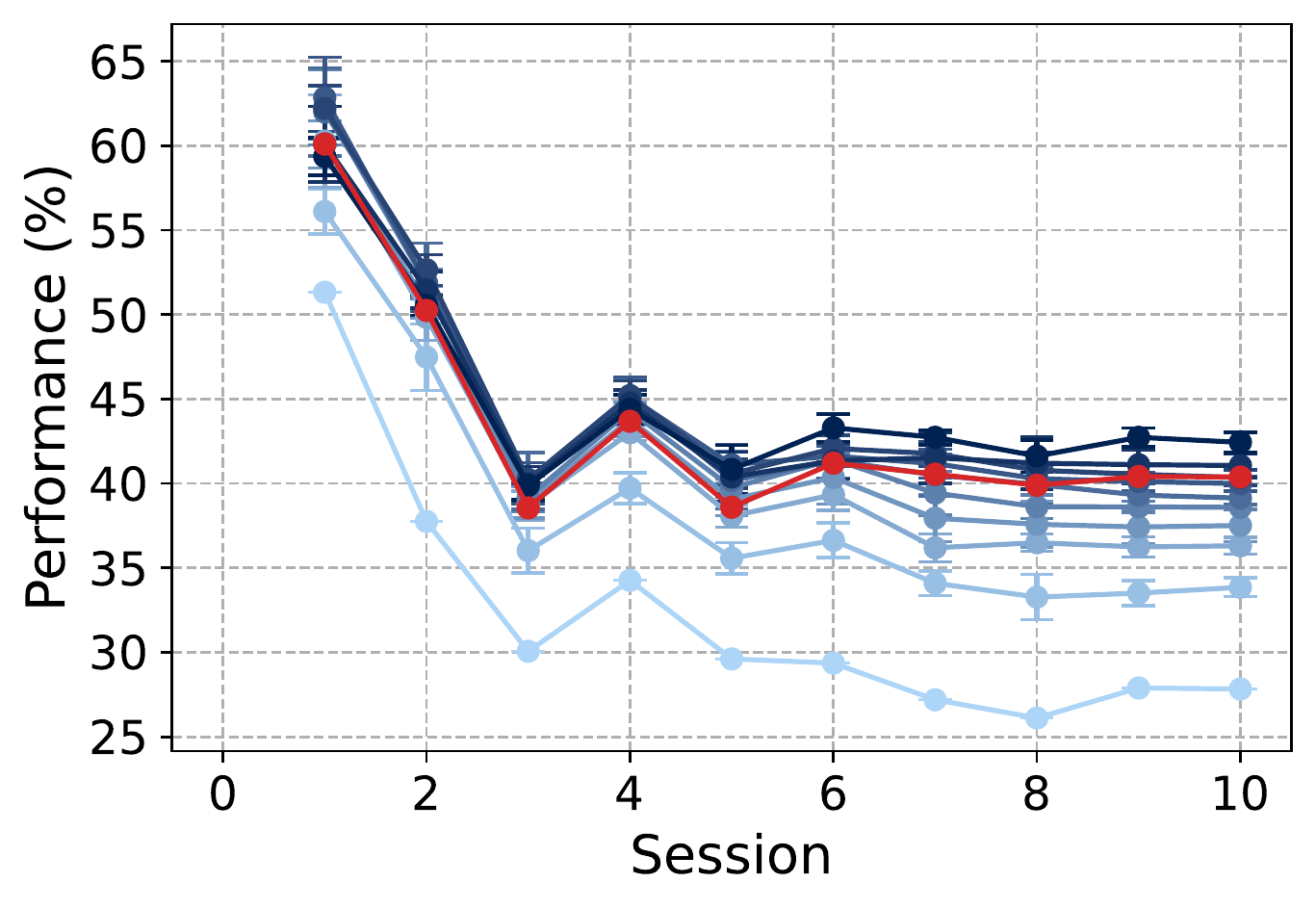}
        \caption{Novel}
    \end{subfigure}
    \hfill
    \begin{subfigure}[h]{0.29\linewidth}
    \includegraphics[width=\linewidth]{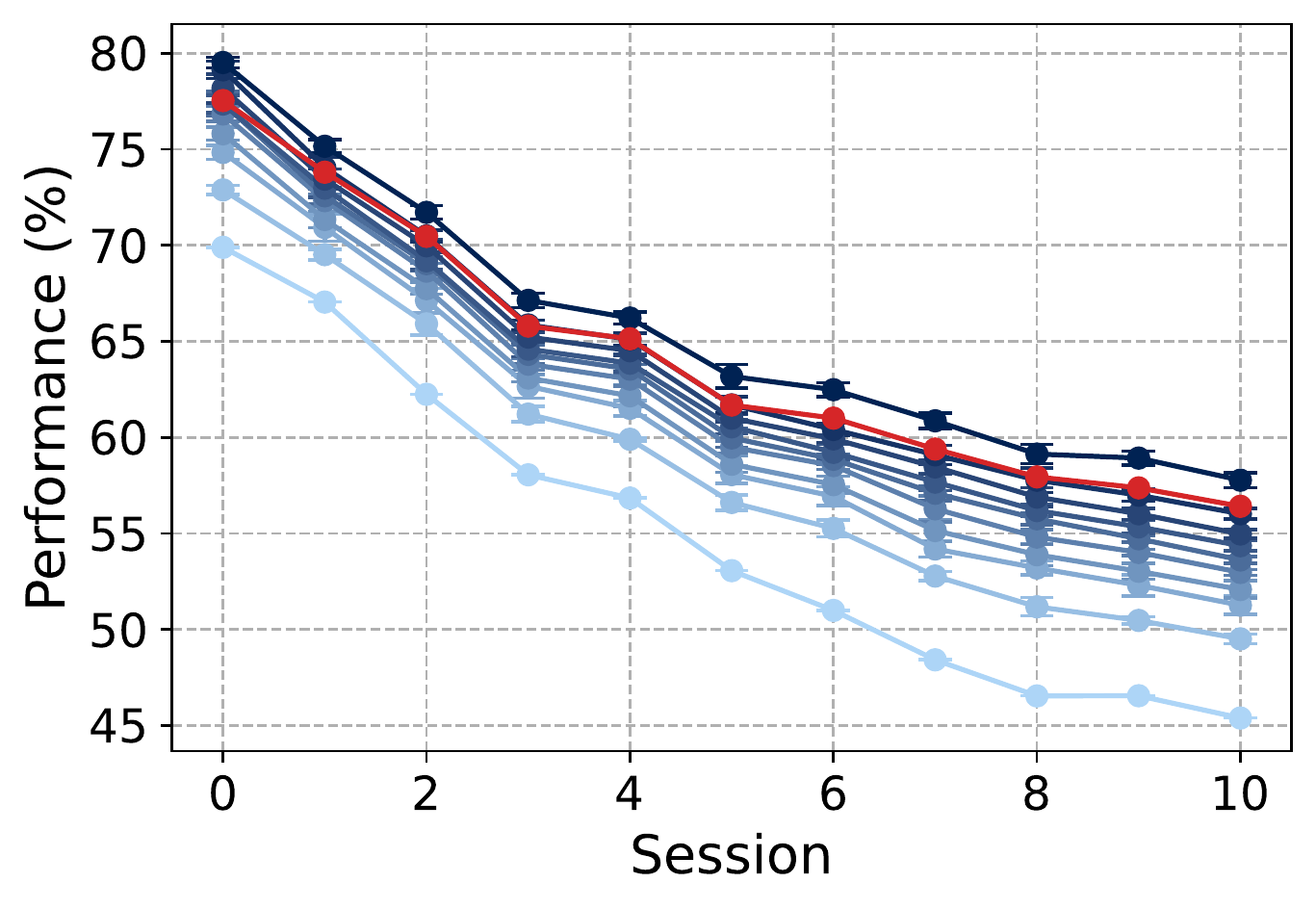}
        \caption{Weighted}
    \end{subfigure}
    \hfill
    \begin{subfigure}[h]{0.10\linewidth}
    \includegraphics[width=\linewidth]{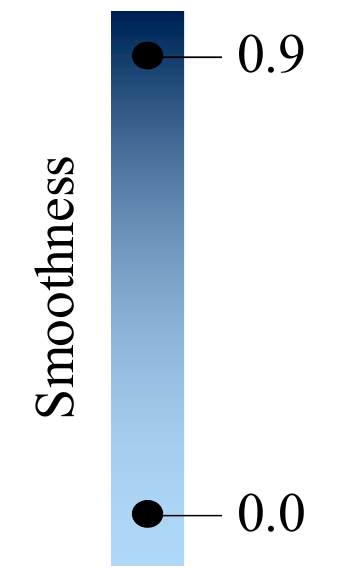}
    \end{subfigure}
    \caption{Base, novel, and weighted performances according to the degree of label smoothing on CUB200. The most light and dark blue colors indicate smoothness of 0.0 and 0.9, respectively. The value of 0.0 means label smoothing is not used and the value increases by 0.1 from 0.0 to 0.9. The red line indicates the performance of FACT\,\citep{zhou2022forward}.}
    \label{fig:cub_ls}
\end{figure}

CUB is one of the fine-grained datasets, which implies that the distance in the representation spaces between different classes can be closer.
However, for both base and novel performances, it seems crucial that representations within the same class are clustered tightly and representations between different classes maintain the distance, before the novel classes increase.
Therefore, we believe that label smoothing is appropriate for fine-grained datasets during the base session. This is because label smoothing helps tight clustering, making equidistance between different classes\,\citep{muller2019does}.


Figure~\ref{fig:cub_ls} describes performances according to the degree of label smoothing on CUB200. Surprisingly, both base and novel performances consistently increase until the smoothness reaches 0.9.
In addition, this outperforms FACT\,\citep{zhou2022forward}. The results on other datasets are reported in Appendix~\ref{appx:ls}.
\section{Conclusion}\label{sec:conclusion}
In this paper, we decomposed the entire network into four partial parameters to investigate the relationship between the update parts and the two performances.
Based on observations, we proposed a simple yet powerful method called NoNPC, which does not learn a model for the novel sessions and only makes inferences with normalized prototype classifiers. This straightforward method achieved comparable performance with the state-of-the-art algorithms.
Furthermore, we showed that the label smoothing technique on the fine-grained dataset boosts the performance.
We hope that our NoNPC will be used as a baseline in future studies on FSCIL.




\newpage
\appendix
\section{Implementation Details}\label{appx:detail}
\subsection{Dataset Details}
Following the benchmark setting for FSCIL\,\citep{tao2020few}, we evaluate the base, novel, and weighted performances on CIFAR100\,\citep{krizhevsky2009learning}, CUB200-2011\,\citep{wah2011caltech}, and miniImageNet\,\citep{russakovsky2015imagenet}.
The number of novel classes is the product of the number of incremental novel classes and novel sessions.
\vspace*{-4pt}
\begin{table}[h]
    \centering
    \begin{tabular}{c|ccc}
    \toprule
    Dataset      & \# of Base classes & \# of incremental novel classes & \# of novel sessions \\
    \midrule
    CIFAR100     & 60  & 5  & 8  \\
    CUB200       & 100 & 10 & 10 \\
    miniImageNet & 60  & 5  & 8  \\
    \bottomrule
    \end{tabular}
    \caption{Class incremental setup.}
    \label{tab:dataset_detail}
    \vspace*{-15pt}
\end{table}

\subsection{Training Details}
Our algorithm requires the base training only.
\vspace*{-4pt}
\begin{table}[h]
    \centering
    \begin{tabular}{ll}
    \toprule
    \multicolumn{2}{c}{Configuration} \\
    \midrule
    Models        & ResNet20 (for CIFAR100) and ResNet18 (for CUB200 and miniImageNet) \\
    Epochs        & 200 \\
    Optimizer     & SGD with momentum 0.9 (nesterov=True) \\
    Batch size    & 256 \\
    Learning rate & 0.1 with milestone scheduler at 120 and 160 epochs (gamma: 0.1) \\
    Weight decay  & 5e-4 (for CIFAR100 and miniImageNet) and 5e-5 (for CUB200) \\
    \bottomrule
    \end{tabular}
    \caption{Training setup.}
    \label{tab:train_detail}
    \vspace*{-15pt}
\end{table}

\section{Performance According to Decomposition}\label{appx:decomp}

\begin{figure}[h]
    \centering
    \includegraphics[width=0.45\linewidth]{figure/legend1.pdf}
    
    \begin{subfigure}[h]{0.32\linewidth}
    \includegraphics[width=\linewidth]{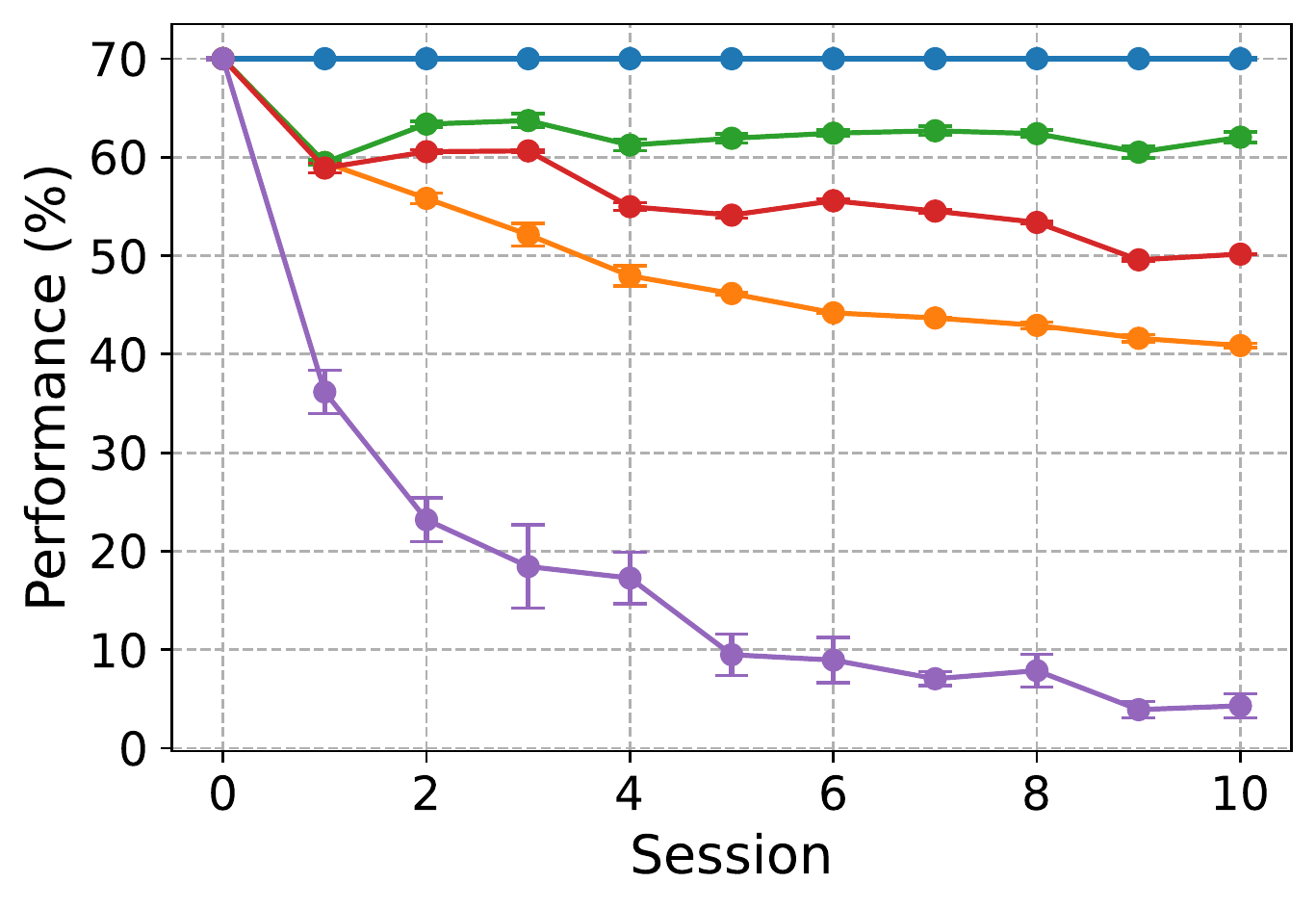}
        \caption{Base}
    \end{subfigure}
    \hfill
    \begin{subfigure}[h]{0.32\linewidth}
    \includegraphics[width=\linewidth]{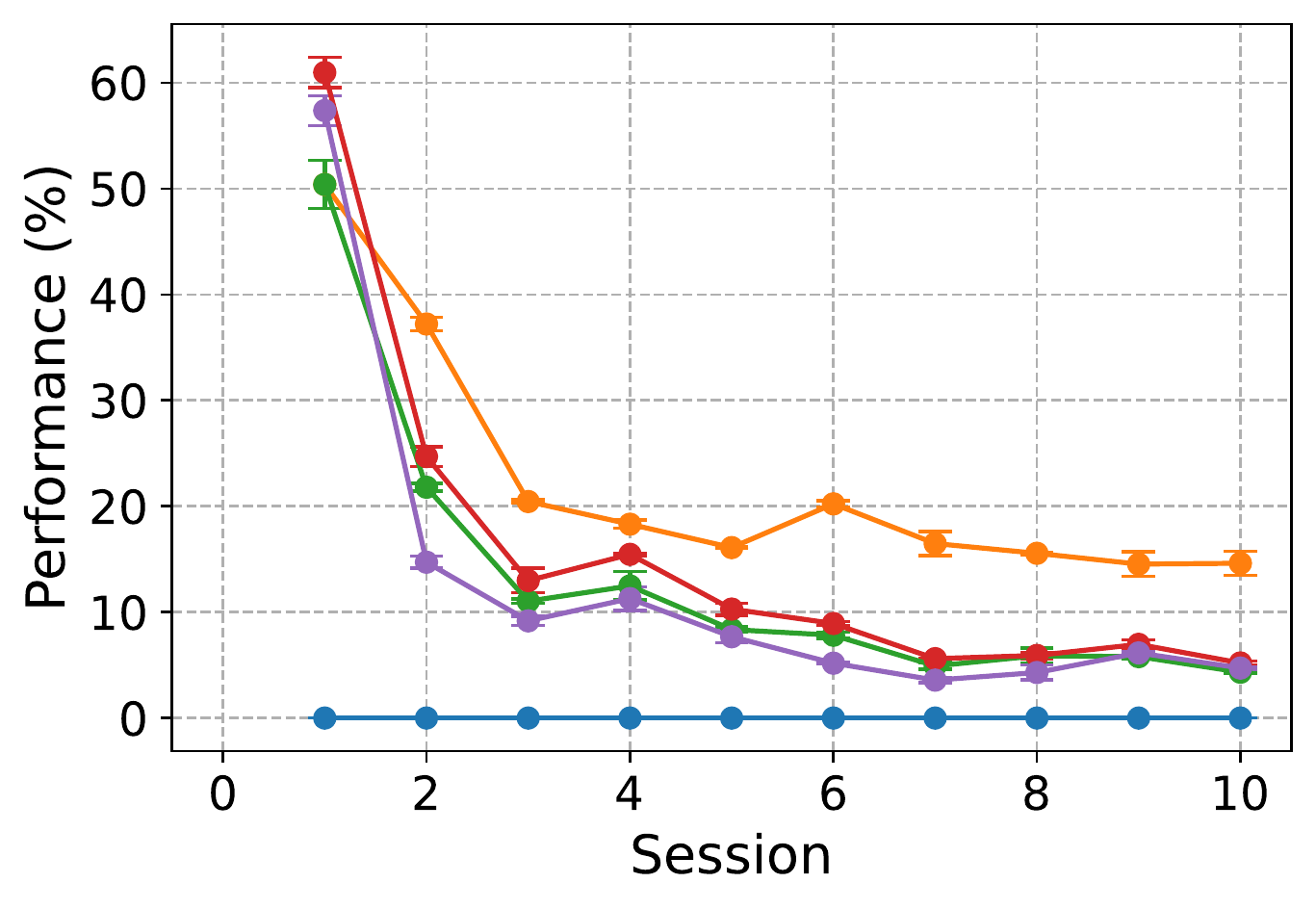}
        \caption{Novel}
    \end{subfigure}
    \hfill
    \begin{subfigure}[h]{0.32\linewidth}
    \includegraphics[width=\linewidth]{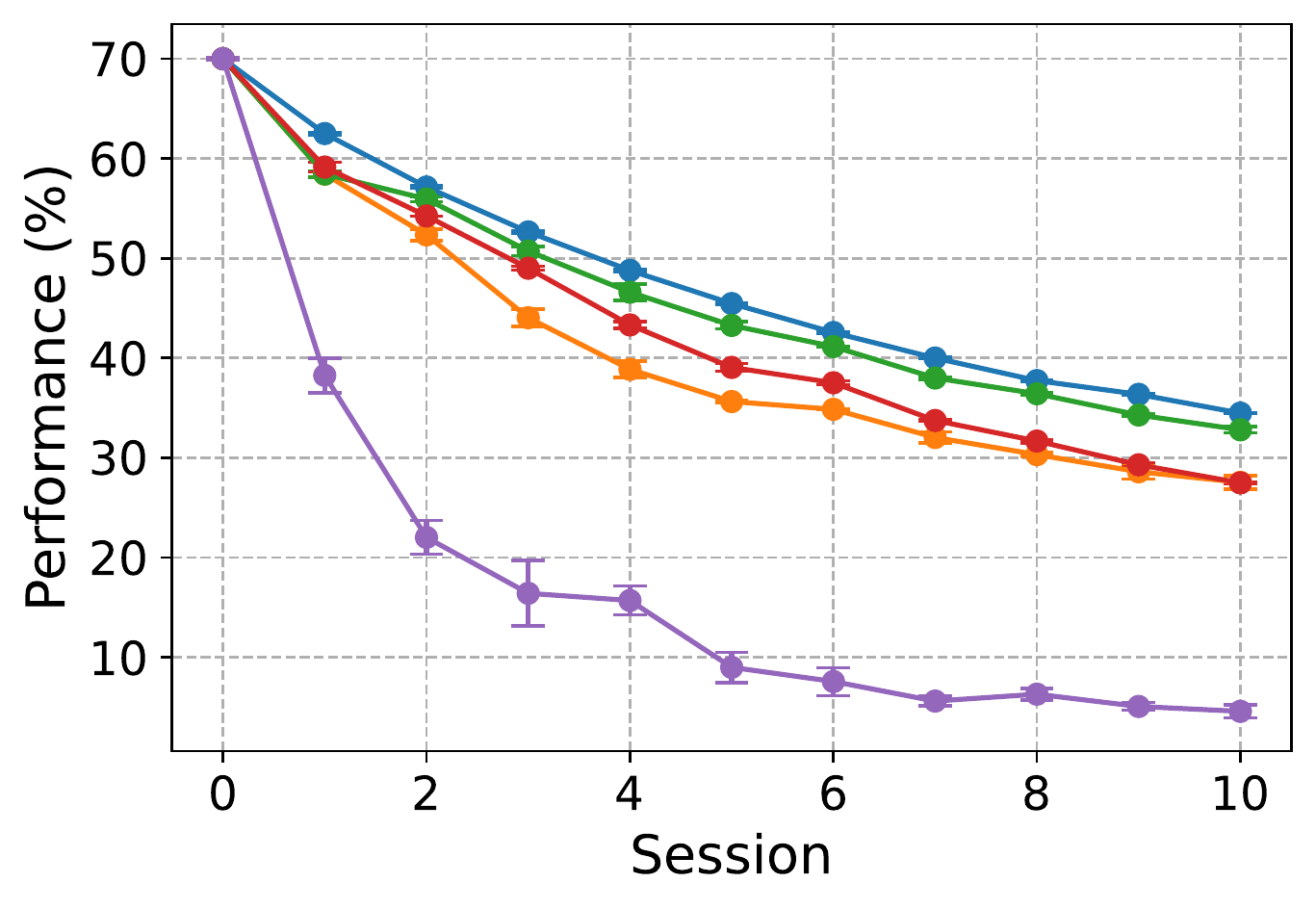}
        \caption{Weighted}
    \end{subfigure}
    \caption{Performances according to the update parts on CUB200.}
    \label{fig:cub200_decomp}
\end{figure}

\begin{figure}[h]
    \centering
    \includegraphics[width=0.45\linewidth]{figure/legend1.pdf}
    
    \begin{subfigure}[h]{0.32\linewidth}
    \includegraphics[width=\linewidth]{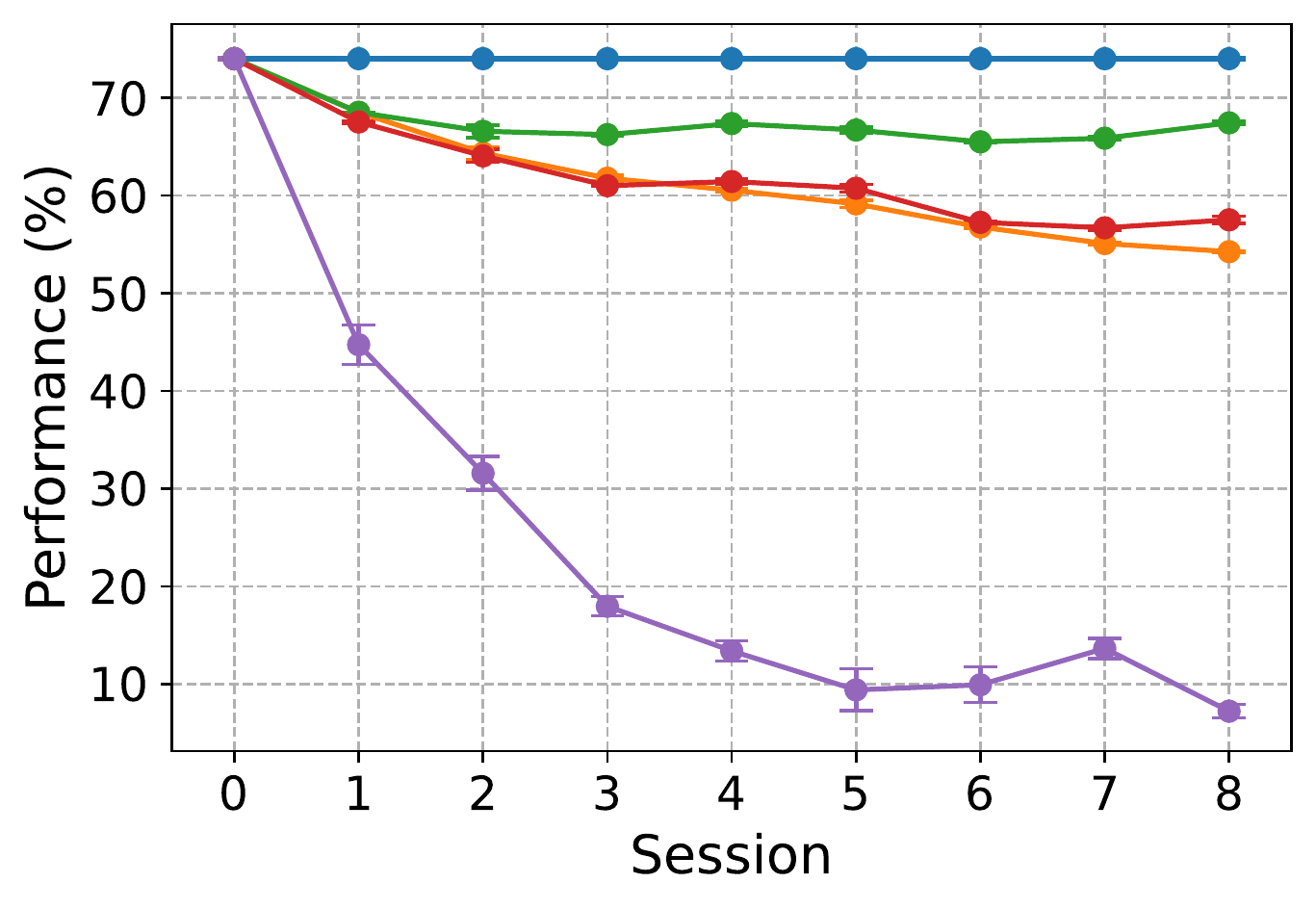}
        \caption{Base}
    \end{subfigure}
    \hfill
    \begin{subfigure}[h]{0.32\linewidth}
    \includegraphics[width=\linewidth]{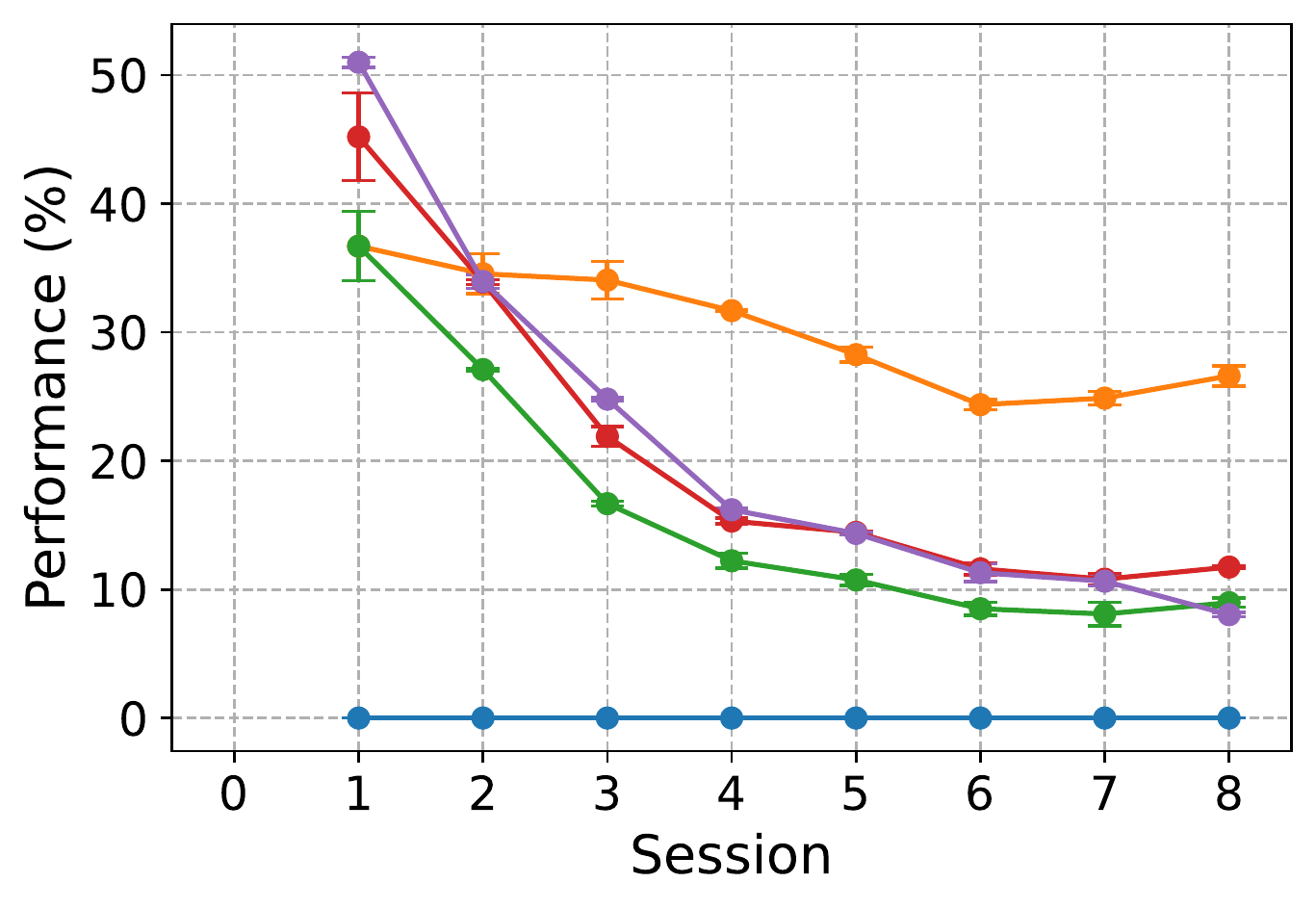}
        \caption{Novel}
    \end{subfigure}
    \hfill
    \begin{subfigure}[h]{0.32\linewidth}
    \includegraphics[width=\linewidth]{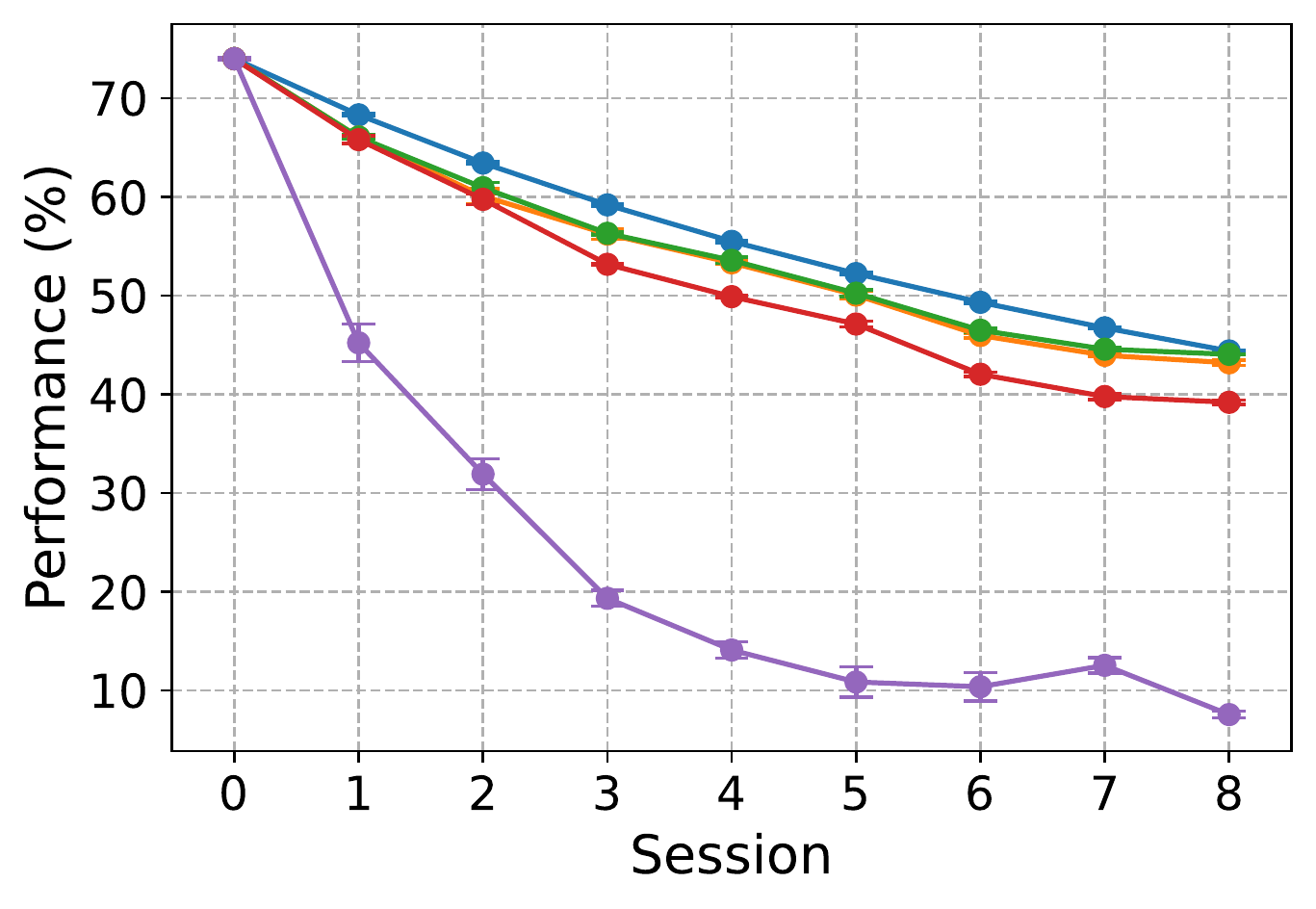}
        \caption{Weighted}
    \end{subfigure}
    \caption{Performances according to the update parts on miniImageNet.}
    \label{fig:miniIN_decomp}
\end{figure}

Figure \ref{fig:cub200_decomp} and \ref{fig:miniIN_decomp} describe the base, novel, and weighted performances according to the decomposition on CUB200 and miniImageNet, respectively. Models are described in Table~\ref{tab:notation}.

\section{Logit Distributions}\label{appx:logit_dist}

\begin{figure}[h]
    \centering
    \begin{subfigure}[h]{0.175\linewidth}
    \includegraphics[width=\linewidth]{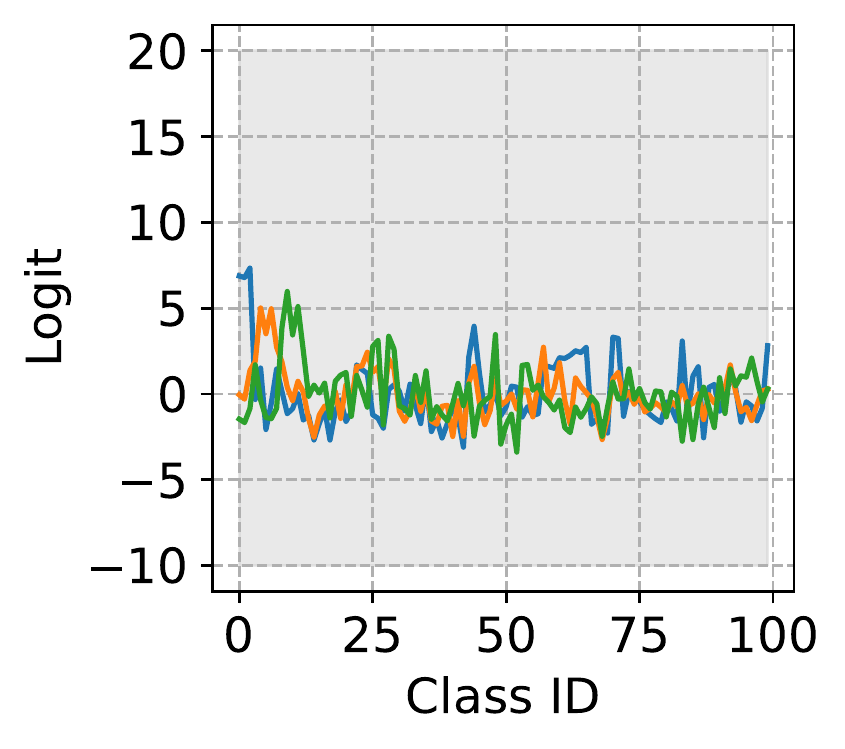}
        \caption{Session 0}
    \end{subfigure}
    \begin{subfigure}[h]{0.155\linewidth}
    \includegraphics[width=\linewidth]{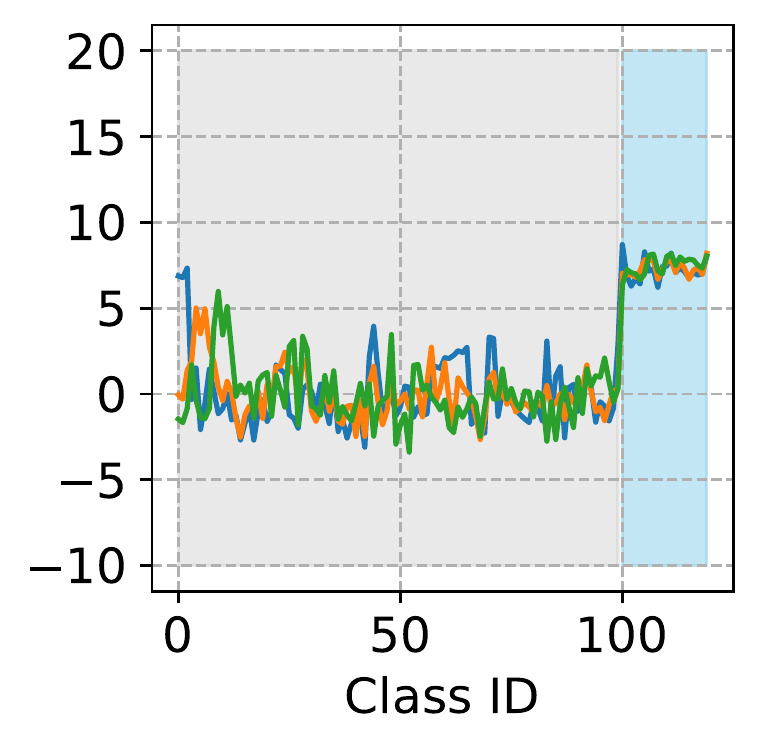}
        \caption{Session 2}
    \end{subfigure}
    \begin{subfigure}[h]{0.155\linewidth}
    \includegraphics[width=\linewidth]{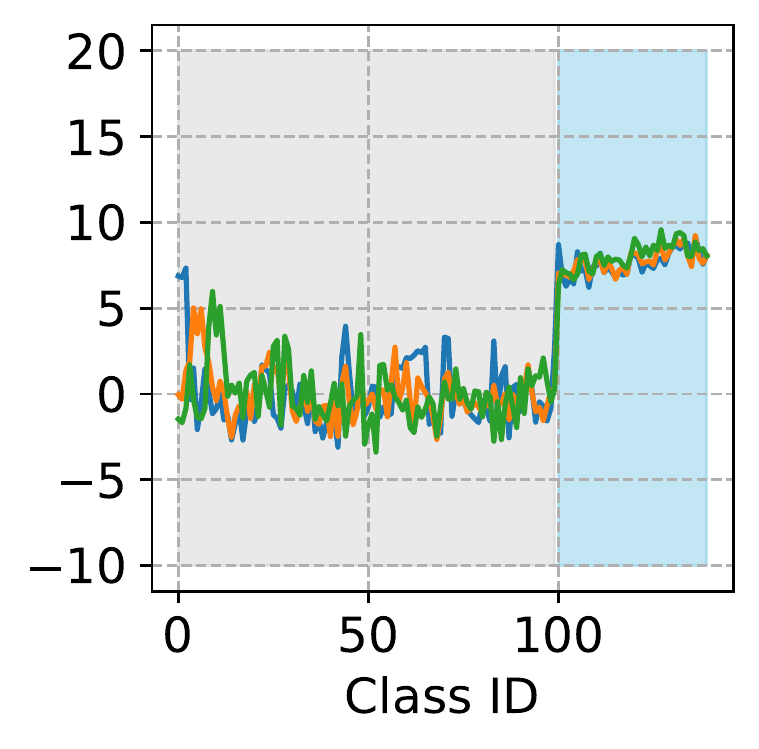}
        \caption{Session 4}
    \end{subfigure}
    \begin{subfigure}[h]{0.155\linewidth}
    \includegraphics[width=\linewidth]{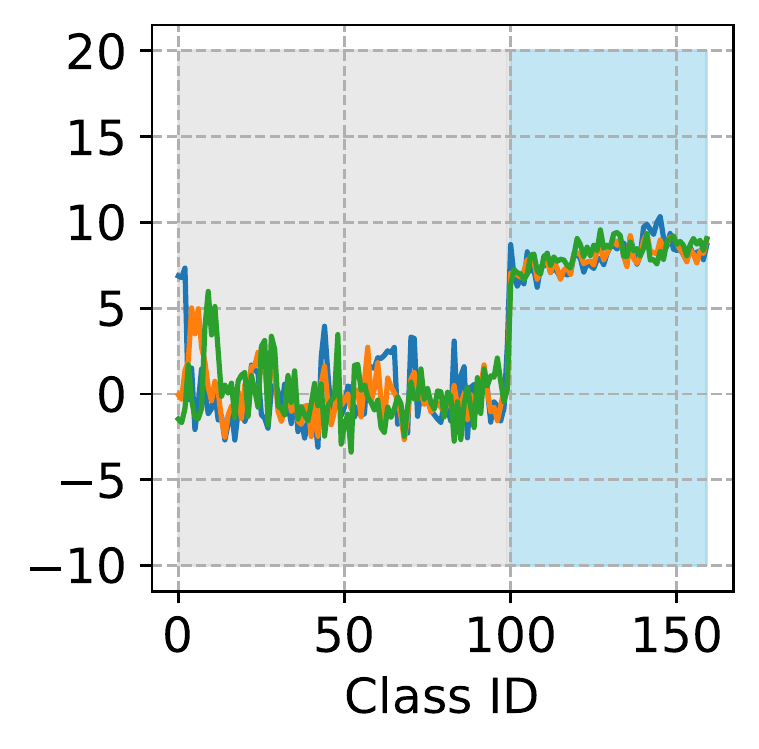}
        \caption{Session 6}
    \end{subfigure}
    \begin{subfigure}[h]{0.155\linewidth}
    \includegraphics[width=\linewidth]{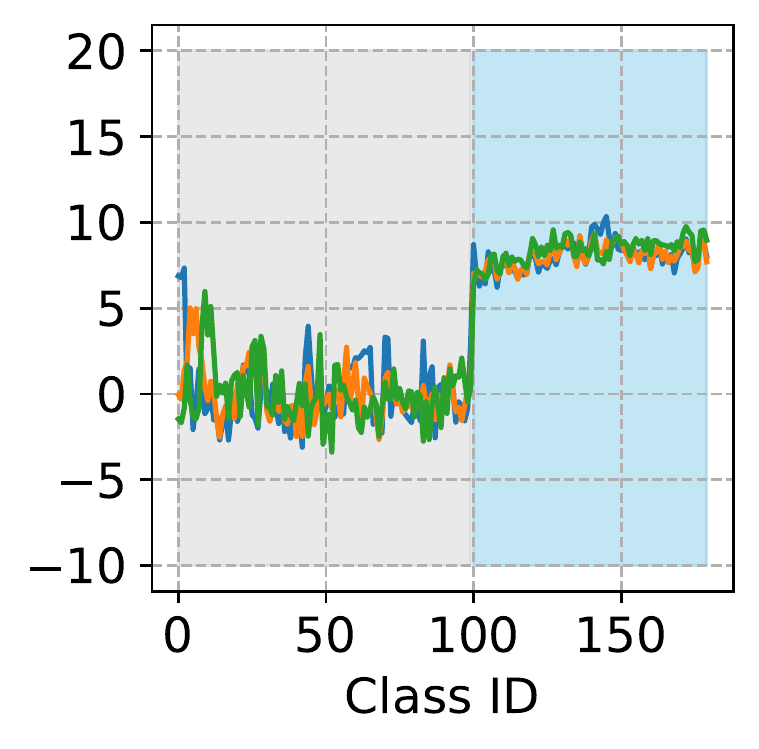}
        \caption{Session 8}
    \end{subfigure}
    \begin{subfigure}[h]{0.155\linewidth}
    \includegraphics[width=\linewidth]{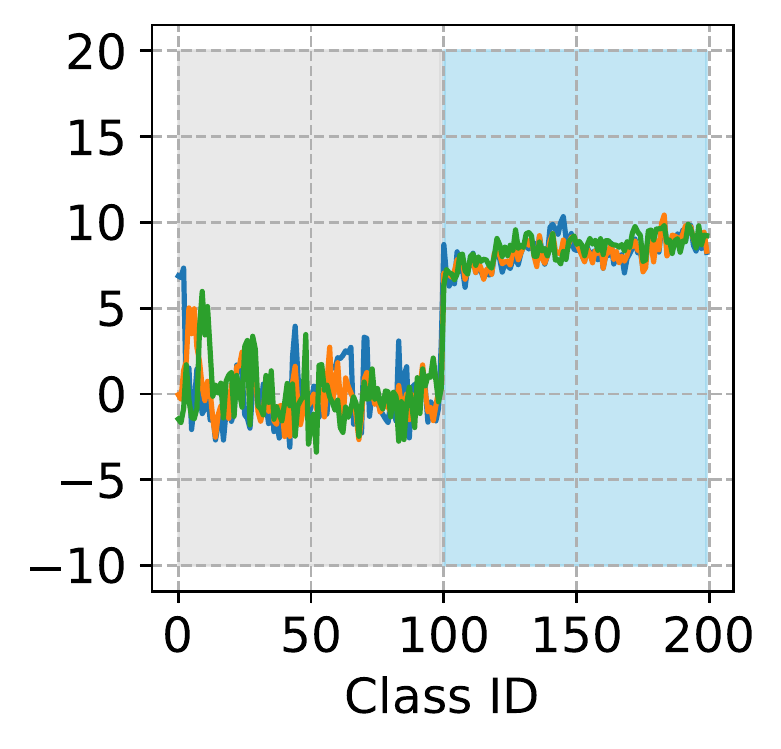}
        \caption{Session 10}
    \end{subfigure}
    %
    \caption{Logit distributions of base samples according to the session on CUB200 when training only $h^i$. Gray and blue backgrounds indicate base and novel classes, respectively. Each line is the average of logits of samples belonging to class 0, 1, and 2.}
    \label{fig:cub200_logit}
\end{figure}

\begin{figure}[h]
    \centering
    \begin{subfigure}[h]{0.195\linewidth}
    \includegraphics[width=\linewidth]{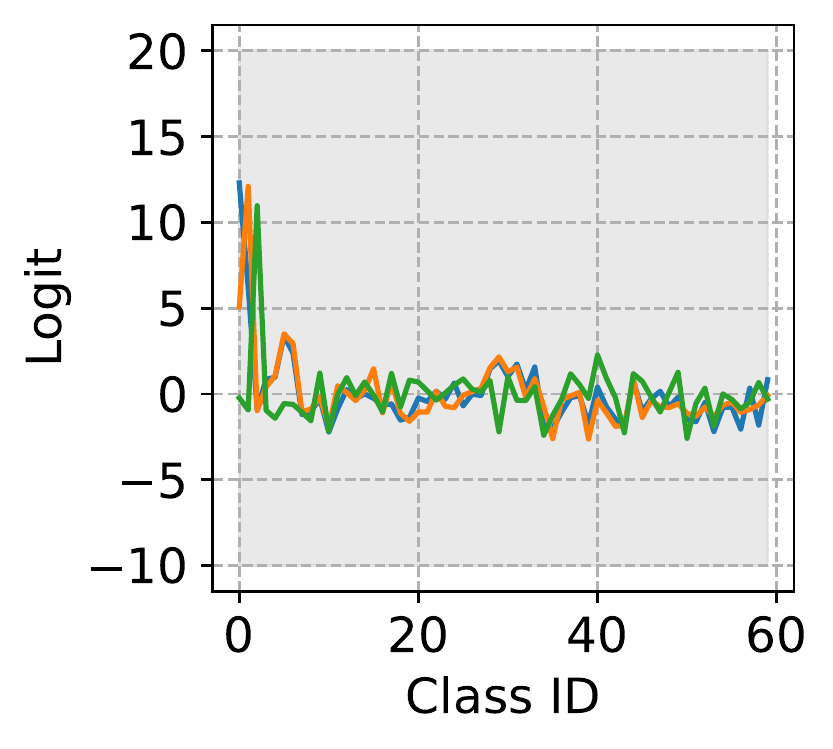}
        \caption{Session 0}
    \end{subfigure}
    \begin{subfigure}[h]{0.18\linewidth}
    \includegraphics[width=\linewidth]{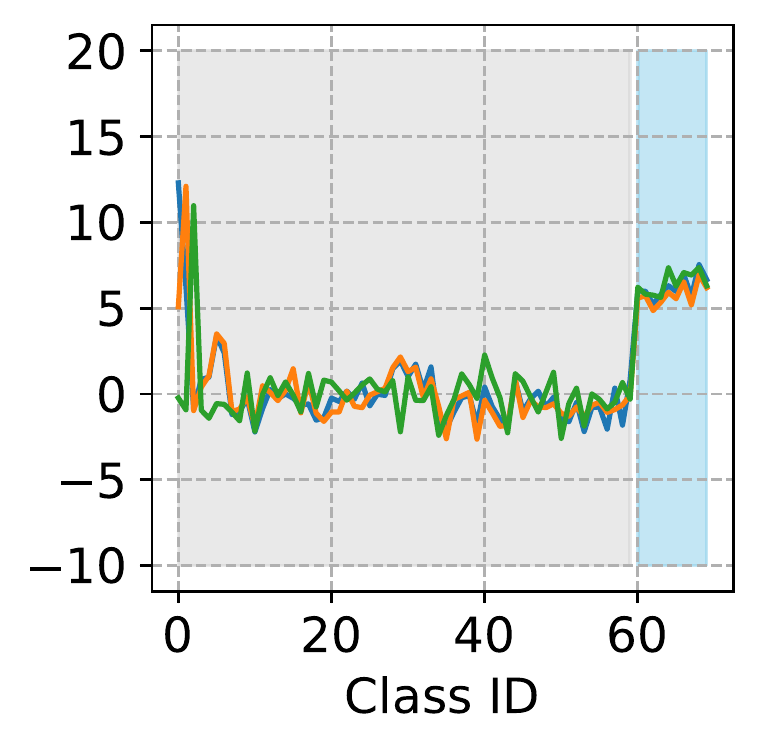}
        \caption{Session 2}
    \end{subfigure}
    \begin{subfigure}[h]{0.18\linewidth}
    \includegraphics[width=\linewidth]{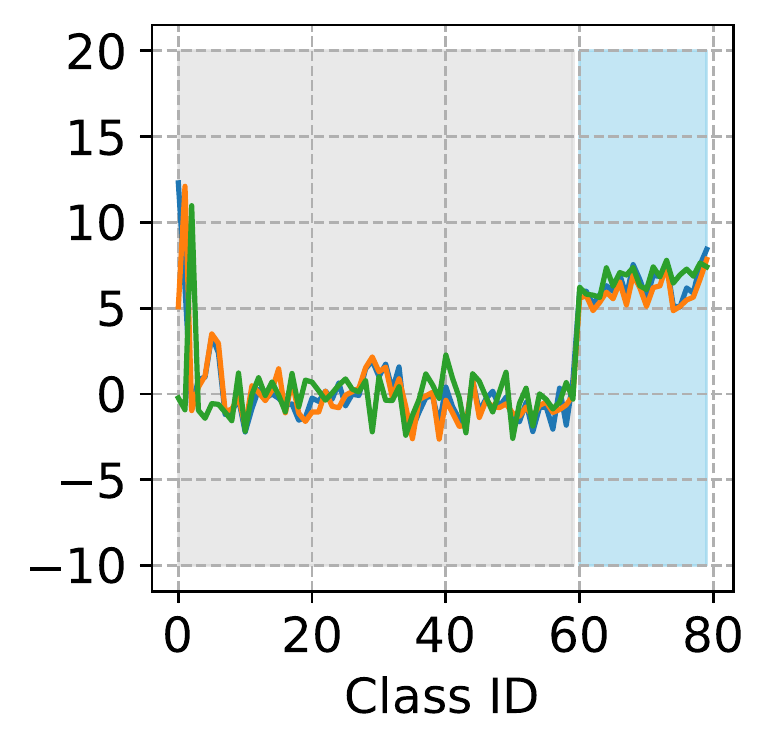}
        \caption{Session 4}
    \end{subfigure}
    \begin{subfigure}[h]{0.18\linewidth}
    \includegraphics[width=\linewidth]{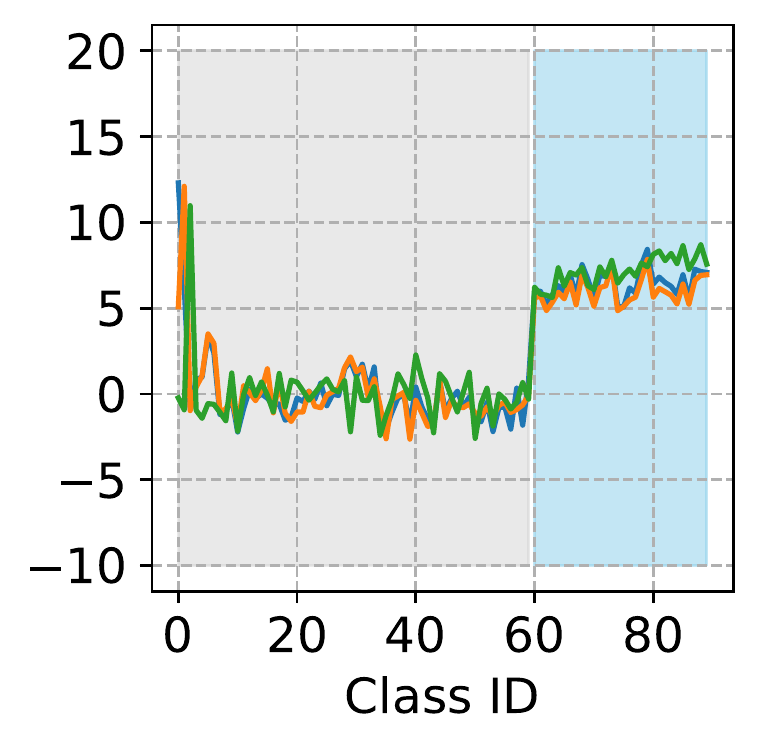}
        \caption{Session 6}
    \end{subfigure}
    \begin{subfigure}[h]{0.18\linewidth}
    \includegraphics[width=\linewidth]{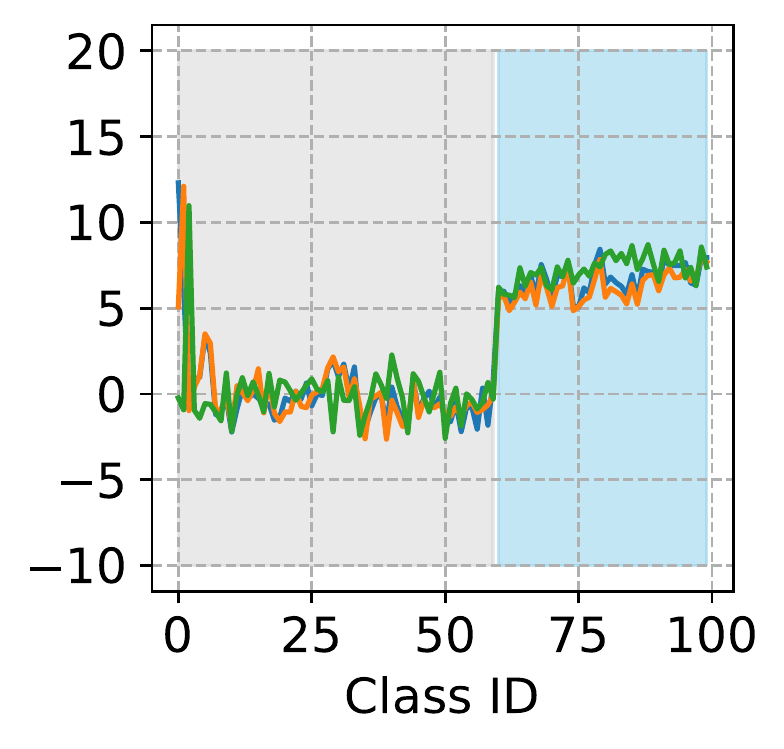}
        \caption{Session 8}
    \end{subfigure}
    %
    \caption{Logit distributions of base samples according to the session on miniImageNet when training only $h^i$. Gray and blue backgrounds indicate base and novel classes, respectively. Each line is the average of logits of samples belonging to class 0, 1, and 2.}
    \label{fig:mini_logit}
\end{figure}

Figure \ref{fig:cub200_logit} and \ref{fig:mini_logit} describe the logit distributions of base samples when training only the current classifier $h^i$ on CUB200 and miniImageNet, respectively.

\clearpage
\section{Base, Novel, and Weighted Performances Comparison}\label{appx:existing}

\begin{figure}[h]
    \centering
    \begin{subfigure}[h]{0.32\linewidth}
    \includegraphics[width=\linewidth]{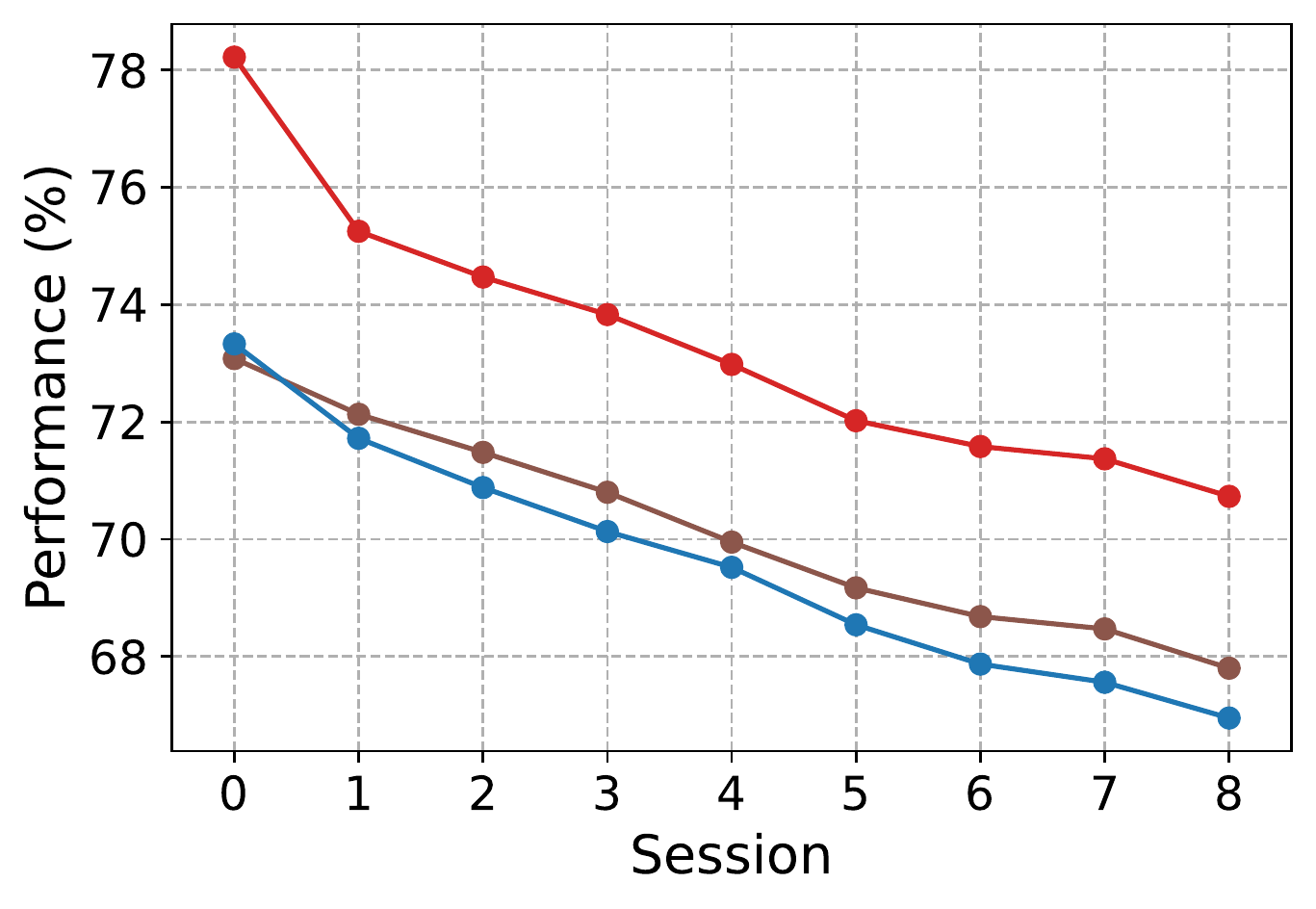}
        \caption{Base}
    \end{subfigure}
    \hfill
    \begin{subfigure}[h]{0.32\linewidth}
    \includegraphics[width=\linewidth]{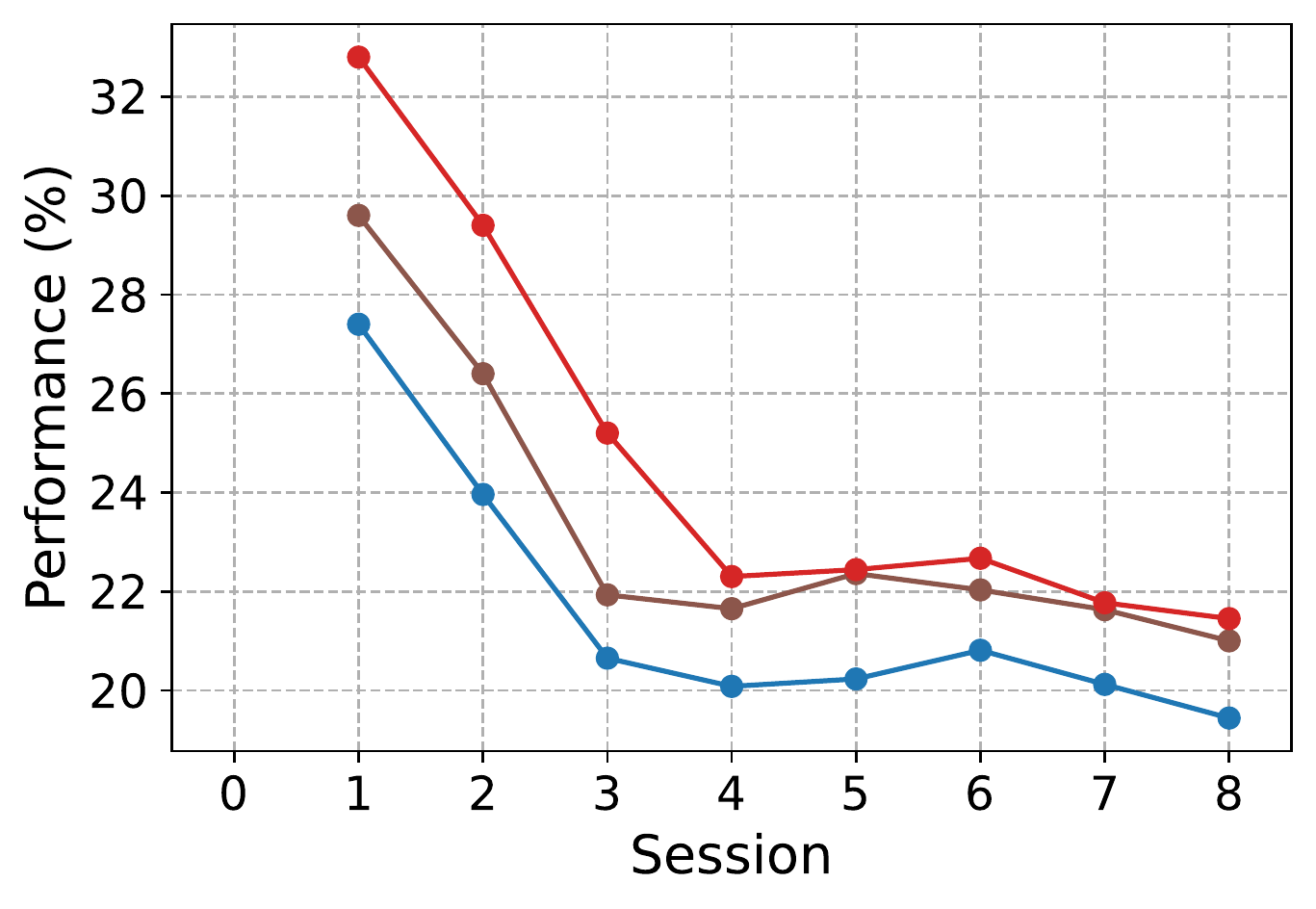}
        \caption{Novel}
    \end{subfigure}
    \hfill
    \begin{subfigure}[h]{0.32\linewidth}
    \includegraphics[width=\linewidth]{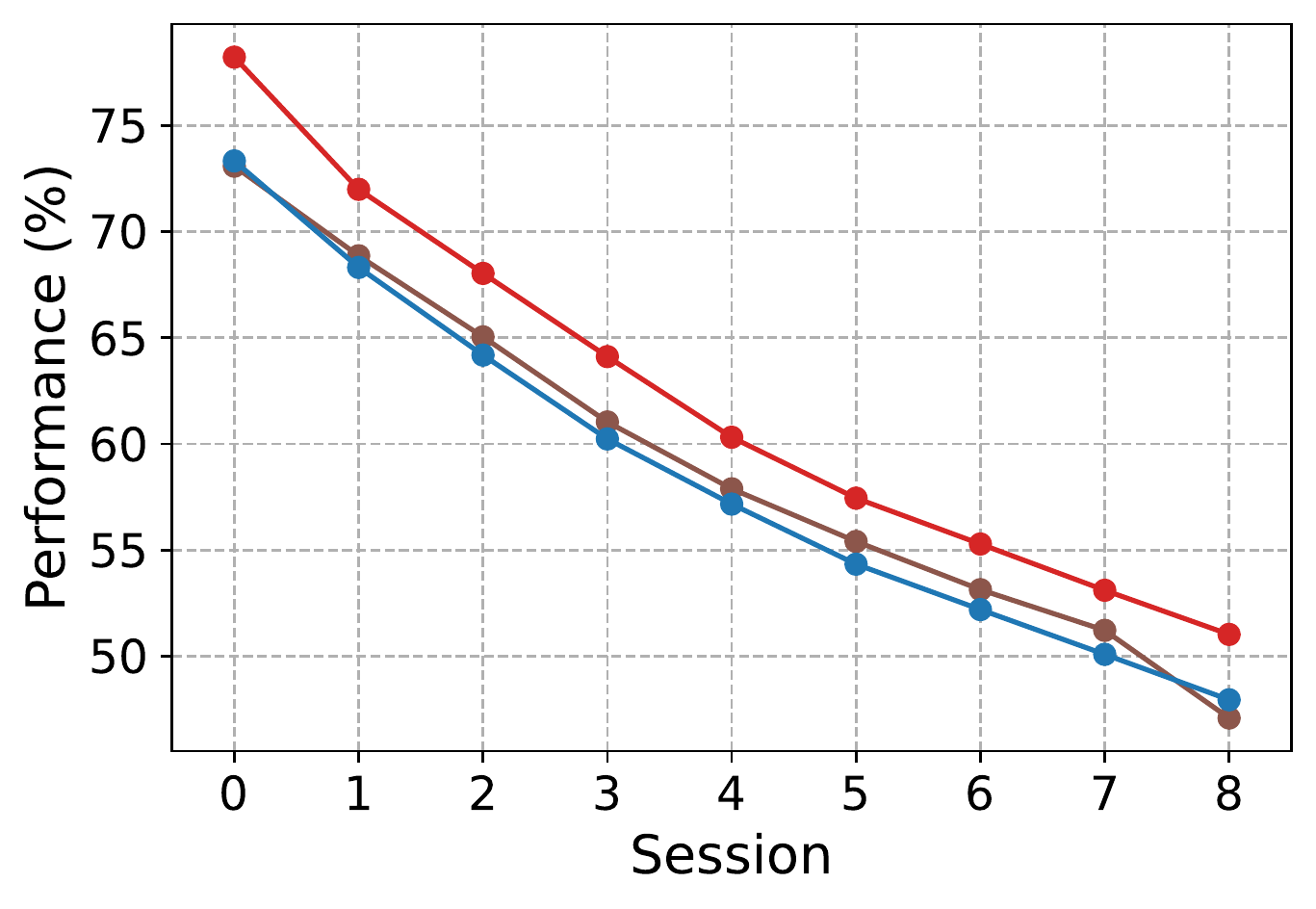}
        \caption{Weighted}
    \end{subfigure}
    \caption{Performances comparison on CIFAR100. The blue, brown, and red lines indicate NoNPC (ours), CEC\,\citep{zhang2021few}, FACT\,\citep{zhou2022forward}.}
    \label{fig:CIFAR_CECFACT}
\end{figure}

\begin{figure}[h]
    \centering
    \begin{subfigure}[h]{0.32\linewidth}
    \includegraphics[width=\linewidth]{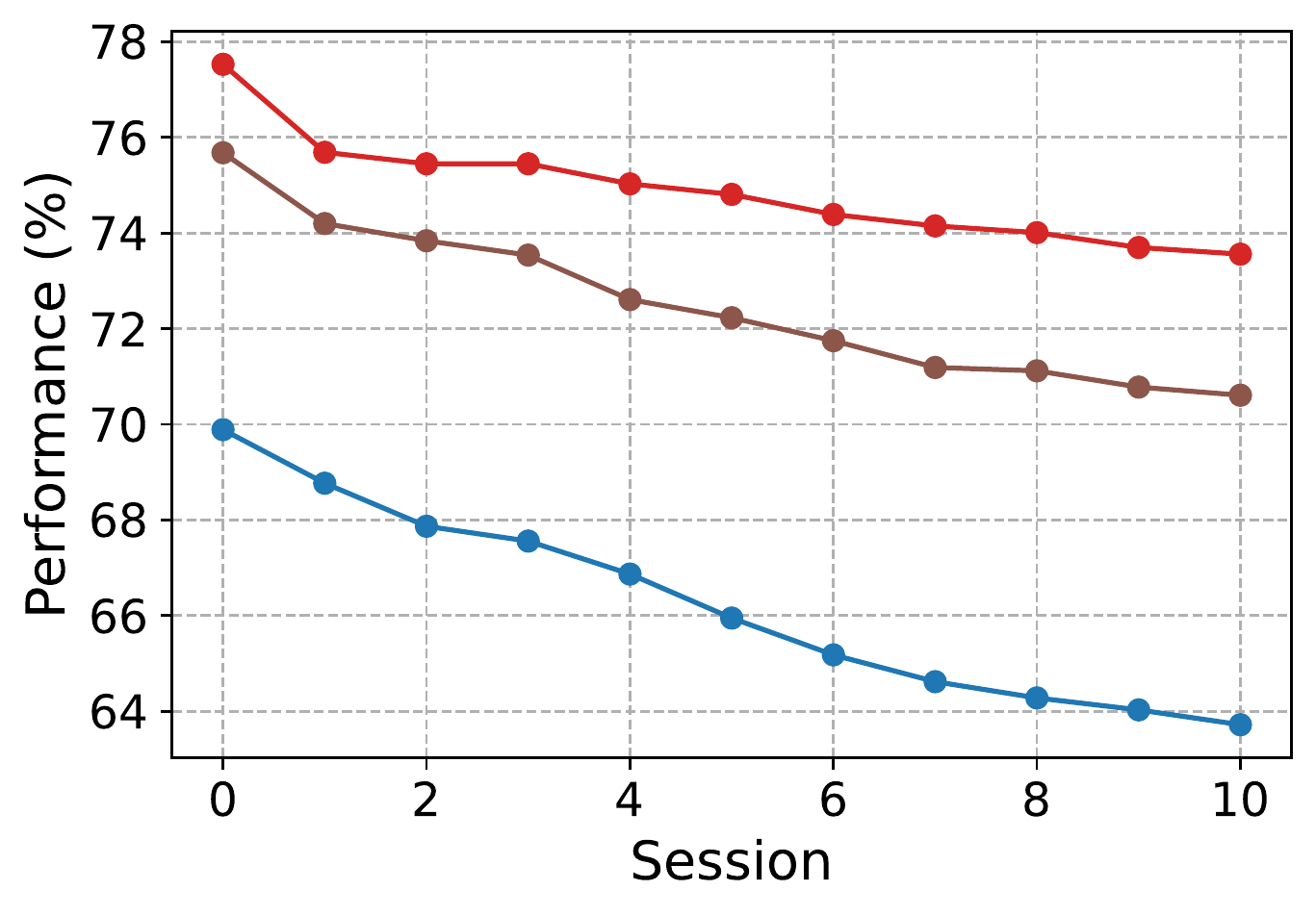}
        \caption{Base}
    \end{subfigure}
    \hfill
    \begin{subfigure}[h]{0.32\linewidth}
    \includegraphics[width=\linewidth]{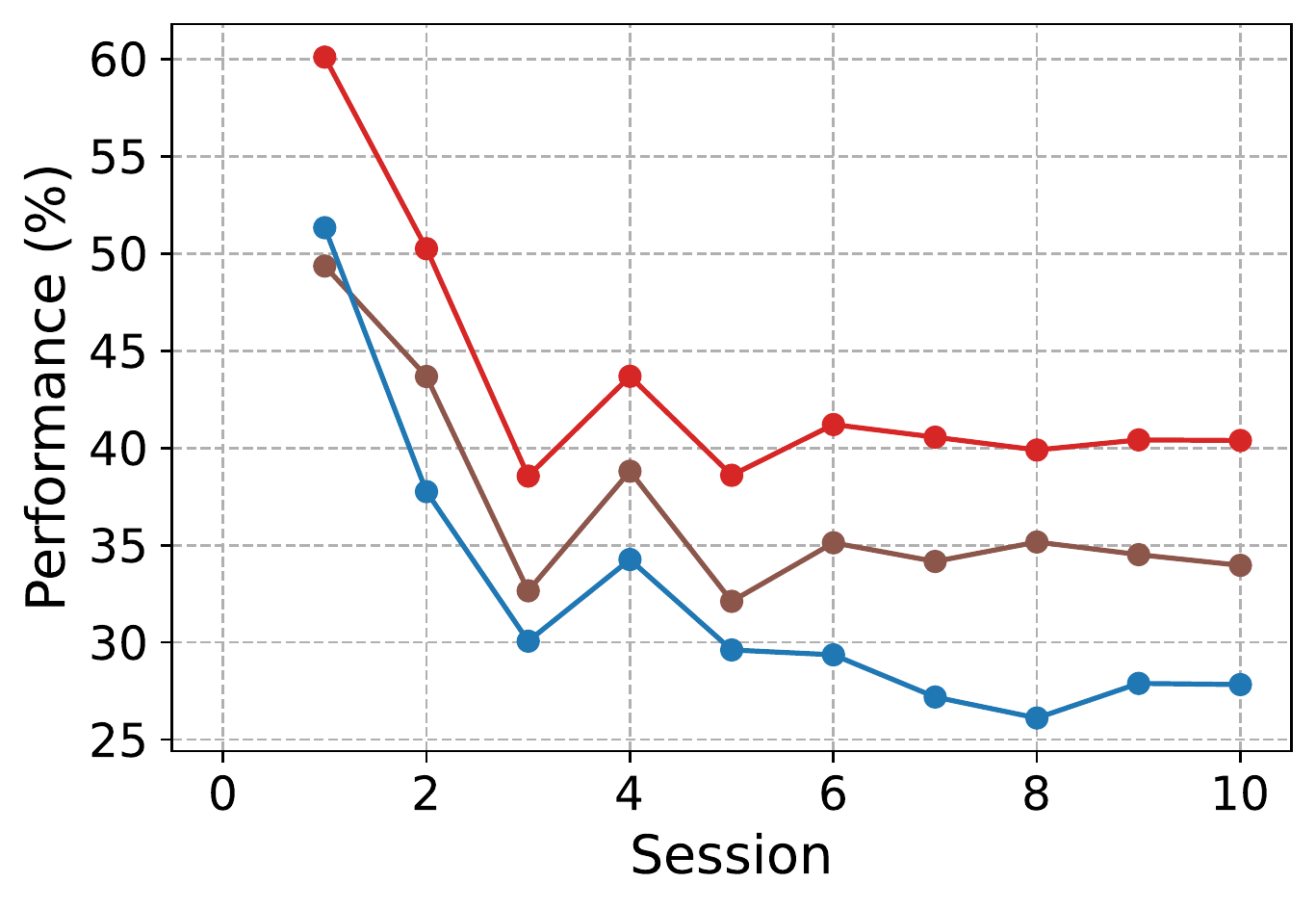}
        \caption{Novel}
    \end{subfigure}
    \hfill
    \begin{subfigure}[h]{0.32\linewidth}
    \includegraphics[width=\linewidth]{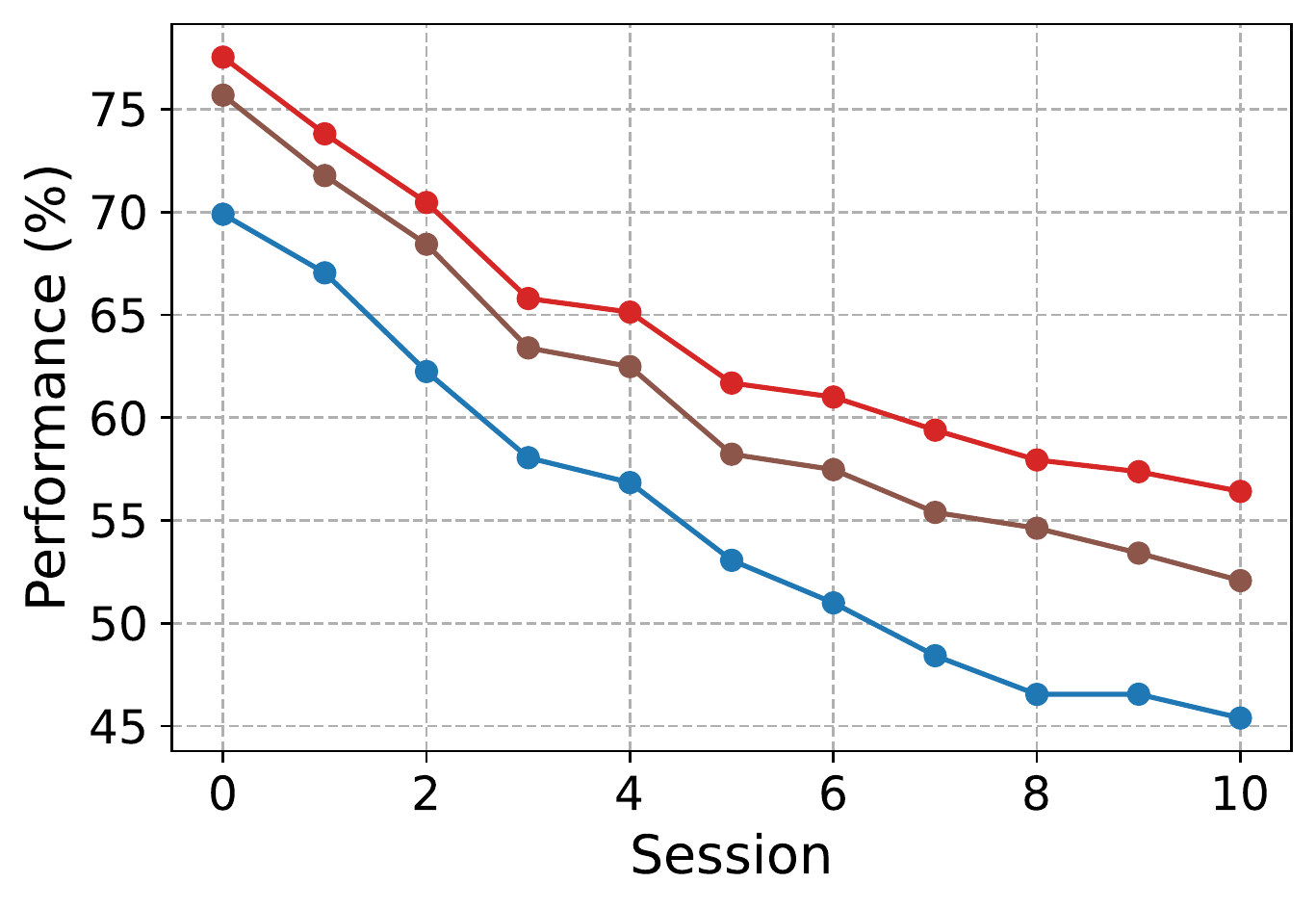}
        \caption{Weighted}
    \end{subfigure}
    \caption{Performances comparison on CUB200. The blue, brown, and red lines indicate NoNPC (ours), CEC\,\citep{zhang2021few}, FACT\,\citep{zhou2022forward}.}
    \label{fig:CUB_CECFACT}
\end{figure}

\begin{figure}[h]
    \centering
    \begin{subfigure}[h]{0.32\linewidth}
    \includegraphics[width=\linewidth]{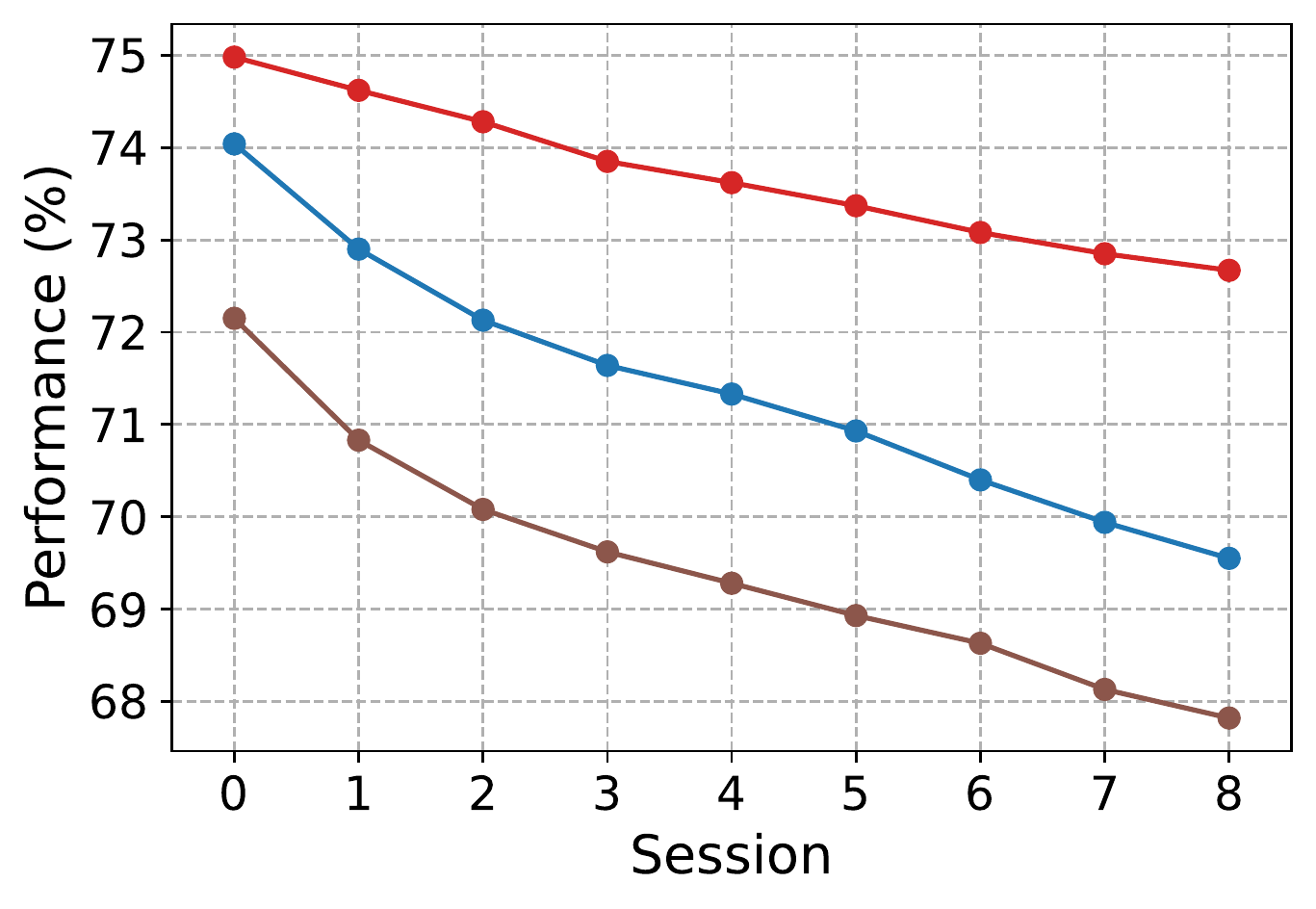}
        \caption{Base}
    \end{subfigure}
    \hfill
    \begin{subfigure}[h]{0.32\linewidth}
    \includegraphics[width=\linewidth]{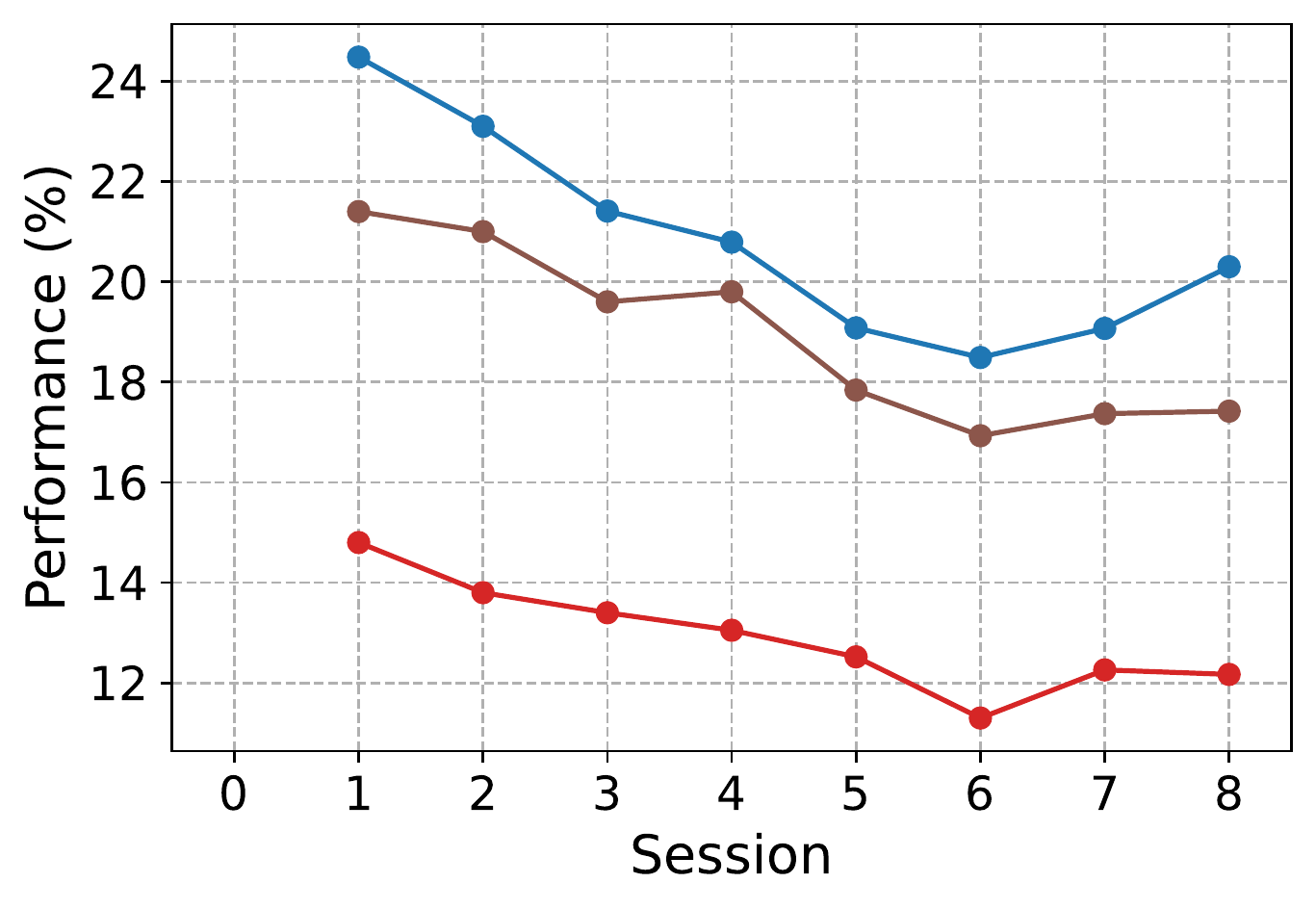}
        \caption{Novel}
    \end{subfigure}
    \hfill
    \begin{subfigure}[h]{0.32\linewidth}
    \includegraphics[width=\linewidth]{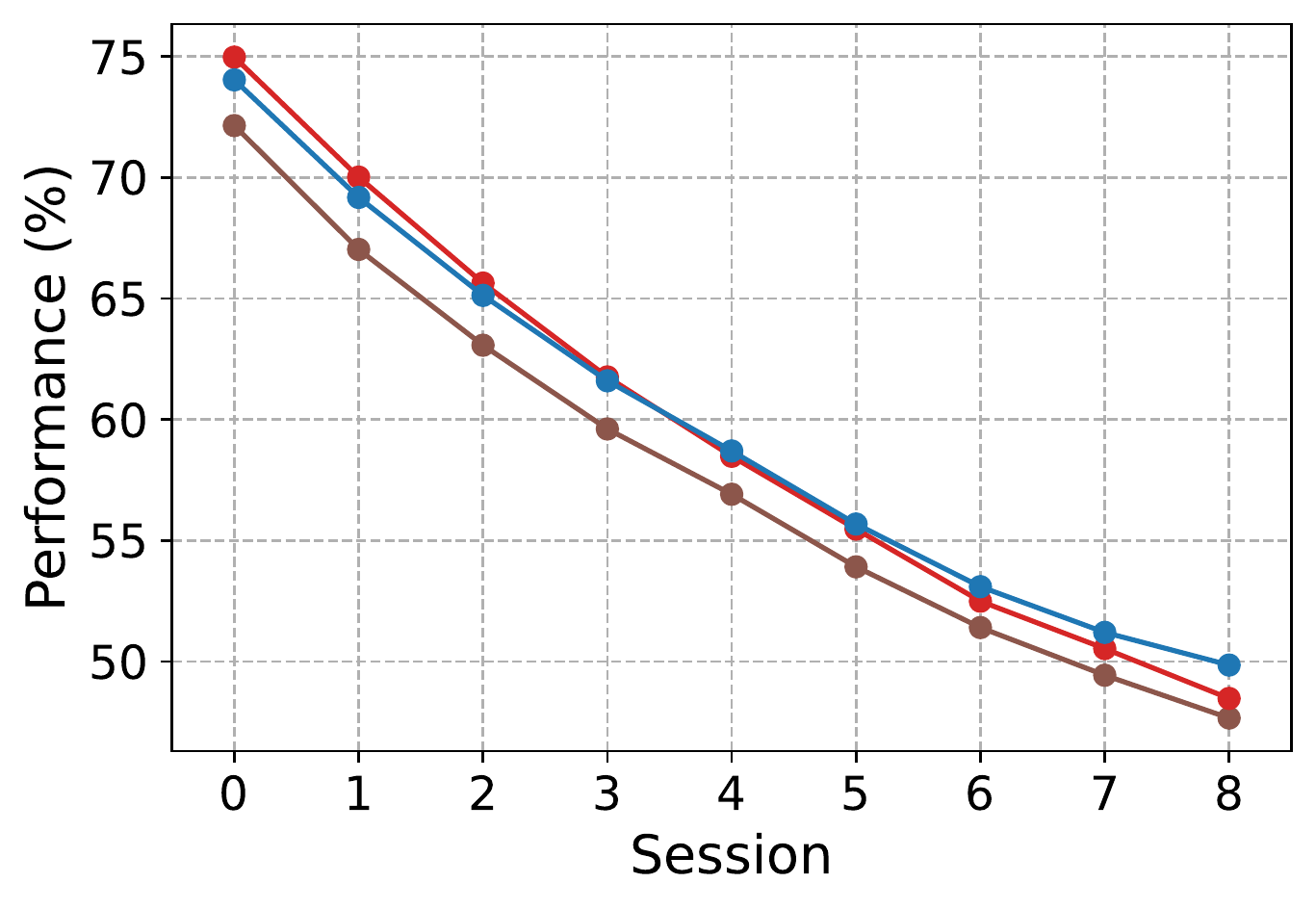}
        \caption{Weighted}
    \end{subfigure}
    \caption{Performances comparison on miniImageNet. The blue, brown, and red lines indicate NoNPC (ours), CEC\,\citep{zhang2021few}, FACT\,\citep{zhou2022forward}.}
    \label{fig:miniIN_CECFACT}
\end{figure}

Table \ref{fig:CIFAR_CECFACT}, \ref{fig:CUB_CECFACT}, and \ref{fig:miniIN_CECFACT} describe the base, novel, and weighted performances on CIFAR100, CUB200, and miniImageNet, respectively.

\newpage
\section{Prototype Normalization}\label{appx:proto}

In this section, for ablation stuides, we compare below models:
\vspace*{-5pt}

\begin{table}[h]
    \centering
    \footnotesize
    \caption{Models for ablation studies related to prototype normalization.}\label{tab:proto_norm_models}
    \vspace*{-0.15cm}
    \begin{tabular}{ccc|ccc}
    \toprule
    Model & Base classifier & Novel classifier & Model & Base classifier & Novel classifier \\
    \midrule
    M1 & -  & -  & M3 & - & NP\\
    M2 & -  & P & M4 (NoNPC) & NP & NP \\
    \bottomrule
    \end{tabular}
    \vspace*{-5pt}
\end{table}

\noindent where `-' of a base classifier and a novel classifier means the optimized classifier via stochastic gradient descent and a random initialized classifier, respectively. `P' and `NP' indicate a prototype classifier and a normalized prototype classifier. Note that M1 in Table~\ref{tab:proto_norm_models} is the same with M1 in Table~\ref{tab:notation}, while other models have nothing to do with models in Table~\ref{tab:notation}.
\vspace*{-5pt}

\begin{figure}[h!]
    \centering
    \includegraphics[width=0.45\linewidth]{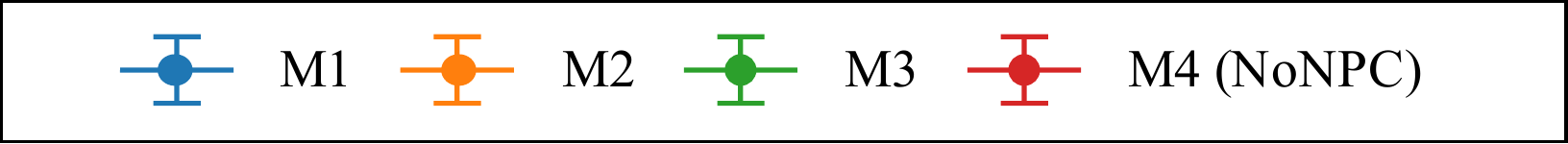}
    
    \begin{subfigure}[h]{0.32\linewidth}
    \includegraphics[width=\linewidth]{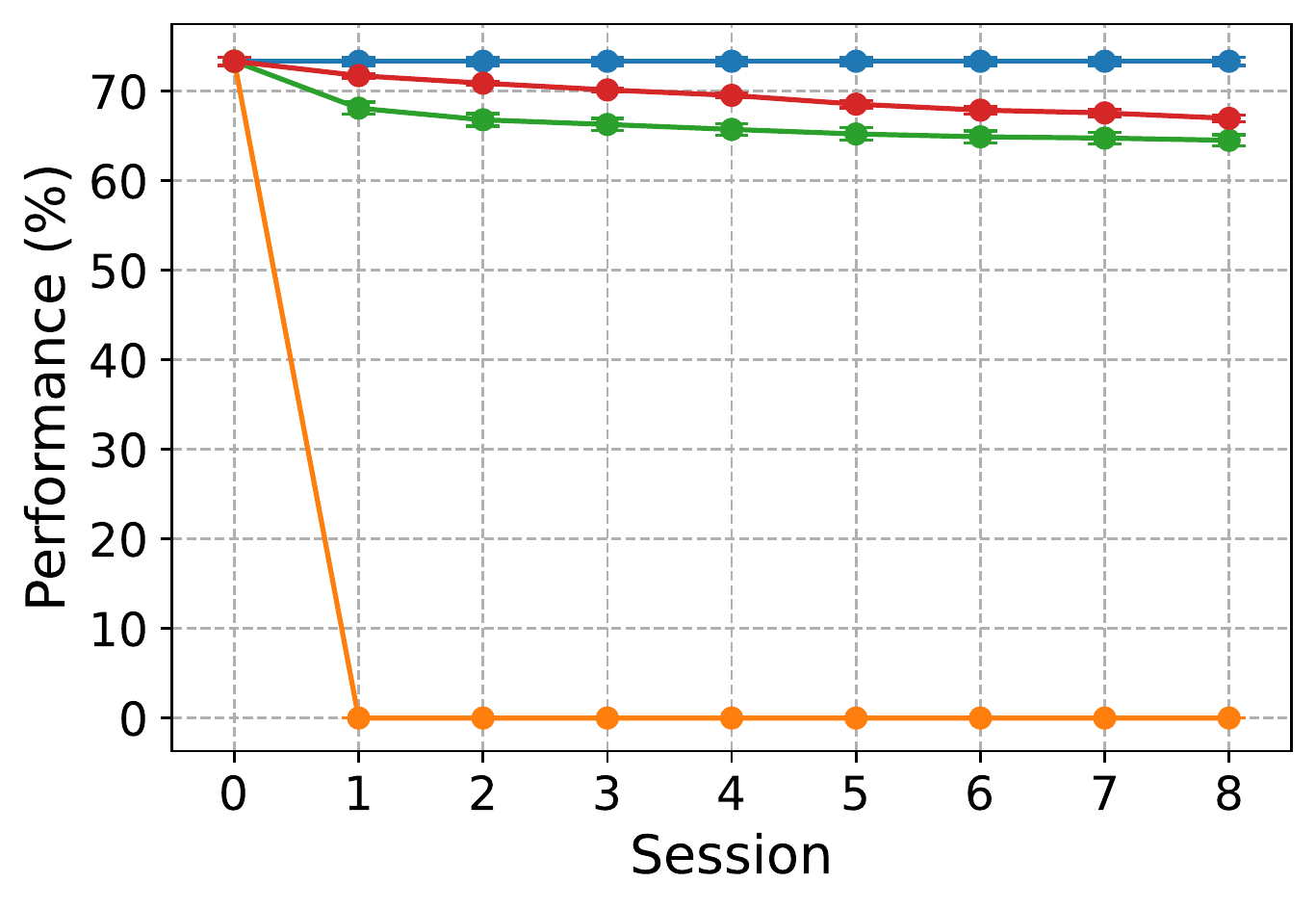}
        \caption{Base}
    \end{subfigure}
    \hfill
    \begin{subfigure}[h]{0.32\linewidth}
    \includegraphics[width=\linewidth]{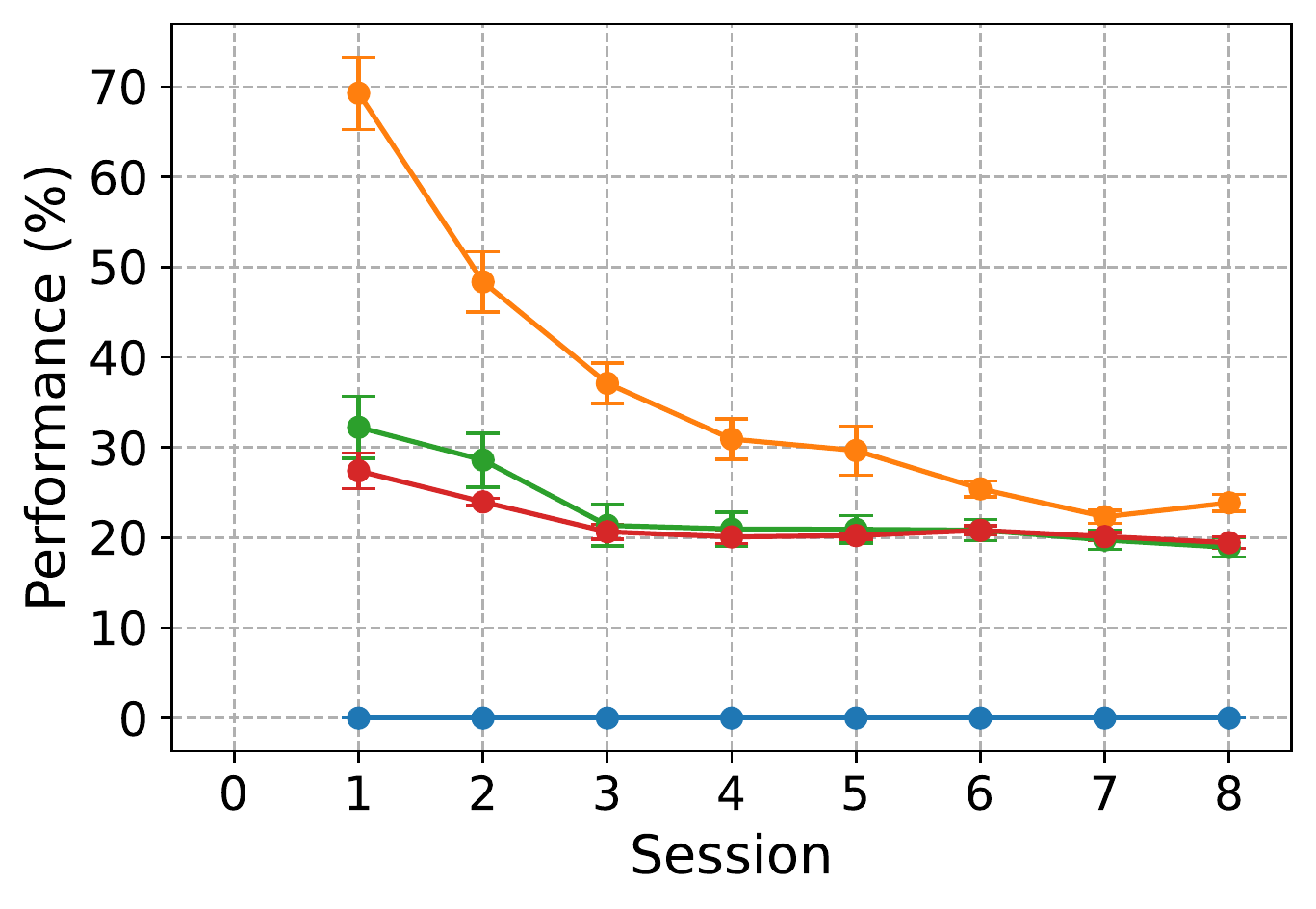}
        \caption{Novel}
    \end{subfigure}
    \hfill
    \begin{subfigure}[h]{0.32\linewidth}
    \includegraphics[width=\linewidth]{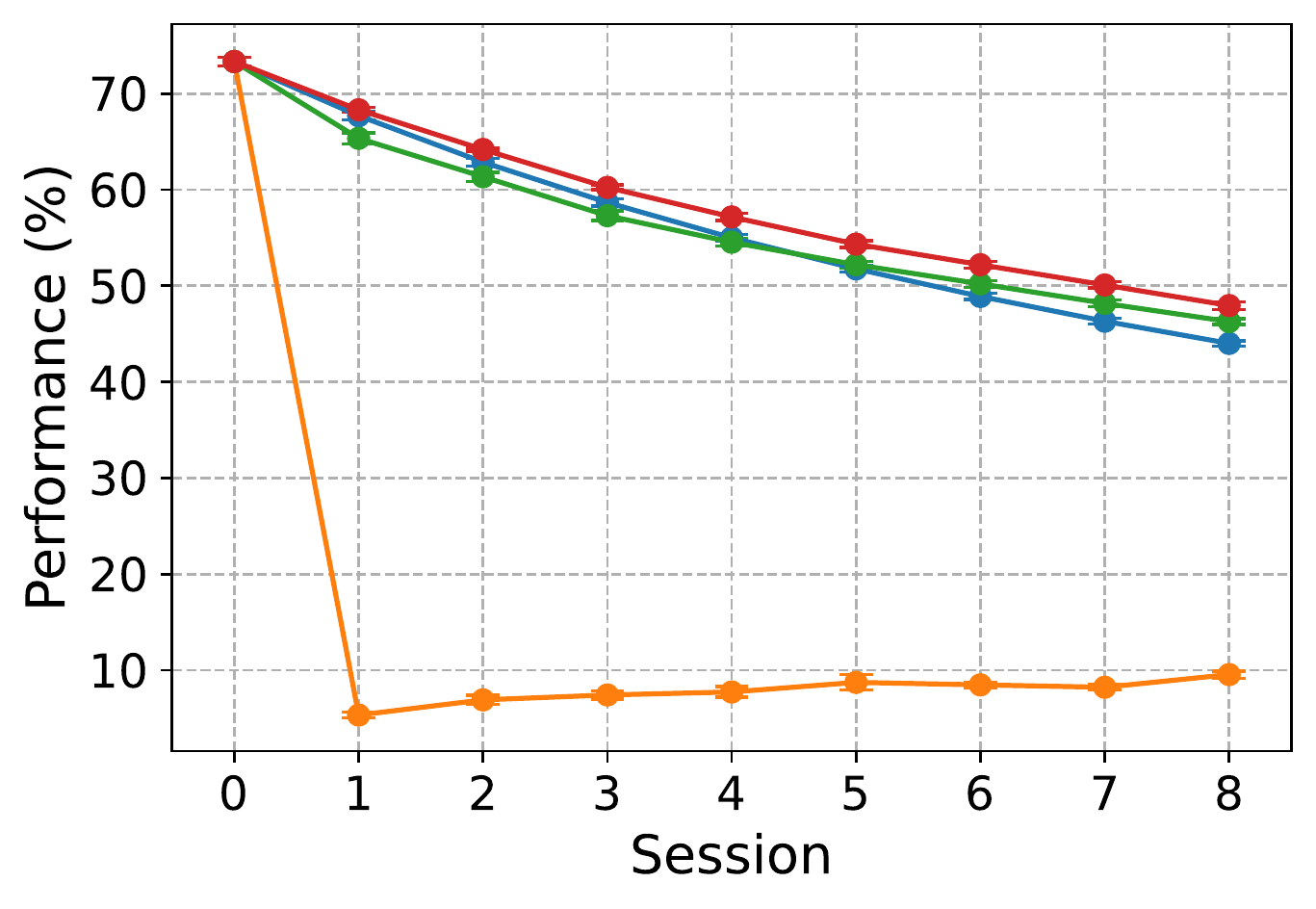}
        \caption{Weighted}
    \end{subfigure}
    \vspace*{-5pt}
    \caption{Ablation study according to prototype normalization on CIFAR100.}
    \label{fig:CIFAR_abla}
    \vspace*{-18pt}
\end{figure}

\begin{figure}[h!]
    \centering
    \begin{subfigure}[h]{0.32\linewidth}
    \includegraphics[width=\linewidth]{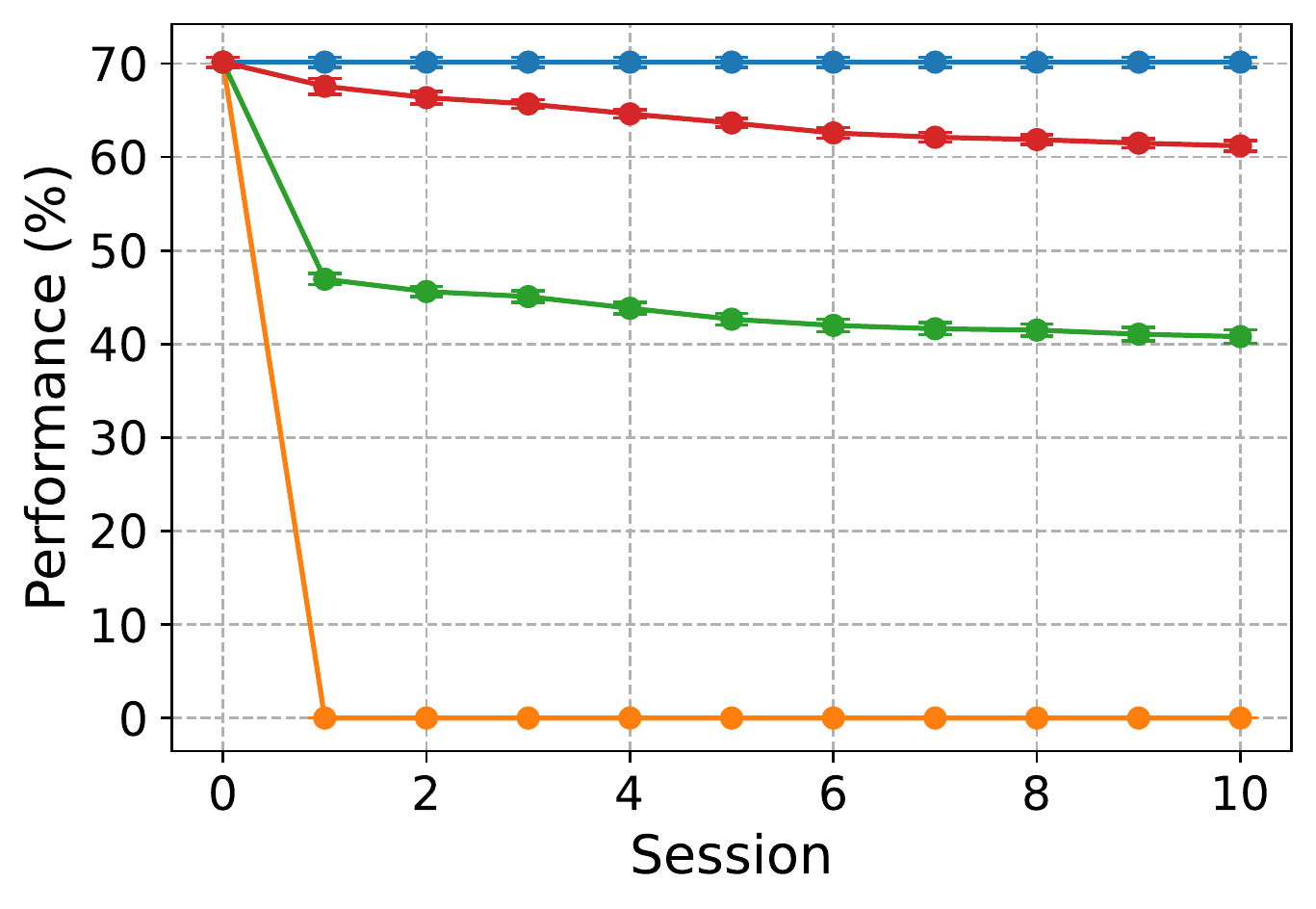}
        \caption{Base}
    \end{subfigure}
    \hfill
    \begin{subfigure}[h]{0.32\linewidth}
    \includegraphics[width=\linewidth]{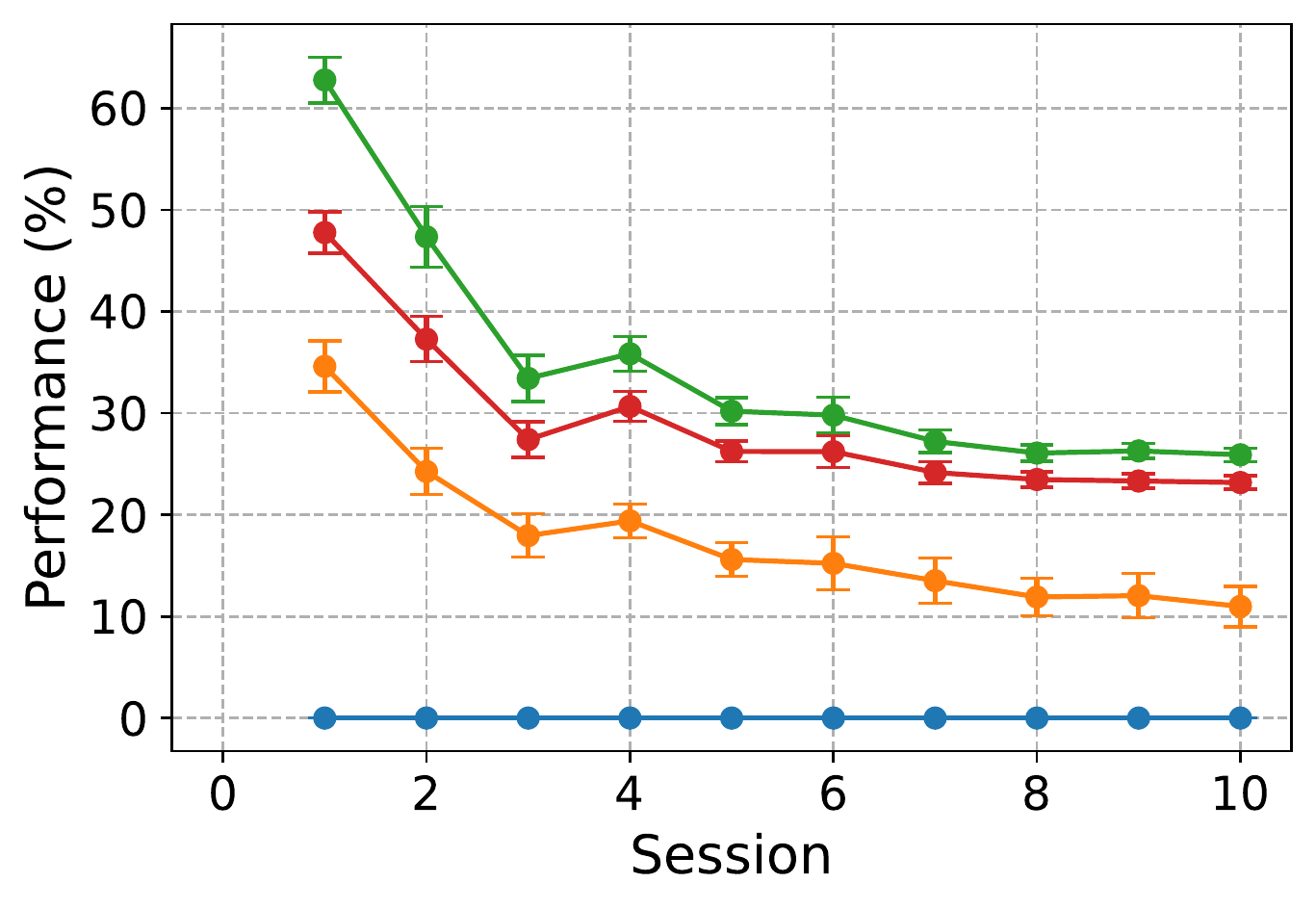}
        \caption{Novel}
    \end{subfigure}
    \hfill
    \begin{subfigure}[h]{0.32\linewidth}
    \includegraphics[width=\linewidth]{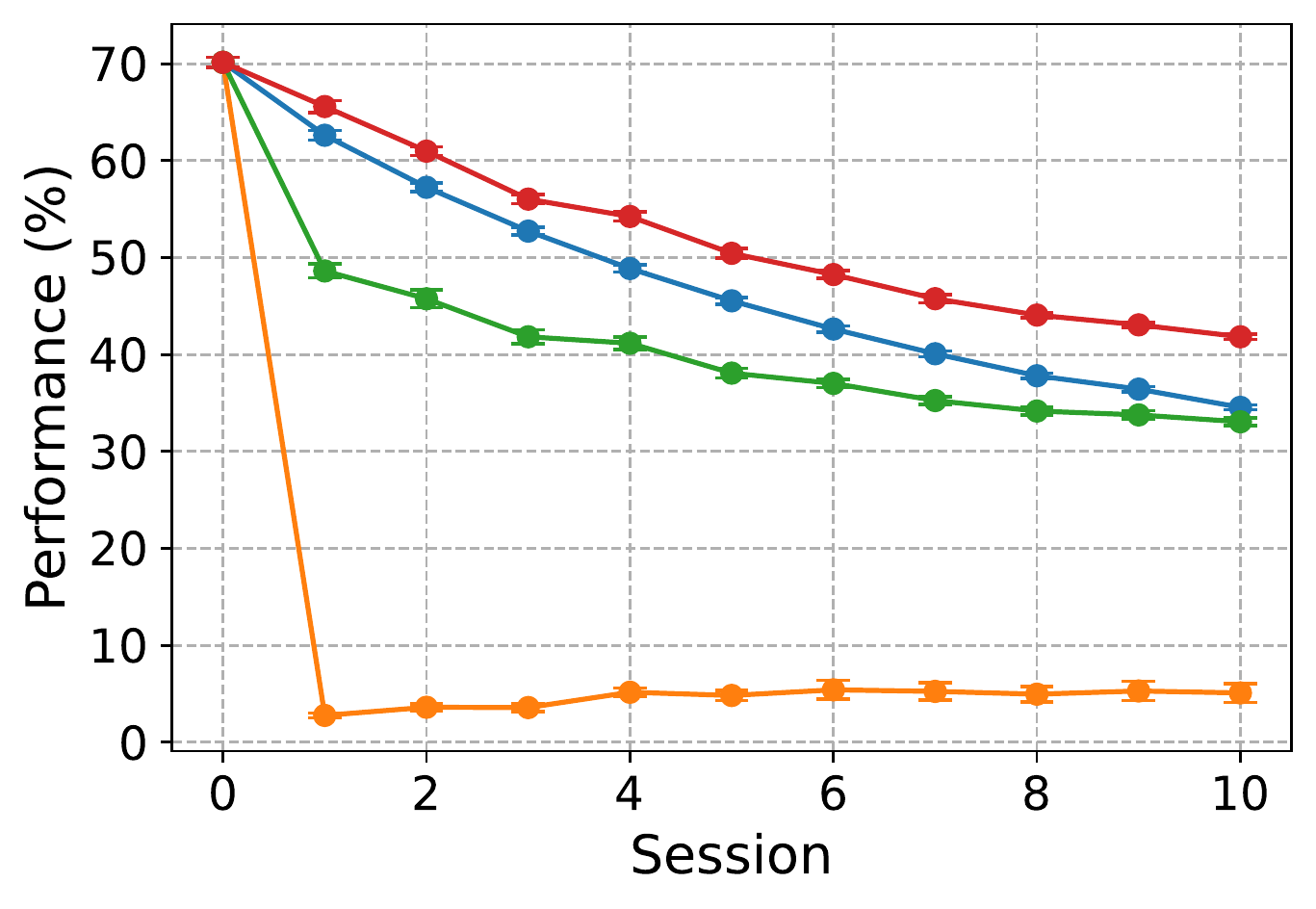}
        \caption{Weighted}
    \end{subfigure}
    \vspace*{-5pt}
    \caption{Ablation study according to prototype normalization on CUB200.}
    \label{fig:cub_abla}
    \vspace*{-15pt}
\end{figure}

\begin{figure}[h!]
    \centering
    \begin{subfigure}[h]{0.32\linewidth}
    \includegraphics[width=\linewidth]{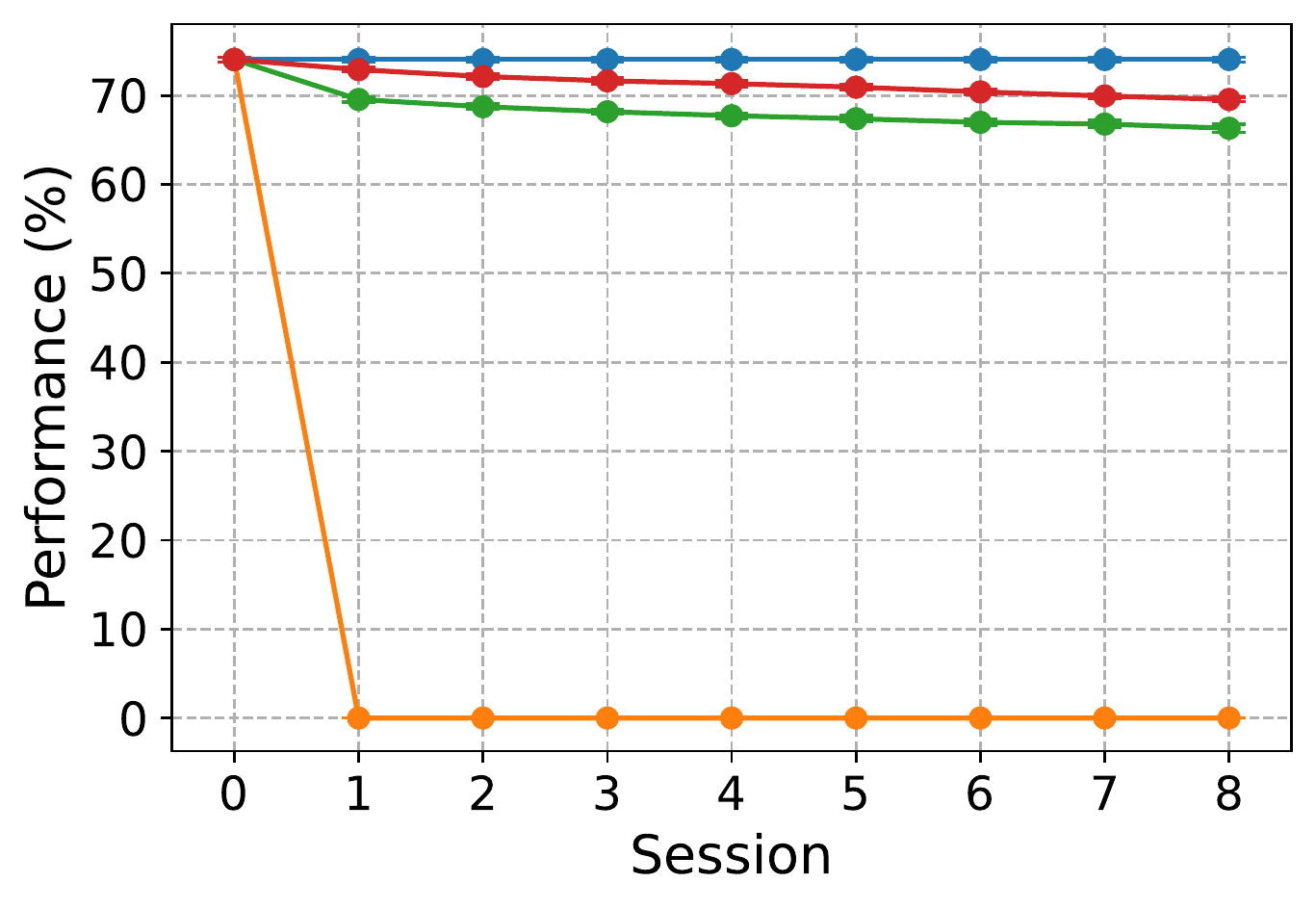}
        \caption{Base}
    \end{subfigure}
    \hfill
    \begin{subfigure}[h]{0.32\linewidth}
    \includegraphics[width=\linewidth]{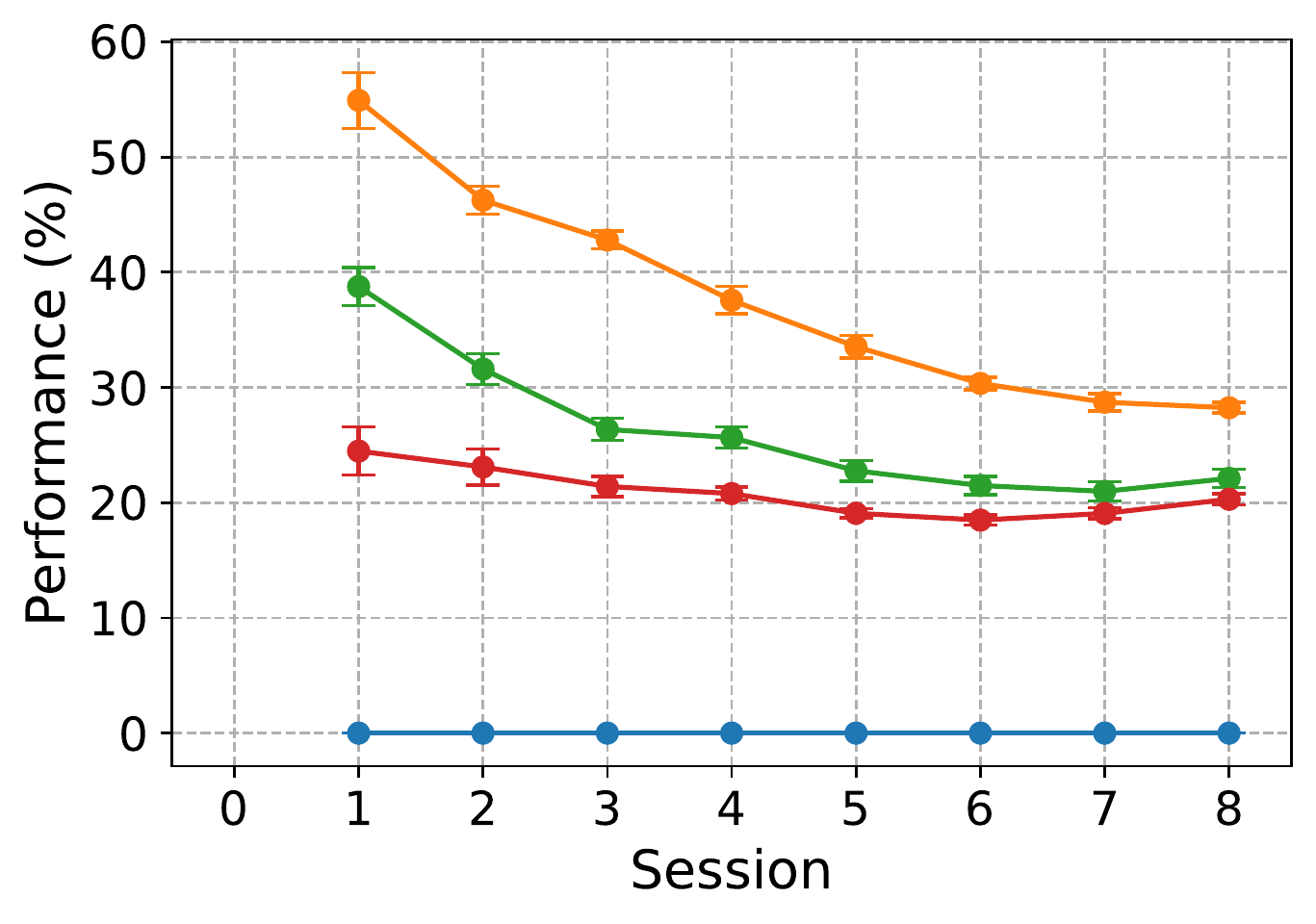}
        \caption{Novel}
    \end{subfigure}
    \hfill
    \begin{subfigure}[h]{0.32\linewidth}
    \includegraphics[width=\linewidth]{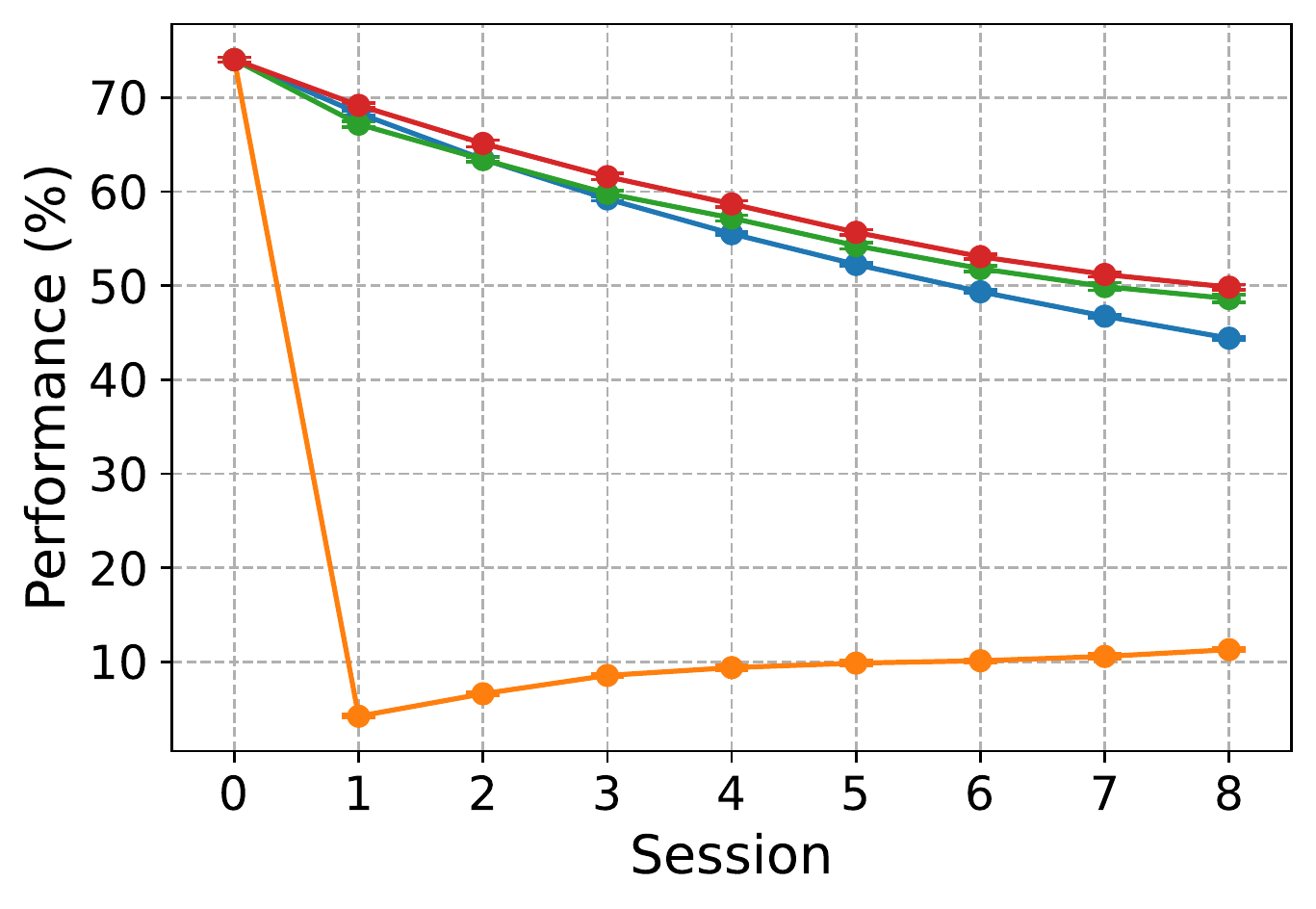}
        \caption{Weighted}
    \end{subfigure}
    \vspace*{-5pt}
    \caption{Ablation study according to prototype normalization on miniImageNet.}
    \label{fig:mini_abla}
\end{figure}

\clearpage
\section{Label Smoothing during the Base Session}\label{appx:ls}

\begin{figure}[h]
    \centering
    \begin{subfigure}[h]{0.29\linewidth}
    \includegraphics[width=\linewidth]{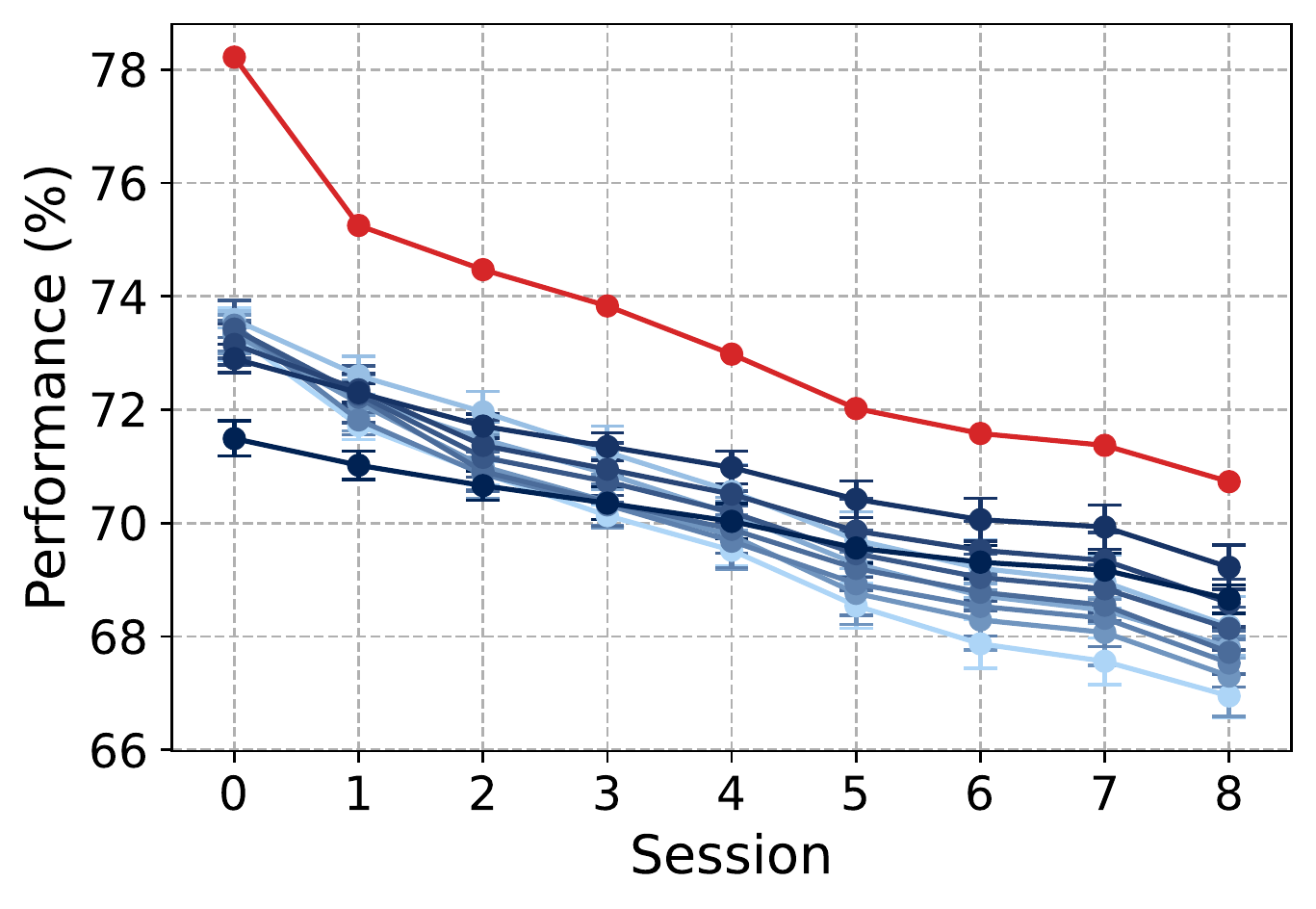}
        \caption{Base}
    \end{subfigure}
    \hfill
    \begin{subfigure}[h]{0.29\linewidth}
    \includegraphics[width=\linewidth]{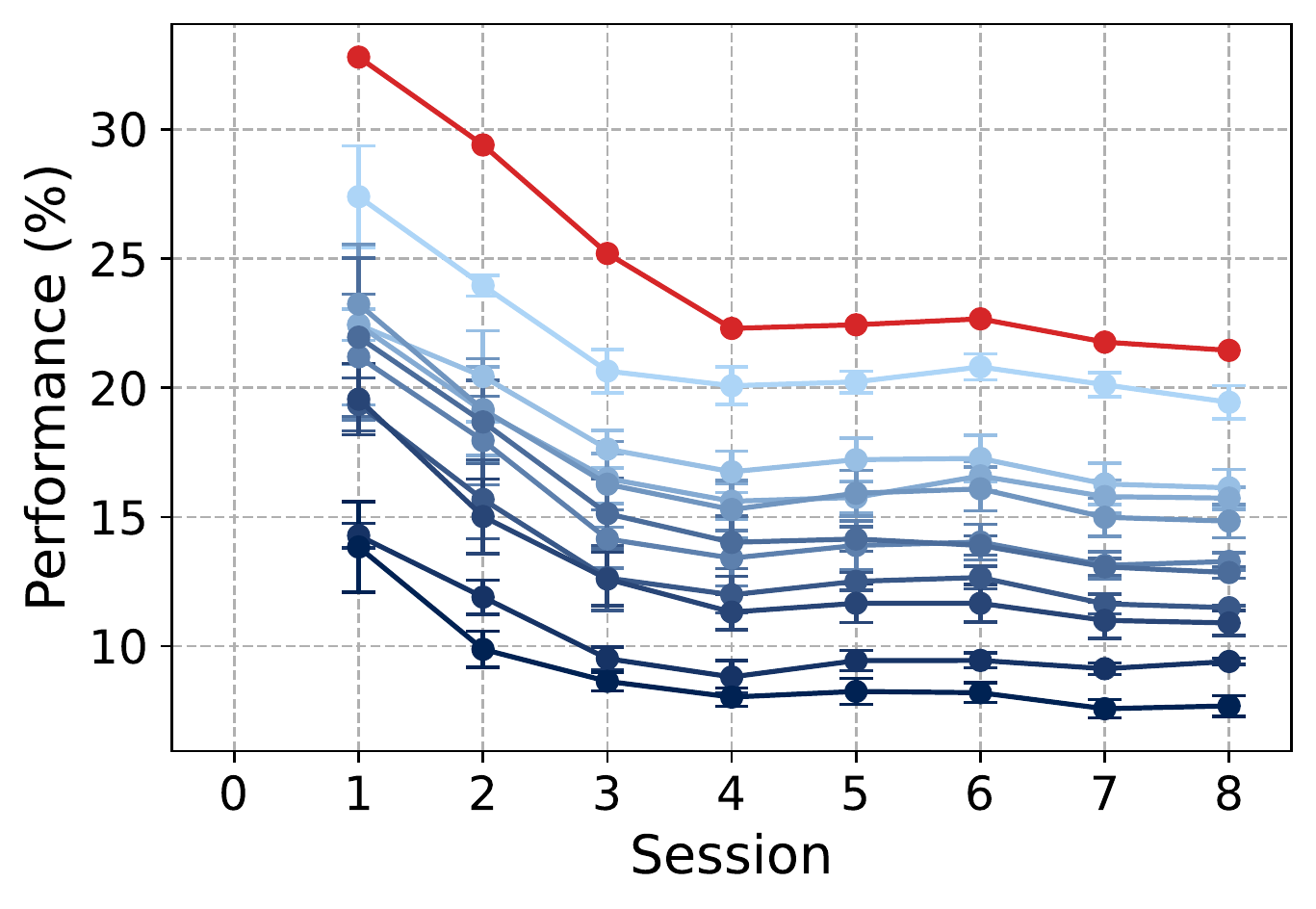}
        \caption{Novel}
    \end{subfigure}
    \hfill
    \begin{subfigure}[h]{0.29\linewidth}
    \includegraphics[width=\linewidth]{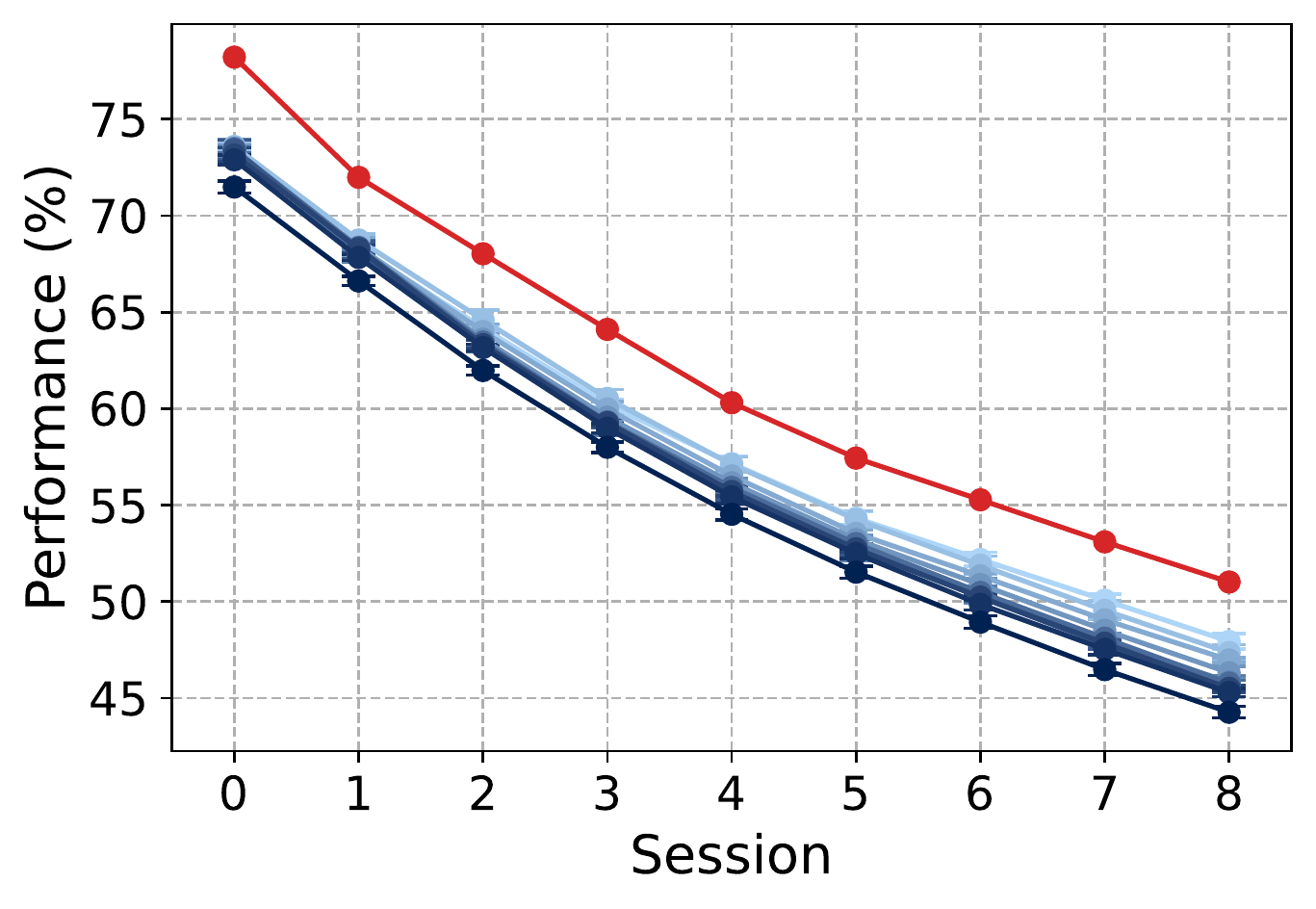}
        \caption{Weighted}
    \end{subfigure}
    \hfill
    \begin{subfigure}[h]{0.10\linewidth}
    \includegraphics[width=\linewidth]{figure/legend3.pdf}
    \end{subfigure}
    \vspace*{-3pt}
    \caption{Base, novel, and weighted performances according to the degree of label smoothing on CIFAR100. The most light and dark blue colors indicate smoothness of 0.0 and 0.9, respectively. The value of 0.0 means label smoothing is not used. The red line indicates the performance of FACT\,\citep{zhou2022forward}.}
    \label{fig:cifar_ls}
\end{figure}

\begin{figure}[h]
    \centering
    \begin{subfigure}[h]{0.29\linewidth}
    \includegraphics[width=\linewidth]{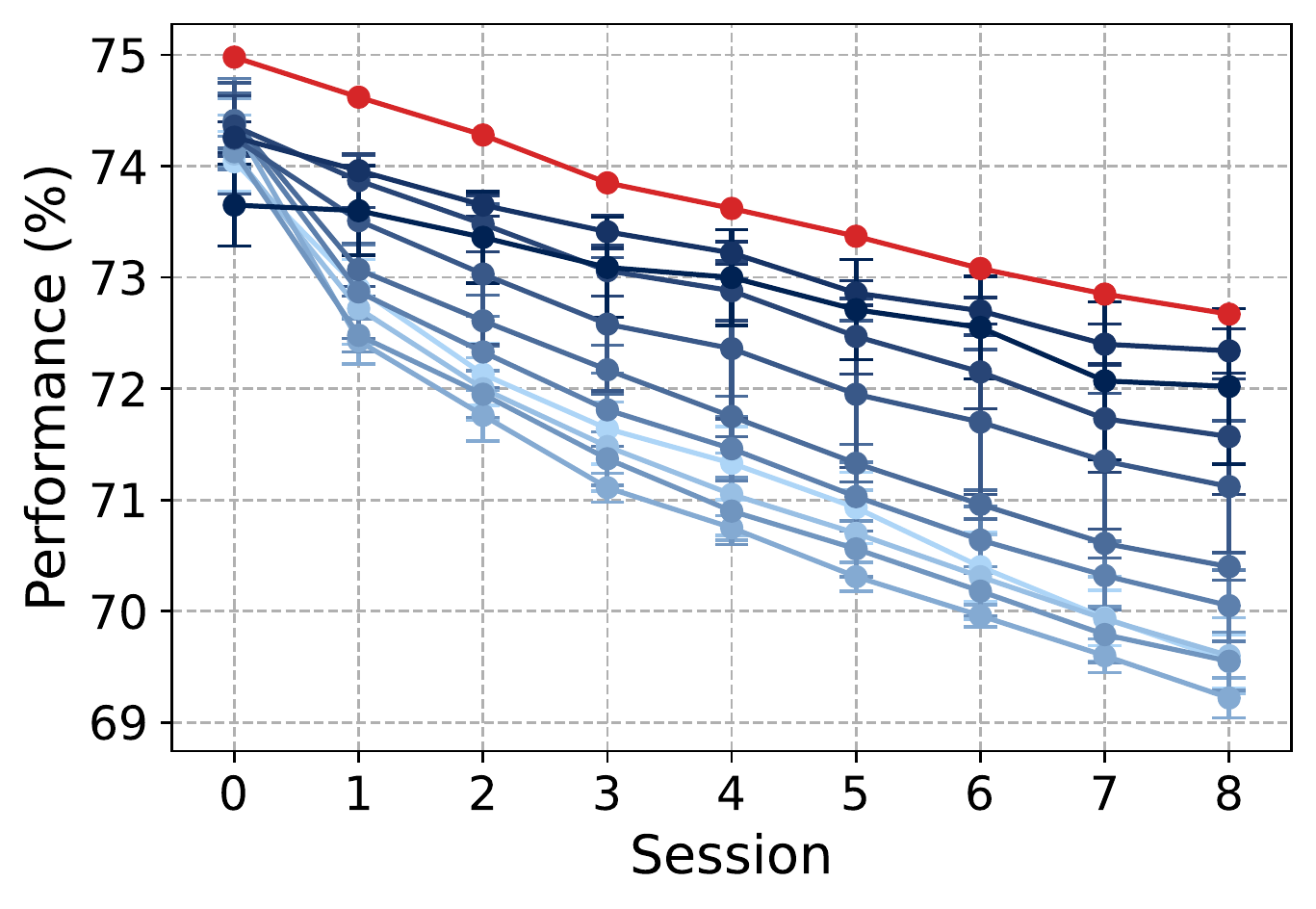}
        \caption{Base}
    \end{subfigure}
    \hfill
    \begin{subfigure}[h]{0.29\linewidth}
    \includegraphics[width=\linewidth]{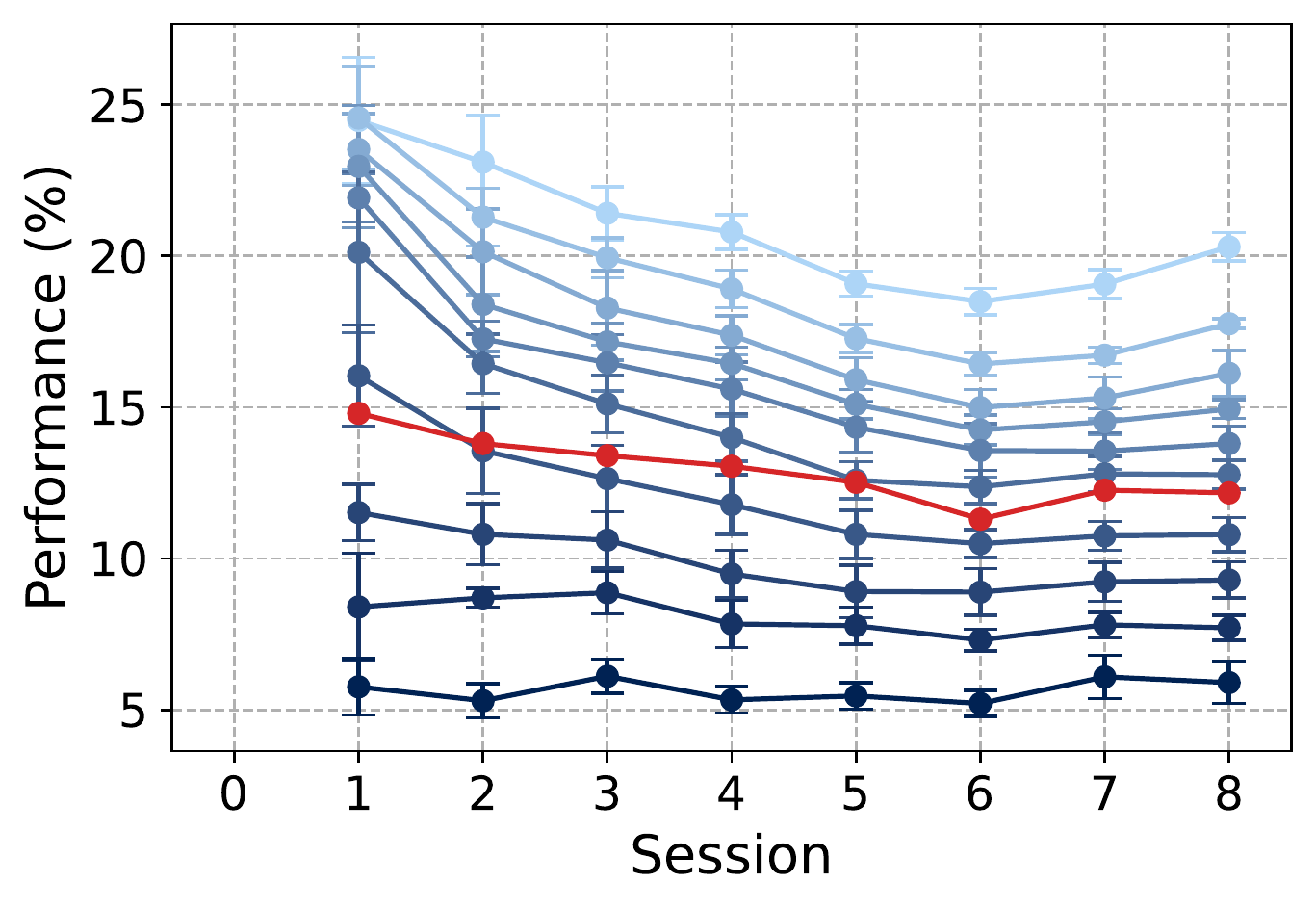}
        \caption{Novel}
    \end{subfigure}
    \hfill
    \begin{subfigure}[h]{0.29\linewidth}
    \includegraphics[width=\linewidth]{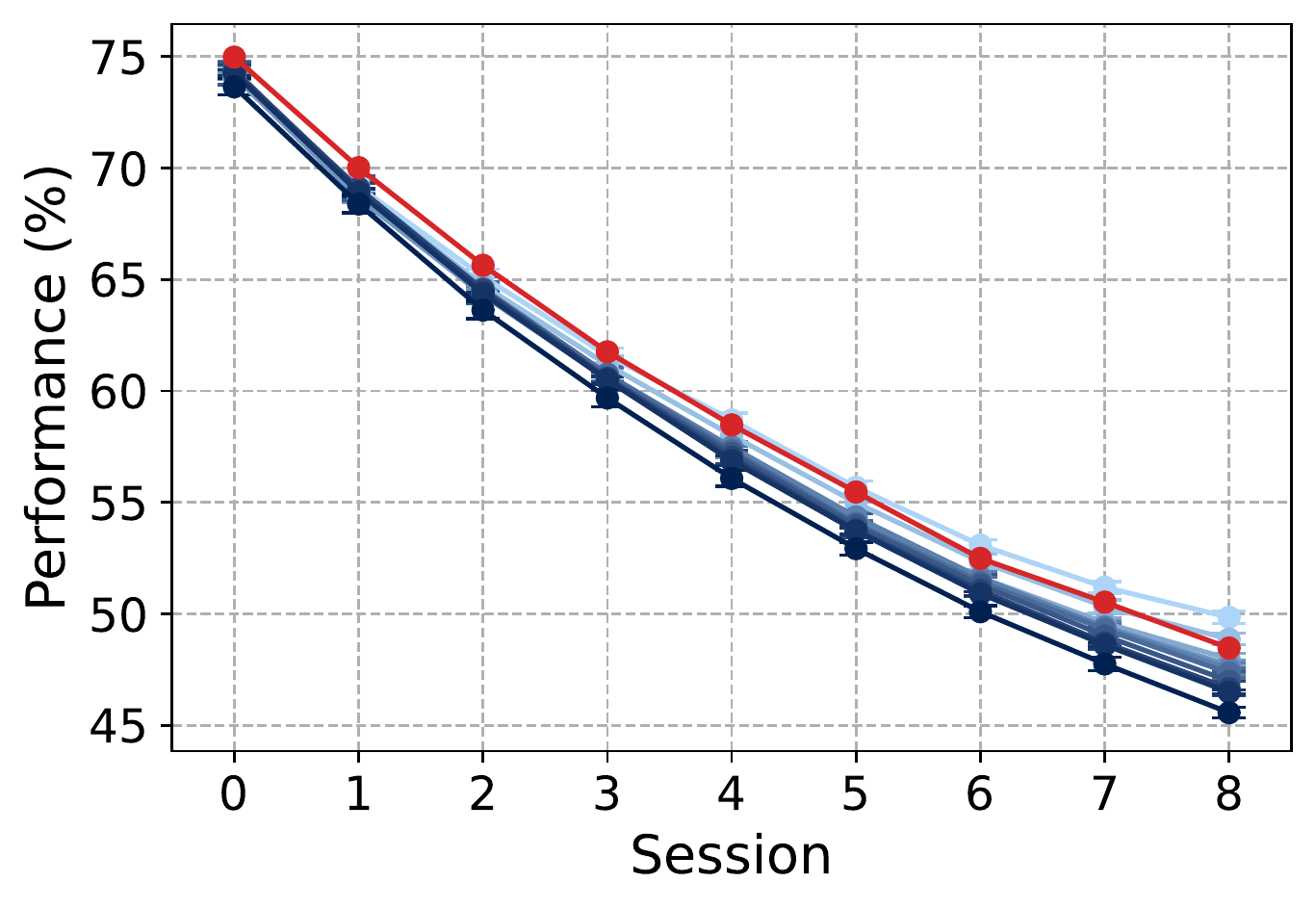}
        \caption{Weighted}
    \end{subfigure}
    \hfill
    \begin{subfigure}[h]{0.10\linewidth}
    \includegraphics[width=\linewidth]{figure/legend3.pdf}
    \end{subfigure}
    \vspace*{-3pt}
    \caption{Base, novel, and weighted performances according to the degree of label smoothing on miniImageNet. The most light and dark blue colors indicate smoothness of 0.0 and 0.9, respectively. The value of 0.0 means label smoothing is not used. The red line indicates the performance of FACT\,\citep{zhou2022forward}.}
    \label{fig:mini_ls}
\end{figure}

Figure \ref{fig:cifar_ls} and \ref{fig:mini_ls} describe the base, novel, and weighted performances according to the degree of label smoothing on CIFAR100 and miniImageNet, respectively. The tendency for the novel classes is opposite to the tendency on CUB200. For coarse-grained datasets, small smoothness can increase the novel performance rather than large smoothness.

\newpage
\bibliography{reference}

\begin{thebibliography}{28}
\providecommand{\natexlab}[1]{#1}
\providecommand{\url}[1]{\texttt{#1}}
\expandafter\ifx\csname urlstyle\endcsname\relax
  \providecommand{\doi}[1]{doi: #1}\else
  \providecommand{\doi}{doi: \begingroup \urlstyle{rm}\Url}\fi

\bibitem[Arivazhagan et~al.(2019)Arivazhagan, Aggarwal, Singh, and
  Choudhary]{arivazhagan2019federated}
Manoj~Ghuhan Arivazhagan, Vinay Aggarwal, Aaditya~Kumar Singh, and Sunav
  Choudhary.
\newblock Federated learning with personalization layers.
\newblock \emph{arXiv preprint arXiv:1912.00818}, 2019.

\bibitem[Belouadah and Popescu(2019)]{belouadah2019IL2M}
Eden Belouadah and Adrian Popescu.
\newblock Il2m: Class incremental learning with dual memory.
\newblock In \emph{Proceedings of the IEEE/CVF International Conference on
  Computer Vision (ICCV)}, October 2019.

\bibitem[Collins et~al.(2021)Collins, Hassani, Mokhtari, and
  Shakkottai]{collins2021exploiting}
Liam Collins, Hamed Hassani, Aryan Mokhtari, and Sanjay Shakkottai.
\newblock Exploiting shared representations for personalized federated
  learning.
\newblock In \emph{International Conference on Machine Learning}, pages
  2089--2099. PMLR, 2021.

\bibitem[Davari et~al.(2022)Davari, Asadi, Mudur, Aljundi, and
  Belilovsky]{davari2022probing}
MohammadReza Davari, Nader Asadi, Sudhir Mudur, Rahaf Aljundi, and Eugene
  Belilovsky.
\newblock Probing representation forgetting in supervised and unsupervised
  continual learning.
\newblock In \emph{Proceedings of the IEEE/CVF Conference on Computer Vision
  and Pattern Recognition (CVPR)}, 2022.

\bibitem[Fini et~al.(2020)Fini, Lathuiliere, Sangineto, Nabi, and
  Ricci]{fini2020online}
Enrico Fini, St{\'e}phane Lathuiliere, Enver Sangineto, Moin Nabi, and Elisa
  Ricci.
\newblock Online continual learning under extreme memory constraints.
\newblock In \emph{European Conference on Computer Vision}, pages 720--735.
  Springer, 2020.

\bibitem[Hersche et~al.(2022)Hersche, Karunaratne, Cherubini, Benini,
  Sebastian, and Rahimi]{hersche2022cfscil}
Michael Hersche, Geethan Karunaratne, Giovanni Cherubini, Luca Benini, Abu
  Sebastian, and Abbas Rahimi.
\newblock Constrained few-shot class-incremental learning.
\newblock In \emph{Proceedings of the IEEE Conference on Computer Vision and
  Pattern Recognition (CVPR)}, 2022.

\bibitem[Jie et~al.(2022)Jie, Deng, and Li]{jie2022alleviating}
Shibo Jie, Zhi-Hong Deng, and Ziheng Li.
\newblock Alleviating representational shift for continual fine-tuning.
\newblock \emph{arXiv preprint arXiv:2204.10535}, 2022.

\bibitem[Joseph et~al.(2022)Joseph, Khan, Khan, Anwer, and
  Balasubramanian]{joseph2022energy}
KJ~Joseph, Salman Khan, Fahad~Shahbaz Khan, Rao~Muhammad Anwer, and Vineeth~N
  Balasubramanian.
\newblock Energy-based latent aligner for incremental learning.
\newblock In \emph{Proceedings of the IEEE/CVF Conference on Computer Vision
  and Pattern Recognition}, 2022.

\bibitem[Kang et~al.(2020)Kang, Xie, Rohrbach, Yan, Gordo, Feng, and
  Kalantidis]{Kang2020Decoupling}
Bingyi Kang, Saining Xie, Marcus Rohrbach, Zhicheng Yan, Albert Gordo, Jiashi
  Feng, and Yannis Kalantidis.
\newblock Decoupling representation and classifier for long-tailed recognition.
\newblock In \emph{International Conference on Learning Representations}, 2020.
\newblock URL \url{https://openreview.net/forum?id=r1gRTCVFvB}.

\bibitem[Krizhevsky et~al.(2009)Krizhevsky, Hinton,
  et~al.]{krizhevsky2009learning}
Alex Krizhevsky, Geoffrey Hinton, et~al.
\newblock Learning multiple layers of features from tiny images.
\newblock 2009.

\bibitem[Lee and Choi(2018)]{lee2018gradient}
Yoonho Lee and Seungjin Choi.
\newblock Gradient-based meta-learning with learned layerwise metric and
  subspace.
\newblock In \emph{International Conference on Machine Learning}, pages
  2927--2936. PMLR, 2018.

\bibitem[Mai et~al.(2022)Mai, Li, Jeong, Quispe, Kim, and
  Sanner]{mai2022online}
Zheda Mai, Ruiwen Li, Jihwan Jeong, David Quispe, Hyunwoo Kim, and Scott
  Sanner.
\newblock Online continual learning in image classification: An empirical
  survey.
\newblock \emph{Neurocomputing}, 469:\penalty0 28--51, 2022.

\bibitem[Masana et~al.(2020)Masana, Liu, Twardowski, Menta, Bagdanov, and
  van~de Weijer]{masana2020class}
Marc Masana, Xialei Liu, Bartlomiej Twardowski, Mikel Menta, Andrew~D Bagdanov,
  and Joost van~de Weijer.
\newblock Class-incremental learning: survey and performance evaluation on
  image classification.
\newblock \emph{arXiv preprint arXiv:2010.15277}, 2020.

\bibitem[Mazumder et~al.(2021)Mazumder, Singh, and Rai]{mazumder2021few}
Pratik Mazumder, Pravendra Singh, and Piyush Rai.
\newblock Few-shot lifelong learning.
\newblock 2021.

\bibitem[M{\"u}ller et~al.(2019)M{\"u}ller, Kornblith, and
  Hinton]{muller2019does}
Rafael M{\"u}ller, Simon Kornblith, and Geoffrey~E Hinton.
\newblock When does label smoothing help?
\newblock \emph{Advances in neural information processing systems}, 32, 2019.

\bibitem[Oh et~al.(2021)Oh, Yoo, Kim, and Yun]{oh2021boil}
Jaehoon Oh, Hyungjun Yoo, ChangHwan Kim, and Se-Young Yun.
\newblock Boil: Towards representation change for few-shot learning.
\newblock In \emph{International Conference on Learning Representations}, 2021.
\newblock URL \url{https://openreview.net/forum?id=umIdUL8rMH}.

\bibitem[Oh et~al.(2022)Oh, Kim, and Yun]{oh2022fedbabu}
Jaehoon Oh, SangMook Kim, and Se-Young Yun.
\newblock Fed{BABU}: Toward enhanced representation for federated image
  classification.
\newblock In \emph{International Conference on Learning Representations}, 2022.
\newblock URL \url{https://openreview.net/forum?id=HuaYQfggn5u}.

\bibitem[Raghu et~al.(2020)Raghu, Raghu, Bengio, and Vinyals]{Raghu2020Rapid}
Aniruddh Raghu, Maithra Raghu, Samy Bengio, and Oriol Vinyals.
\newblock Rapid learning or feature reuse? towards understanding the
  effectiveness of maml.
\newblock In \emph{International Conference on Learning Representations}, 2020.
\newblock URL \url{https://openreview.net/forum?id=rkgMkCEtPB}.

\bibitem[Russakovsky et~al.(2015)Russakovsky, Deng, Su, Krause, Satheesh, Ma,
  Huang, Karpathy, Khosla, Bernstein, et~al.]{russakovsky2015imagenet}
Olga Russakovsky, Jia Deng, Hao Su, Jonathan Krause, Sanjeev Satheesh, Sean Ma,
  Zhiheng Huang, Andrej Karpathy, Aditya Khosla, Michael Bernstein, et~al.
\newblock Imagenet large scale visual recognition challenge.
\newblock \emph{International journal of computer vision}, 115\penalty0
  (3):\penalty0 211--252, 2015.

\bibitem[Shi et~al.(2021)Shi, Chen, Zhang, Zhan, and Wu]{shi2021overcoming}
Guangyuan Shi, Jiaxin Chen, Wenlong Zhang, Li-Ming Zhan, and Xiao-Ming Wu.
\newblock Overcoming catastrophic forgetting in incremental few-shot learning
  by finding flat minima.
\newblock In \emph{Advances in Neural Information Processing Systems},
  volume~34, 2021.

\bibitem[Shim et~al.(2021)Shim, Mai, Jeong, Sanner, Kim, and
  Jang]{shim2021online}
Dongsub Shim, Zheda Mai, Jihwan Jeong, Scott Sanner, Hyunwoo Kim, and Jongseong
  Jang.
\newblock Online class-incremental continual learning with adversarial shapley
  value.
\newblock In \emph{Proceedings of the AAAI Conference on Artificial
  Intelligence}, volume~35, pages 9630--9638, 2021.

\bibitem[Tao et~al.(2020)Tao, Hong, Chang, Dong, Wei, and Gong]{tao2020few}
Xiaoyu Tao, Xiaopeng Hong, Xinyuan Chang, Songlin Dong, Xing Wei, and Yihong
  Gong.
\newblock Few-shot class-incremental learning.
\newblock In \emph{Proceedings of the IEEE/CVF Conference on Computer Vision
  and Pattern Recognition (CVPR)}, June 2020.

\bibitem[Wah et~al.(2011)Wah, Branson, Welinder, Perona, and
  Belongie]{wah2011caltech}
Catherine Wah, Steve Branson, Peter Welinder, Pietro Perona, and Serge
  Belongie.
\newblock The caltech-ucsd birds-200-2011 dataset.
\newblock 2011.

\bibitem[Wang et~al.(2019)Wang, Chao, Weinberger, and van~der
  Maaten]{wang2019simpleshot}
Yan Wang, Wei-Lun Chao, Kilian~Q Weinberger, and Laurens van~der Maaten.
\newblock Simpleshot: Revisiting nearest-neighbor classification for few-shot
  learning.
\newblock \emph{arXiv preprint arXiv:1911.04623}, 2019.

\bibitem[Yu et~al.(2020)Yu, Zhang, Deng, Yuan, Jia, and Chen]{yu2020devil}
Haiyang Yu, Ningyu Zhang, Shumin Deng, Zonggang Yuan, Yantao Jia, and Huajun
  Chen.
\newblock The devil is the classifier: Investigating long tail relation
  classification with decoupling analysis.
\newblock \emph{arXiv preprint arXiv:2009.07022}, 2020.

\bibitem[Zhang et~al.(2021)Zhang, Song, Lin, Zheng, Pan, and Xu]{zhang2021few}
Chi Zhang, Nan Song, Guosheng Lin, Yun Zheng, Pan Pan, and Yinghui Xu.
\newblock Few-shot incremental learning with continually evolved classifiers.
\newblock In \emph{Proceedings of the IEEE/CVF Conference on Computer Vision
  and Pattern Recognition (CVPR)}, pages 12455--12464, June 2021.

\bibitem[Zhou et~al.(2022)Zhou, Wang, Ye, Ma, Pu, and Zhan]{zhou2022forward}
Da-Wei Zhou, Fu-Yun Wang, Han-Jia Ye, Liang Ma, Shiliang Pu, and De-Chuan Zhan.
\newblock Forward compatible few-shot class-incremental learning.
\newblock In \emph{Proceedings of the IEEE/CVF Conference on Computer Vision
  and Pattern Recognition (CVPR)}, 2022.

\bibitem[Zhu et~al.(2021)Zhu, Cheng, Zhang, and Liu]{zhu2021class}
Fei Zhu, Zhen Cheng, Xu-yao Zhang, and Cheng-lin Liu.
\newblock Class-incremental learning via dual augmentation.
\newblock \emph{Advances in Neural Information Processing Systems}, 34, 2021.

\end{thebibliography}

\end{document}